\documentclass[a4paper,fleqn]{cas-sc}

\usepackage[numbers]{natbib}

\usepackage{amssymb}
\usepackage{amsmath}
\usepackage{amsthm}

\usepackage{algorithm}
\usepackage{array}
\usepackage[caption=false,font=normalsize,labelfont=sf,textfont=sf]{subfig}

\theoremstyle{plain}
\newtheorem{theorem}{Theorem}[section]

\theoremstyle{definition}
\newtheorem{definition}[theorem]{Definition}

\theoremstyle{remark}

\usepackage{booktabs}
\usepackage{multirow}
\usepackage{threeparttable}
\usepackage{color}
\usepackage{threeparttable}
\usepackage{setspace}

\usepackage{hyperref}

\usepackage{algorithm}
\usepackage{algorithmic}

\def\tsc#1{\csdef{#1}{\textsc{\lowercase{#1}}\xspace}}
\tsc{WGM}
\tsc{QE}

\makeatletter
\let\printorcid\relax   
\makeatother

\begin{document}

\shorttitle{}    

\shortauthors{J. Zhang et~al.}

\title [mode = title]{Understanding Token-level Topological Structures in Transformer-based Time Series Forecasting}  



%









\author[label1,label2]{Jianqi Zhang}\fnmark[1]
\ead{jluzhangjianqi@163.com}   

\author[label1,label2]{Wenwen Qiang}\cormark[1]\fnmark[1]
\ead{qiang.ww0922@gmail.com}
\cortext[1]{Corresponding author}
\fntext[1]{These authors contributed equally to this work.}

\author[label1,label2]{Jingyao Wang}
\ead{wangjingyao2023@iscas.ac.cn}   

\author[label3]{Jiahuan Zhou}
\ead{jiahuanzhou@pku.edu.cn} 

\author[label1,label2]{Changwen Zheng}
\ead{changwen@iscas.ac.cn}   

\author[label4]{Hui Xiong}
\ead{xionghui@ust.hk}

\affiliation[label1]{organization={University of Chinese Academy of Sciences},
            city={Beijing}, country={China}}
\affiliation[label2]{organization={National Key Laboratory of Space Integrated Information System, Institute of Software, Chinese Academy of Sciences},
            city={Beijing}, country={China}}
\affiliation[label3]{organization={Wangxuan Institute of Computer Technology, Peking University},
            city={Beijing}, country={China}}
\affiliation[label4]{organization={Hong Kong University of Science and Technology},
            city={Hong Kong}, country={China}}




%















\begin{abstract}
Transformer-based methods have achieved state-of-the-art performance in time series forecasting (TSF) by capturing positional and semantic topological relationships among input tokens. However, it remains unclear whether existing Transformers fully leverage the intrinsic topological structure among tokens throughout intermediate layers. Through empirical and theoretical analyses, we identify that current Transformer architectures progressively degrade the original positional and semantic topology of input tokens as the network deepens, thus limiting forecasting accuracy. Furthermore, our theoretical results demonstrate that explicitly enforcing preservation of these topological structures within intermediate layers can tighten generalization bounds, leading to improved forecasting performance. Motivated by these insights, we propose the Topology Enhancement Method (TEM), a novel Transformer-based TSF method that explicitly and adaptively preserves token-level topology. TEM consists of two core modules: 1) the Positional Topology Enhancement Module (PTEM), which injects learnable positional constraints to explicitly retain original positional topology; 2) the Semantic Topology Enhancement Module (STEM), which incorporates a learnable similarity matrix to preserve original semantic topology. To determine optimal injection weights adaptively, TEM employs a bi-level optimization strategy. The proposed TEM is a plug-and-play method that can be integrated with existing Transformer-based TSF methods. Extensive experiments demonstrate that integrating TEM with a variety of existing methods significantly improves their predictive performance, validating the effectiveness of explicitly preserving original token-level topology. Our code is publicly available at: \href{https://github.com/jlu-phyComputer/TEM}{https://github.com/jlu-phyComputer/TEM}.
\end{abstract}


\begin{highlights}
\item We empirically find that existing Transformer-based TSF methods progressively degrade the original token-level topological structure (both positional and semantic) as network depth increases, thereby harming prediction performance.
\item We propose a plug-and-play method, TEM, which explicitly preserves token-level topology in the intermediate layers of the network to improve prediction performance.
\item Theoretically, we prove that TEM can tighten the model’s generalization bound. Empirically, we show that applying TEM to multiple Transformer-based TSF methods significantly improves their prediction performance.
\end{highlights}


\maketitle

\section{Introduction}
\label{Introduction}
Time series forecasting (TSF) is crucial in many practical applications, including weather prediction \cite{abhishek2012weather, karevan2020transductive,li2024functional}, energy planning \cite{boussif2024improving,novo2022planning}, and traffic flow forecasting \cite{fang2023stwave+,li2022dmgan,wang2022st}. Its goal is to predict future value trends over a certain historical observation. Recently, deep learning approaches (e.g., RNNs \cite{sbrana2020n}, CNNs \cite{liu2022scinet}, Transformer \cite{Transformer}, and MLPs \cite{lin2024sparsetsf}) dominate TSF due to their ability to model complex, nonlinear patterns. Among these, Transformer-based models \cite{wu2020deep, lim2021temporal, liu2023itransformer} leverage self-attention to capture both short-range and long-range dependencies in parallel, consistently achieving state-of-the-art (SOTA) performance on TSF benchmarks.

Essentially, Transformer-based TSF methods can be understood as learning from the underlying topological structures among tokens \cite{zhou2023expanding, PatchTST, liu2023itransformer}. Specifically, tokens are obtained by dividing the input sequence into multiple subsequences, each corresponding to one token. The topological structures between these tokens generally fall into two categories. The first is positional topology, which refers to the positional relations between tokens (Definition \ref{def:GNI}). The second is semantic topology, which refers to the similarity of features between tokens (Definition \ref{def:SNI}). Positional topology is typically modeled using positional encoding (PE). At the input layer, the model injects PE into each token to provide explicit positional information, helping the model identify temporal patterns such as trends, thereby improving performance. For example, in the Weather dataset \cite{liu2023itransformer}, if the model knows that ``9 AM precedes 1 PM'' and the temperature at 9 AM is lower, it can recognize an upward trend in temperature during that period. Semantic topology is commonly modeled through the attention mechanism, which computes pairwise token similarity to capture topological structure within the sequence. This capability enables the model to identify temporally significant patterns—such as periodicity or recurring trends—by attending to tokens with similar semantic features. For instance, in the Weather dataset, attention weights may highlight strong similarities in variables observed at 24-hour intervals, allowing the model to discern and utilize these periodic spans for more accurate forecasting.

Although recent strategies for modeling token-level topology have substantially boosted the forecasting performance of Transformer-based TSF models, it remains unclear whether these methods fully leverage the intrinsic topological structures present at the token level. To clarify this, we revisit existing approaches through the perspective of structure preservation, a concept inspired by Locality Preserving Projection (LPP) \cite{he2003locality, he2005face, cai2005document, wang2020stacked}. Structure preservation emphasizes maintaining the ``original topological structure'' among the input data—characterized by distances, similarities, and adjacency—during feature transformations. This practice produces models that are adept at capturing discriminative information in patterns, thereby improving downstream prediction performance \cite{zhang2022unified, zhang2022feature, qiang2021robust}. In the context of Transformer-based TSF, we specifically define the original topological structure among the input data as the positional and semantic topology among input tokens in the input space, e.g., the Original Positional Topology (OPT)
and the Original Semantic Topology (OST) (see Definition \ref{def:Local_s}). A detailed analysis (see Section \ref{motivation} for empirical evidence) indicates that, A detailed analysis (see Section \ref{motivation} for empirical evidence) indicates that, although existing Transformer-based TSF models implicitly capture certain topological structures, they do not explicitly enforce intermediate layers to preserve these structures. Consequently, successive transformations within these layers may unintentionally distort the information about the original topological structure among input tokens. Motivated by this observation, we hypothesize that explicitly enforcing structure preservation throughout the Transformer architecture could serve as a stronger inductive bias, thereby further enhancing TSF accuracy.

To empirically validate our hypothesis, we conduct a series of experiments (see Section \ref{motivation} for details) to investigate the impact of enforcing structure preservation across the intermediate layers of Transformer models. Specifically, we explicitly incorporate the OPT and the OST into intermediate-layer features, aiming to help the model better retain these original topological structures. In addition, since both Rotary Position Embedding (RoPE) and Convolutional Position Embedding (CPE) can directly inject OPT into the intermediate layers of the network, we further extend our method by incorporating OST into both the input and intermediate layers based on these two position-encoding strategies, and conduct experiments to validate its effectiveness. The detailed procedure of this incorporation is presented in Section \ref{motivation}. After retraining the models under these conditions, we evaluate their predictive performance. As shown in Figures \ref{fig_moti_3} and \ref{fig_moti_4}, the experimental results demonstrate improved forecasting accuracy in most scenarios, although the magnitude of improvement varies across different forecasting tasks. To further understand the underlying causes of these empirical observations, we theoretically analyze how structure preservation influences the performance of Transformer-based TSF methods. Specifically, we prove two key findings: 1) explicitly maintaining the OPT and OST throughout deeper Transformer layers helps tighten the model’s generalization bound, thereby potentially improving its predictive generalization; 2) the extent of this improvement is inherently layer-dependent, explaining the observed variability in performance gains across different tasks. Consequently, we argue that adaptively enforcing the preservation of the original topological structure according to the characteristics of each layer in the model is essential for consistently achieving good forecasting performance.

To this end, we propose a novel Transformer-based method for TSF, termed Topology Enhancement Method (TEM). The core idea of TEM is to adaptively enforce the preservation of the original topological structure among input tokens, thus improving the predictive accuracy of Transformer-based models. Specifically, TEM consists of two primary modules: 1) Positional Topology Enhancement Module (PTEM), which strengthens the preservation of OPT by injecting a learnable weighted positional encoding (PE) constraint into intermediate-layer features; 2) Semantic Topology Enhancement Module (STEM), which enhances the preservation of OST by incorporating a learnable weighted similarity matrix among input tokens into intermediate-layer representations. To ensure optimal topology preservation, we introduce a bi-level optimization strategy, enabling the modules to learn optimal injection weights adaptively. The proposed TEM is a plug-and-play method that can be easily integrated with existing Transformer-based TSF methods. Extensive experiments on various real-world datasets demonstrate the effectiveness and superiority of our proposed framework. The main contributions of this paper are summarized as follows:
\begin{itemize}
    \item Through comprehensive empirical and theoretical analyses, we reveal two critical insights: 1) the original topological structure among input tokens is progressively weakened within intermediate layers of existing Transformer-based TSF models, potentially harming forecasting accuracy; 2) explicitly preserving the original topological structures among input tokens throughout the Transformer’s intermediate layers substantially enhances forecasting performance, building effectively upon existing Transformer-based TSF methods.
    \item Motivated by the above insights, we propose a novel Transformer-based approach, called Topology Enhancement Method (TEM), which adaptively preserves the original topological structure among input tokens. TEM consists of two modules: 1) the Positional Topology Enhancement Module (PTEM), which injects a learnable weighted PE to explicitly maintain OPT among tokens; and 2) the Semantic Topology Enhancement Module (STEM), which introduces a learnable weighted similarity matrix to explicitly preserve OST among tokens.
    \item To ensure optimal preservation of the original topological structure, we introduce a bi-level optimization strategy that adaptively learns optimal weights for OPT and OST injection. Extensive experimental evaluations on multiple real-world benchmark datasets verify the effectiveness and superiority of the proposed TEM.
\end{itemize}

\section{Related Work}
\textbf{Transformer-based TSF methods:} Transformer-based methods are one of the mainstream approaches for TSF. Currently, many improvements have been made to these methods. Some methods \cite{Autoformer,fedformer,liang2023does} combine the Fourier transform with the Transformer to leverage the periodicity of time series. GCformer \cite{zhao2023gcformer} integrates convolution with the Transformer. Others \cite{du2023preformer,tang2023infomaxformer} enhance the attention mechanism to improve TSF performance. \cite{han2024capacity} enhances the Transformer's forecasting performance through residual regularization (PRReg). PatchTST \cite{PatchTST} embeds multiple adjacent time points into one token to address the insufficient temporal information of a single token.  Crossformer \cite{Crossformer} and DSformer \cite{yu2023dsformer} utilize both temporal and variable tokens for TSF. iTransformer \cite{liu2023itransformer} shows that using only variable tokens can yield good results. More details about temporal/variable tokens' details can be found in Appendix \ref{Token_Acquisition}. Regardless of how these methods are designed, they all rely on the positional/semantic topology to assist the model in making predictions. 

\textbf{Positional \& semantic topology in transformer-based TSF methods:} Regarding positional topology (see Definition \ref{def:GNI} for details), existing studies mainly enhance a model’s ability to capture such topological relations by designing new positional encodings (PE). There is extensive work on PE in other domains, such as natural language processing (NLP) and vision. \cite{su2024roformer} proposes Rotary Position Embedding (RoPE) to improve Transformer performance in NLP. \cite{shaw2018self} uses relative positional encoding for NLP tasks. \cite{ke2020rethinking} decouples token series from PE and explores a new way to inject PE. \cite{chu2021conditional} and \cite{liu2021swin} propose Convolutional Position Embedding (CPE) and relative positional encoding, respectively, to boost Transformer performance in vision. Inspired by these above works, the researchers in TSF have also proposed a series of methods. \cite{irani2025dywpe} uses a discrete wavelet transform to generate PE and injects it at the input layer. \cite{irani2025dywpe} proposes continuous PE, injected at the input layer. \cite{messou2025tsformer} introduces a learnable PE, injected at the input layer. \cite{lv2025toward} injects a fixed relative PE at every layer. Of course, beyond these TSF-specific designs, the most commonly used choices in TSF remain sinusoidal PE and learnable PE \cite{irani2025positional}. For semantic topology (see Definition \ref{def:SNI} for details), recent works primarily strengthen the semantics encoded in tokens by improving tokenization schemes \cite{PatchTST,liu2023itransformer} or employing multiple types of tokens \cite{Crossformer,yu2023dsformer}. This enables the attention mechanism to better capture semantic topological relations. Unlike the above methods, which focus on only one topology, this paper jointly considers both positional and semantic topologies in transformer-based TSF methods, leading to improved forecasting performance.

\textbf{Structure Preservation in TSF: } Building on the works in Locality Preserving Projections (LPP) \cite{he2003locality, he2005face, cai2005document, wang2020stacked}, preserving the original topological structure of input data during feature mapping has a positive effect on improving a model’s predictive performance. The positional and semantic topologies discussed above can be regarded as forms of the input data’s original topology. For preserving positional topology in TSF, \cite{lv2025toward} can achieve this preservation by injecting fixed relative PE into every layer. Moreover, applying positional encodings that are injected at each layer from other domains (such as RoPE \cite{su2024roformer} and CPE \cite{chu2021conditional}) to Transformer-based TSF methods can also maintain positional topology. However, these approaches do not finely control the amount of positional topology injected at each layer. This may result in too little injection, yielding limited predictive gains, or too much injection, which can adversely affect predictions. Our experimental results in Subsection \ref{motivation} further prove this. In TSF, research on preserving semantic topology remains scarce and warrants further exploration. This paper addresses the preservation of both positional and semantic topologies in Transformer-based TSF and proposes an adaptively fine-controlled injection mechanism, aiming to inject the optimal amount of positional and semantic topology to further enhance predictive performance.

Notably, this paper is the first in the TSF field to adaptively preserve and control both positional and semantic topology at the intermediate layers. Although methods such as RoPE \cite{su2024roformer}, CPE \cite{chu2021conditional}, and GL-PE \cite{lv2025toward} preserve positional topology by injecting PE into intermediate layers, they lack fine-grained control over the injection strength. As a result, overly weak injection may yield limited gains, whereas overly strong injection can interfere with prediction and degrade performance. Our experiment (see Figure \ref{fig_bar}) further demonstrates that under the same experimental conditions as our method, the predictive gains of these methods are significantly lower than ours. This indicates that fine-grained regulation of the amount of topology injected is crucial for improving prediction. In this work, we not only adaptively preserve and control positional and semantic topology at the intermediate layers, aiming to maximize the benefits of injecting topological information, but also provide a theoretical justification for the necessity of finely tuning the amount of topological information injected at each layer.

\section{Problem Analysis and Motivation}

In this section, we first introduce the important definitions and notations related to this work. Then, through experiments and theoretical analysis, we investigate the ability of existing models to preserve the original topological structure of input tokens in intermediate features, as well as their impact on predictive performance, which provides important insights for the design of our method.

\subsection{Definition and Notation}
\label{SEC_notation}

\textbf{Time Series Forecasting:} Given a historical time series composed of observations $X=\{\mathbf{x}_1,\mathbf{x}_2,\cdots,\mathbf{x}_T\}\in\mathbb{R}^{T\times N}$, the goal of time series forecasting (TSF) is to predict future observations $Y=\{\mathbf{x}_{T+1},\mathbf{x}_{T+2},\cdots,\mathbf{x}_{T+S}\}\in\mathbb{R}^{S\times N}$. Here, $T$ is the lookback window length (number of historical timestamps), and $S$ is the forecast horizon (number of future timestamps to predict). Specifically, each $\mathbf{x}_i\in\mathbb{R}^{N}$ represents the multivariate observation vector at timestamp $i$, consisting of values from $N$ variables.

\textbf{Tokenizer:} In Transformer-based TSF, the input time series is first segmented into a sequence of tokens, which are then processed through a stack of network layers to generate the final forecast. We refer to the outputs at each intermediate layer as intermediate-layer token features. Various tokenization schemes have been proposed—such as patch-level, channel-wise, and multiscale tokens—each designed to capture different structural or semantic aspects of the input sequence. A comprehensive overview of these token types is provided in Appendix \ref{Token_Acquisition}.

\textbf{Topological Structures:} 
As discussed in the introduction, the topological structures between tokens consist of two categories: positional topology and semantic topology. Positional topology can help the model identify temporal patterns such as trends, while semantic topology enables the model to discover key signals that share similar characteristics (e.g., periodicity). Formal definitions of these two topologies are provided below.

\begin{definition}[Positional Topology]
\label{def:GNI}
Let $N_t$ be the number of tokens and
$\mathcal{P}=\{1,\cdots,{N_t}\}$ be the index set of tokens' positions. The \emph{positional topology} is presented as:
\begin{equation}
  \mathcal{G} = \bigl\{\, p(k) \bigm|\;
  k \in \mathcal{P},  p \in \Upsilon \bigr\},
\end{equation}
where $\Upsilon$ denotes the class of functions used to generate positional encodings (PE).
\end{definition}

In Transformer-based TSF methods, the input token is first converted into intermediate-layer token features through an embedding layer. Initially, these tokens features do not carry any positional information \cite{Transformer}. Positional information is modeled explicitly by adding PE to this tokens features \cite{Transformer}. Since the number of tokens typically remains constant across Transformer layers, the positional information associated with each token is expected to stay consistent throughout the layers.

\begin{definition}[Semantic Topology] 
\label{def:SNI}
Let the $i$-th intermediate-layer token feature output by the $l$-th layer be denoted as $H_i^l$, where $i \in \{ 1,\cdots,N_t\}$ denotes the index of token feature, $l \in \{ 1,\cdots,L\} $ denotes the index of layer. $l=0$ means that the token feature equals to the input token. The \textit{semantic topology} at the $l$-th layer is defined as the set of feature similarity among these tokens: 
\begin{equation}
\label{S_temporal}
        \mathcal{S}^l = \left\{\text{sim}(H_i^l, H_j^l)|\, i, j \in \{1, \cdots, N_t\} \right\}, l \in \{0, 1, \cdots, L\},  
\end{equation}
where $\text{sim}(\cdot)$ is the similarity metric. 
\end{definition} 

In Transformer-based TSF methods, the similarity between token features is naturally modeled by the Transformer's attention mechanism. Specifically, the attention mechanism first computes pairwise token similarities by calculating the dot product between the Query and Key vectors. These similarities are then normalized through the softmax function to generate attention scores, which directly quantify the strength of the semantic dependencies among tokens. Thus, the attention mechanism inherently captures and the semantic topology in Transformer models.

\textbf{Original Topological Structure among Input Tokens in TSF: }
Locality Preserving Projection (LPP) \cite{he2003locality} defines the ``original topological structure'' of the input data based on two key aspects: the spatial distances and the feature similarities between neighboring samples in the input space. Analogously, in Transformer-based TSF, the original topological structure can be represented by the positional distances and feature similarities among input tokens in the input space. According to Definitions \ref{def:GNI} and \ref{def:SNI}, these positional distances and feature similarities exactly correspond to the positional and semantic topology among tokens. Therefore, we define the original topological structure for Transformer-based TSF methods as follows:

\begin{definition}[Original Topological Structure among Input Tokens of Transformer-based TSF] 
\label{def:Local_s}
In the context of Transformer-based TSF, the original topological structure among input tokens, denoted as $\Theta$, is composed of the positional topology and semantic topology among the input tokens in the input space, expressed as:
\begin{equation}
\label{eq_Original_Local}
        \Theta = \{\mathcal{G}, \mathcal{S}^0\}  
\end{equation}
Thus, we name $\mathcal{G}$ as the Original Positional Topology (OPT) and $\mathcal{S}^0$ as the Original Semantic Topology (OST).
\end{definition}

By definition, positional topology is computed directly based on the positional relationships among tokens in the input space. Therefore, it is reasonable to adopt $\mathcal{G}$ as the OPT. Similarly, $\mathcal{S}^0$ represents the semantic topology at the $0$-th layer, which corresponds to the semantic relationships in the original input space. Hence, using $\mathcal{S}^0$ as the OST is also well-justified.

\begin{figure*}
  \centering
  \subfloat[]{\includegraphics[width=.25\textwidth]{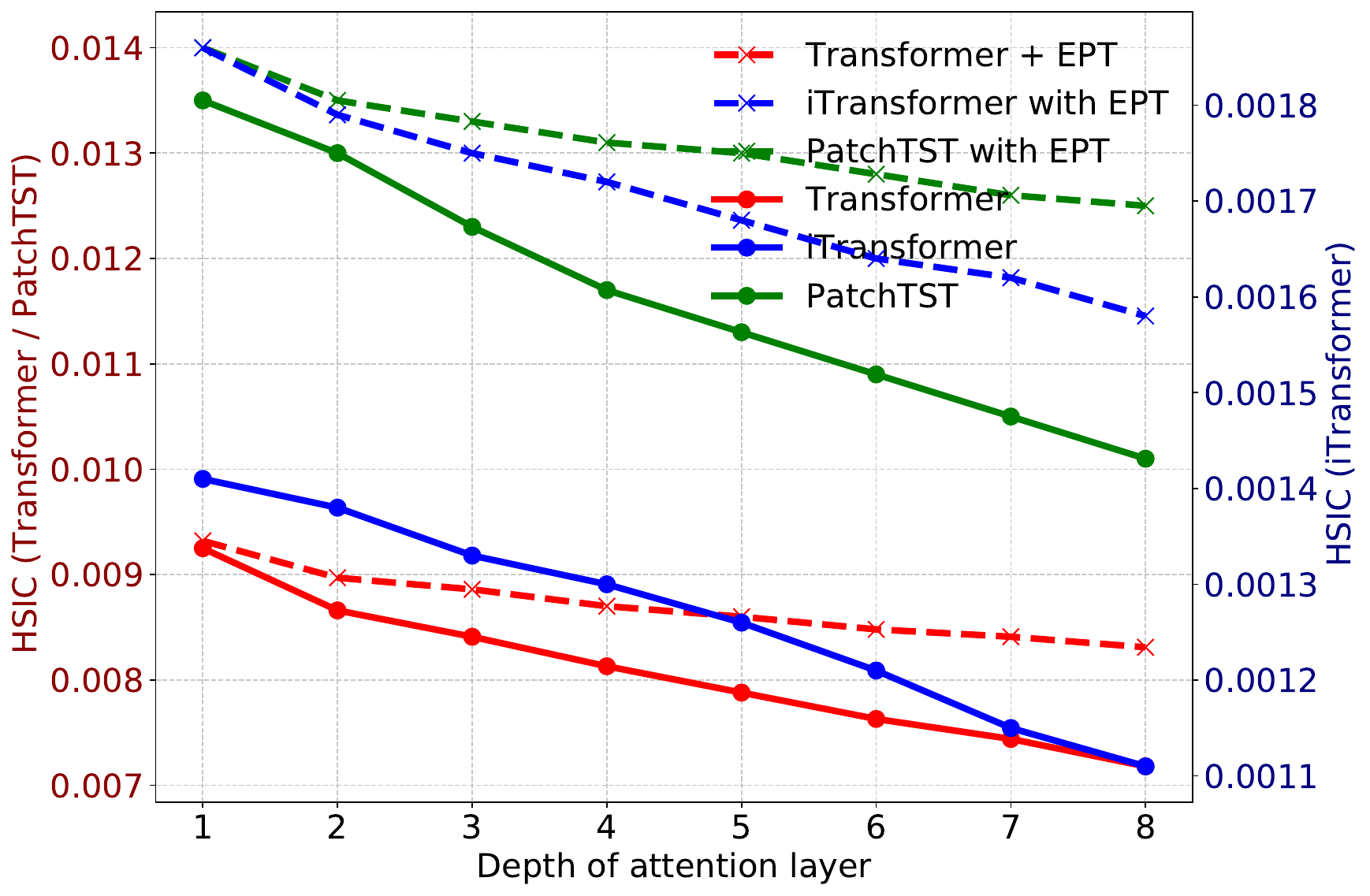}\label{fig_moti_1}}
  \hfill
  \subfloat[]{\includegraphics[width=.23\textwidth]{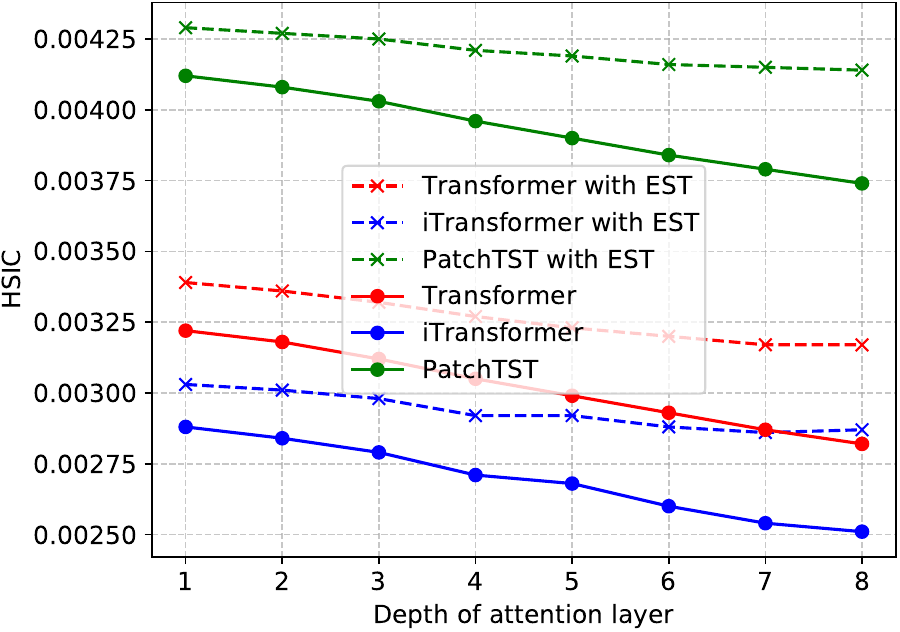}\label{fig_moti_2}}
  \subfloat[]{\includegraphics[width=.25\textwidth]{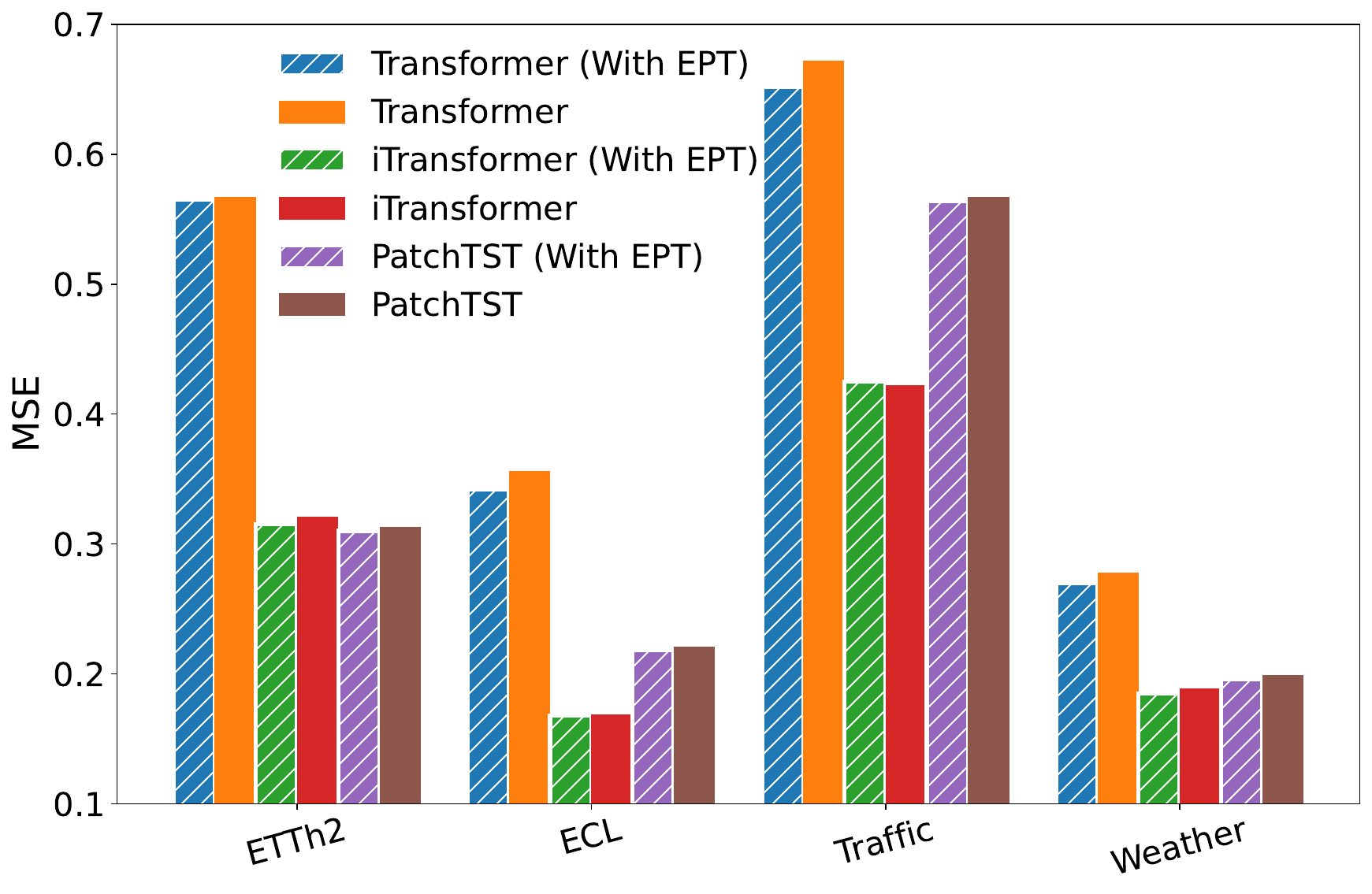}\label{fig_moti_3}}
  \hfill
  \subfloat[]{\includegraphics[width=.25\textwidth]{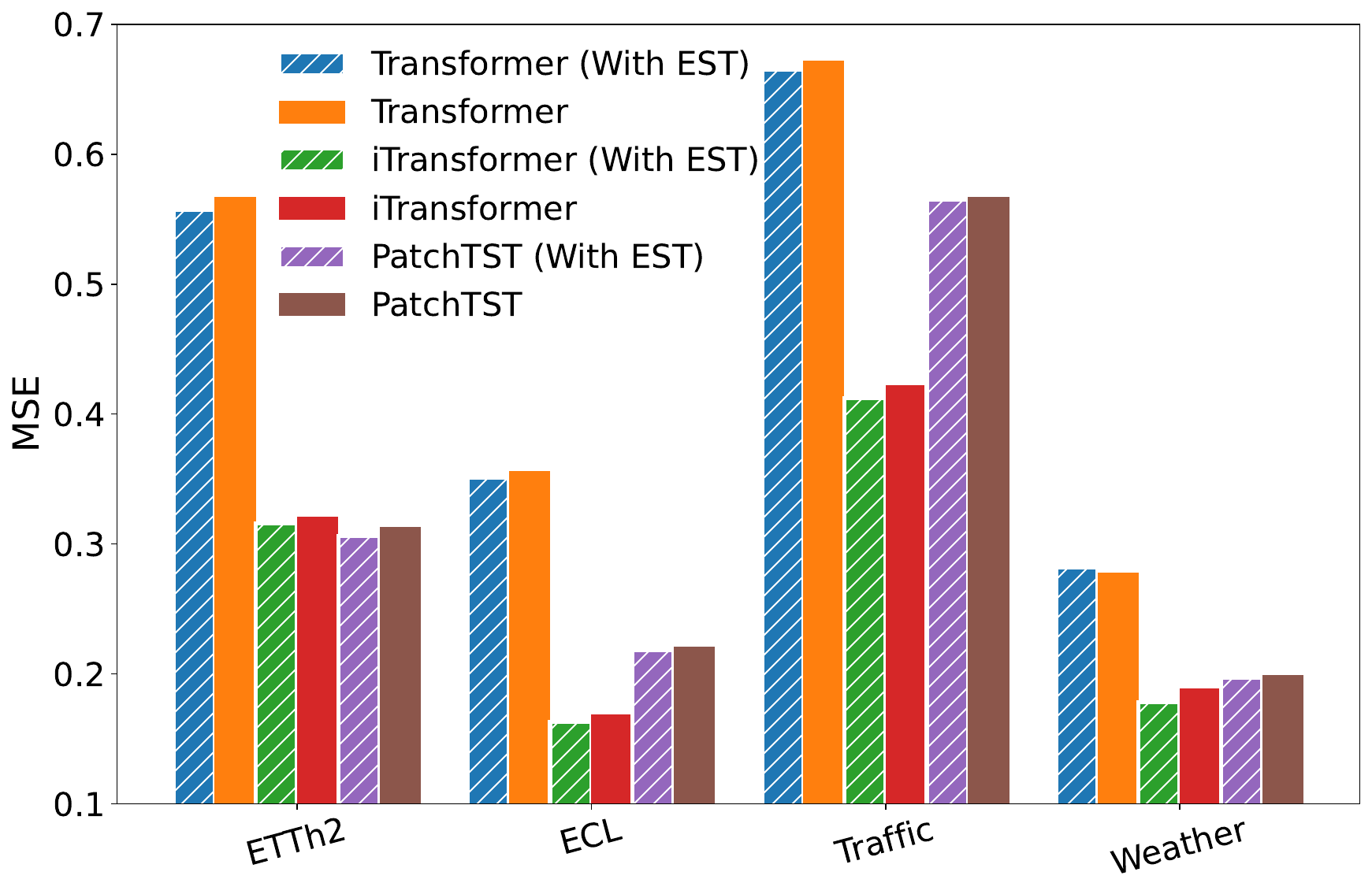}\label{fig_moti_4}}
  \caption{(a) Illustration of the change in the value of HSIC between the PE and the output feature of each encoder layer. (b) Illustration of the change in the value of HSIC between the similarity matrix of input tokens and the similarity matrix of output tokens of each encoder layer. Changes in model performance after enhancing positional topology of the input tokens (c) or semantic topology of the input tokens (d) in deep layers. EPT/EST stands for ``enhanced positional/semantic topology of the input tokens''. }
  \label{fig_moti}
\end{figure*}

\begin{figure*}
  \centering
  \subfloat[]{\includegraphics[width=.33\textwidth]{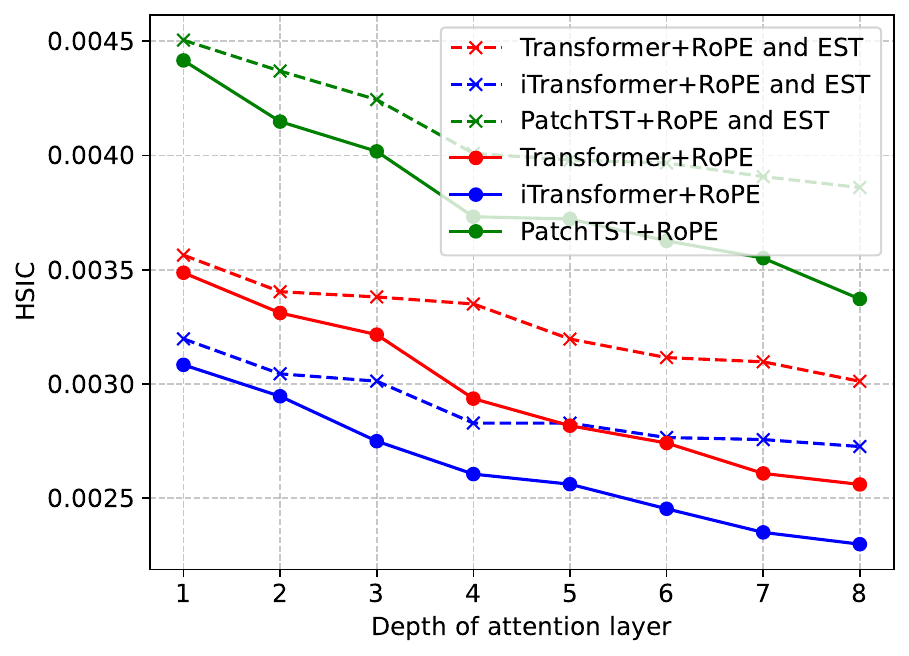}\label{fig_moti1_1}}
  \hfill
  \subfloat[]{\includegraphics[width=.33\textwidth]{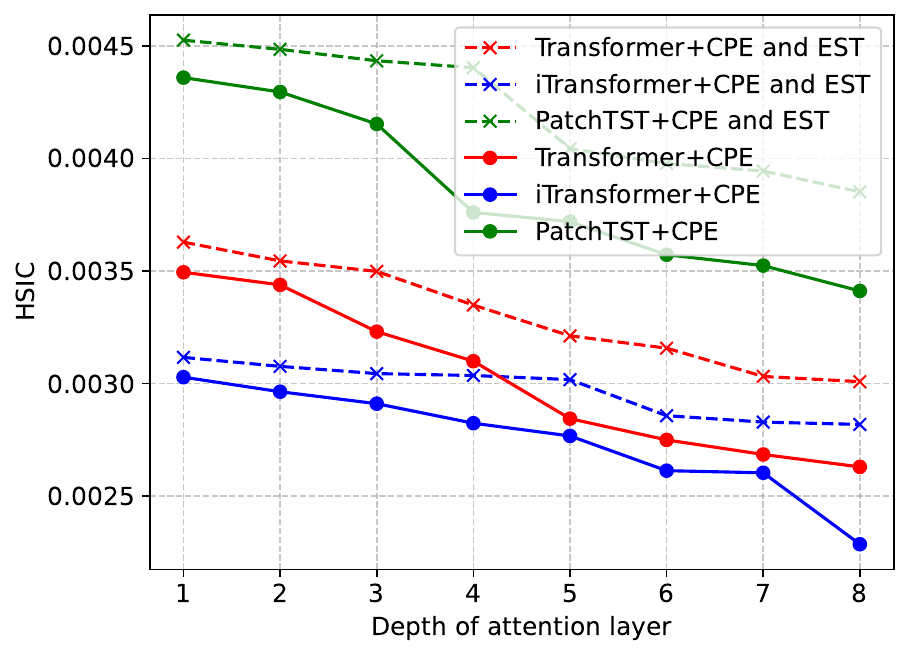}\label{fig_moti1_2}}
  \subfloat[]{\includegraphics[width=.33\textwidth]{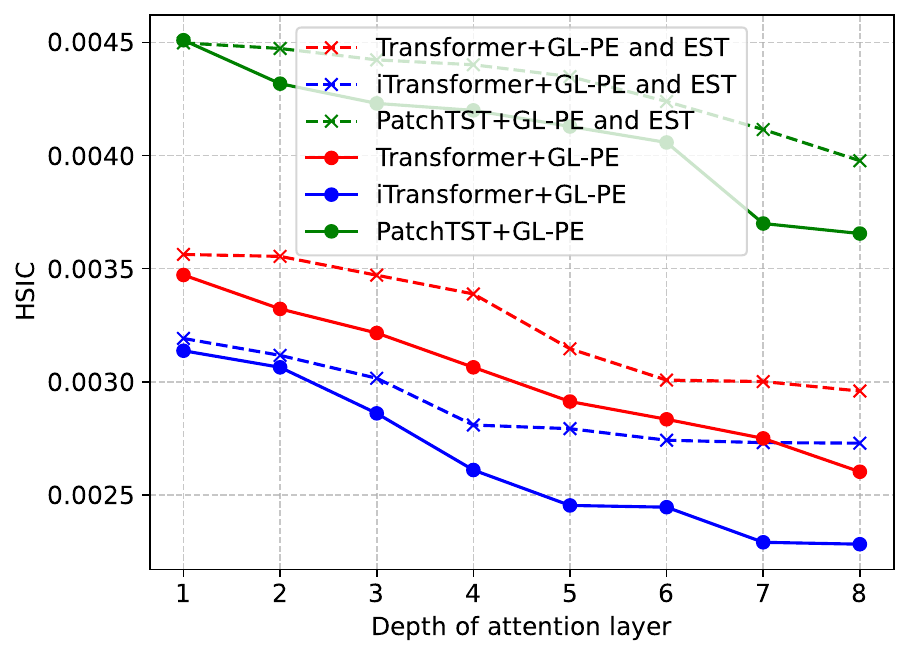}\label{fig_moti1_3}}
  \hfill
  \subfloat[]{\includegraphics[width=.33\textwidth]{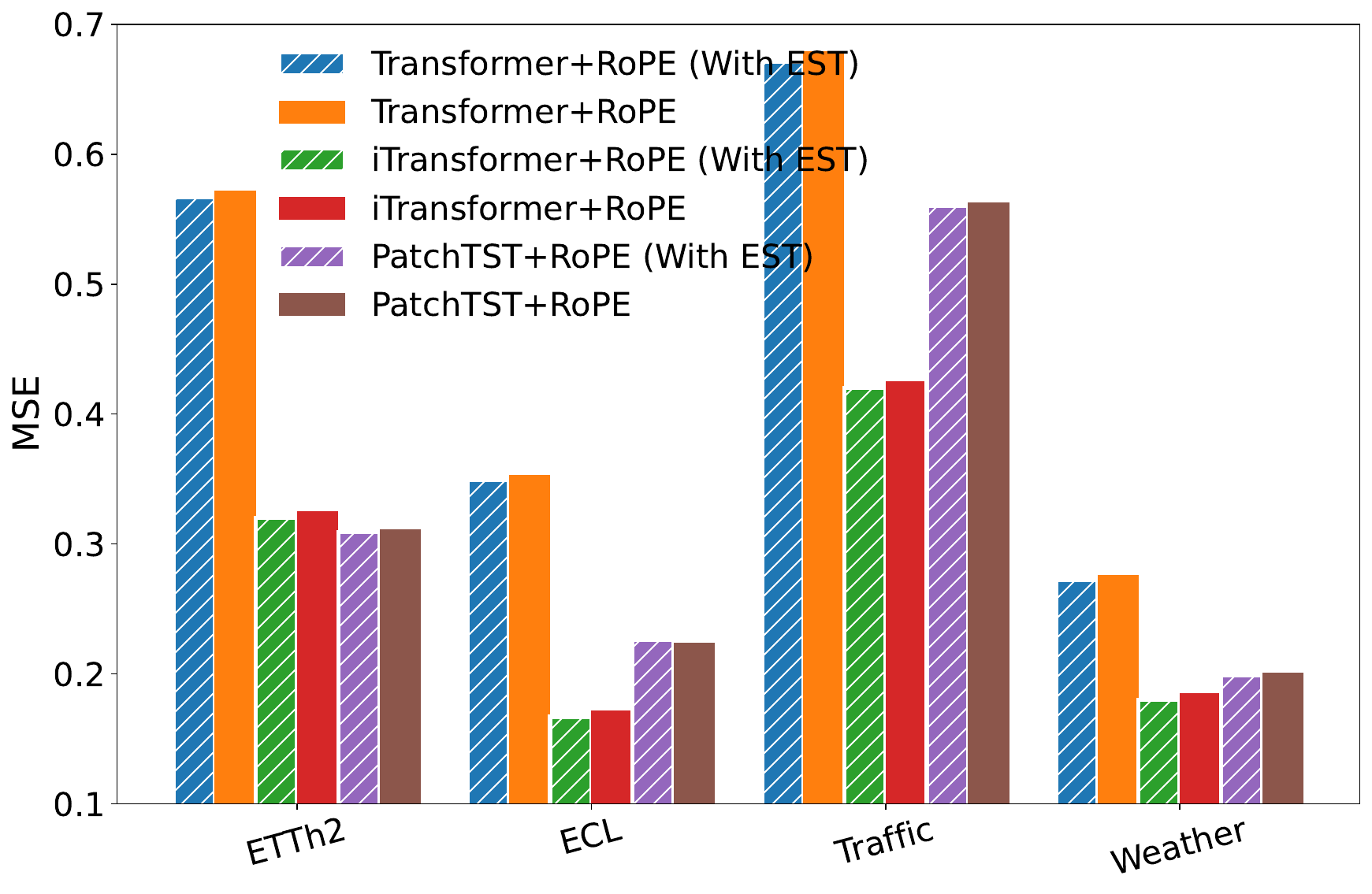}\label{fig_moti1_4}}
  \hfill
  \subfloat[]{\includegraphics[width=.33\textwidth]{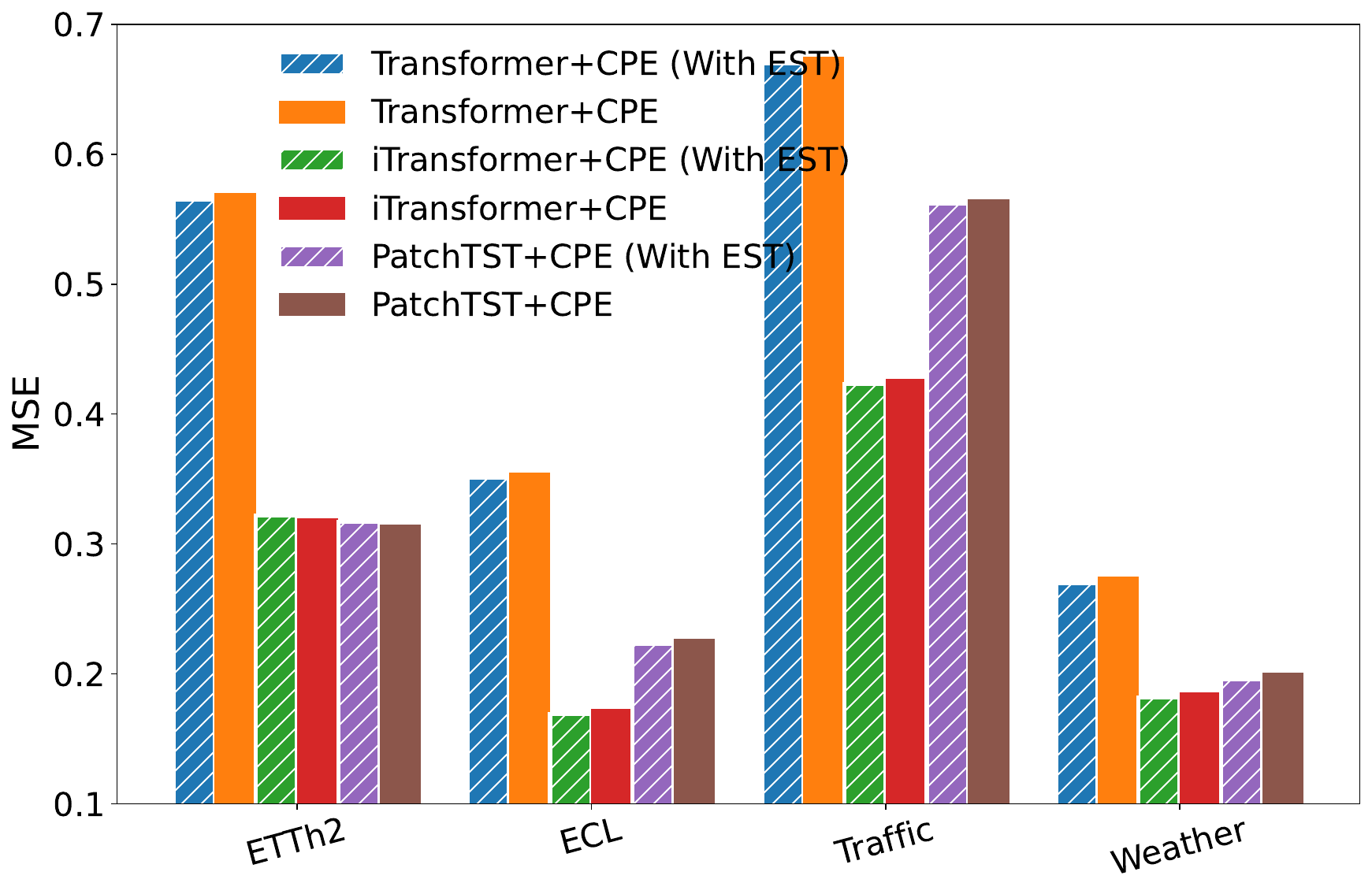}\label{fig_moti1_5}}
  \hfill
  \subfloat[]{\includegraphics[width=.33\textwidth]{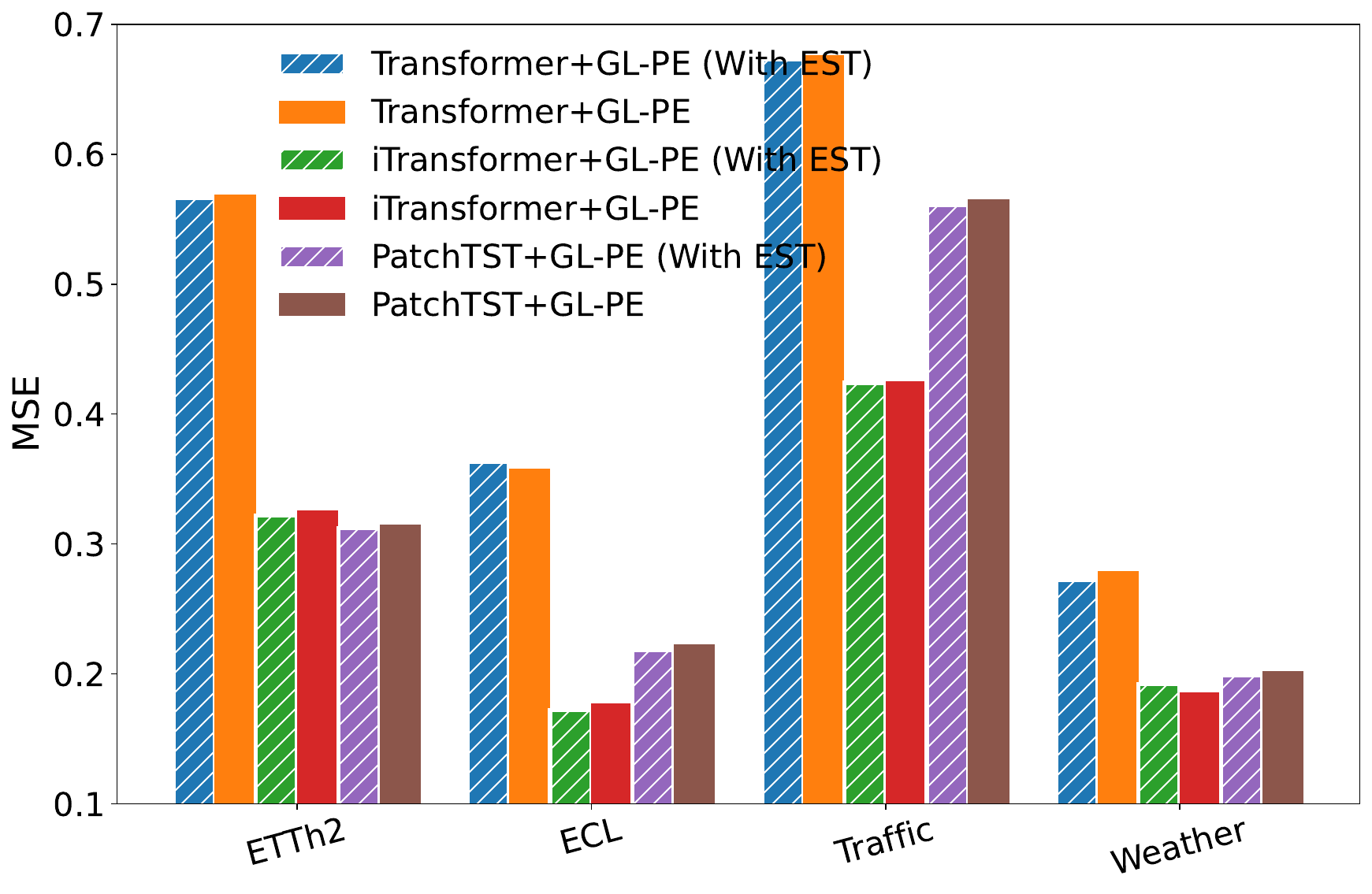}\label{fig_moti1_6}}
  \caption{ (a), (b), (c): Illustration of the change in the value of HSIC between the similarity matrix of input tokens and the similarity matrix of output tokens of each encoder layer.  (d), (e), (f): Changes in model performance after enhancing the semantic topology of the input tokens. EST stands for ``enhanced semantic topology of the input tokens''. Each row from left to right shows the results of the baseline methods using RoPE \cite{su2024roformer}, CPE \cite{chu2021conditional}, and GL-PE \cite{lv2025toward}.
  }
  \label{fig_moti1}
\end{figure*}

\subsection{Empirical Findings}
\label{motivation}
In this subsection, we empirically verify two key findings: 1) the intermediate layers of existing Transformer-based TSF methods do not effectively preserve the original topological structure among input tokens; 2) explicitly enforcing the preservation of this original topological structure can further improve TSF performance.

First, we conduct experiments to evaluate how well Transformer-based models preserve the original topological structures among input tokens across different encoder layers. Specifically, we train three representative models—vanilla Transformer \cite{Transformer}, PatchTST \cite{PatchTST}, and iTransformer \cite{liu2023itransformer}—on four standard datasets: ETTh2, ECL, Traffic, and Weather \cite{liu2023itransformer}. The vanilla Transformer and iTransformer adopt sinusoidal positional encoding (PE) \cite{Transformer}, while PatchTST employs learnable PE \cite{devlin2018bert}. In all experiments, the look-back and prediction lengths are set to 96, and the number of encoder layers is fixed to 8 for consistency and ease of analysis. To assess the preservation of the original local structure, we utilize the Hilbert-Schmidt Independence Criterion (HSIC), which measures the statistical dependence between two variables and is capable of capturing both linear and nonlinear relationships \cite{greenfeld2020robust}. A higher HSIC value indicates a stronger dependency. In our context, a high dependence between 1) PE and the output features, and 2) the input-token similarity matrix and the corresponding output-token similarity matrix at each layer, suggests that the model retains the OPT and OST, respectively. Therefore, HSIC serves as a reasonable proxy for evaluating the extent to which original local structures are preserved in intermediate layers. The detailed computation process of HSIC is provided in Appendix \ref{sec:app_hsic}. Experimental results show that, as the network depth increases, the HSIC values for both OPT preservation and OST preservation consistently decrease across all models (Figures \ref{fig_moti_1} and \ref{fig_moti_2}). This indicates that the preservation of original local structures weakens progressively with depth, highlighting a potential limitation in current Transformer-based TSF models.

Second, we conduct experiments to evaluate the impact of explicitly injecting the original topological structures among input tokens into the intermediate layers of Transformer models. The dataset and model configurations remain the same as those used in the first set of experiments. To enhance the preservation of OPT, we add the $\mathcal{G}$ directly to the output features of each Transformer layer. To enhance the preservation of OST, we incorporate the $\mathcal{S}^0$ into each layer's attention mechanism. Specifically, we multiply this similarity matrix by a scalar and add it to the raw attention logits before the softmax operation. Please refer to Appendix \ref{Sec_why} for further discussion. Then, we retrain and evaluate the above models accordingly. To further verify the generality of the phenomenon, in addition to the experiments above, we include three other PE to compare: RoPE \cite{su2024roformer}, CPE \cite{chu2021conditional}, and GL-PE \cite{lv2025toward}. Because these three methods add PE at every layer of the model, they naturally inject OPT into intermediate layers. We examine how these three PEs affect the three Transformer baselines before and after injecting OST into the intermediate layers. The experimental results in Figures \ref{fig_moti_1}, \ref{fig_moti_2}, \ref{fig_moti1_1}, \ref{fig_moti1_2}, and \ref{fig_moti1_3} indicate that both strategies effectively slow the degradation of layer-wise HSIC values, suggesting stronger preservation of the original topological structure. Furthermore, the results in Figures \ref{fig_moti_3}, \ref{fig_moti_4}, \ref{fig_moti1_4}, \ref{fig_moti1_5}, and \ref{fig_moti1_6} also show that, regardless of the PE adopted by the models, injecting OPT or OST into intermediate layers improves predictive performance in most cases, although the magnitude of improvement varies across tasks.

These two experiments reveal two key observations: 1) under existing strategies employed by Transformer-based TSF methods, the original topological structures among input token are progressively degraded across intermediate layers as the network deepens; 2) explicitly enforcing the preservation of these structures can lead to improved forecasting performance. However, the current findings remain heuristic, and the simple quantitative enhancement strategies applied in these experiments may result in limited and inconsistent gains. These limitations underscore the need for a theoretical understanding of how the original topological structures influence Transformer-based model performance. 

\subsection{Theoretical Analysis}
\label{SEC_TA}

To explain the empirical observations in Section \ref{motivation}, we provide a theoretical analysis of how preserving the OPT and OST affects the performance of Transformer-based TSF methods. Specifically, we first derive a generalization error bound for Transformer-based TSF models based on the Rademacher complexity of each Transformer layer. Then, to investigate the impact of structure preservation, we analyze how the degree to which the original topological structures are retained in intermediate representations influences the Rademacher complexity of each layer, and thus the overall generalization error bound of the model.

Without loss of generality, consider a training set $\mathcal{D} = \{(X_i, Y_i)\}_{i=1}^m$ for the TSF problem. This dataset can equivalently be viewed either as samples drawn from a $\beta$-mixing distribution or as segments extracted from an extended, finite-length time series. Here, the mixing coefficient $\beta(\cdot)$ is defined according to Eq.(1) in \cite{mohri2008rademacher}. Furthermore, this extended time series can be partitioned into $\mu$ mutually independent segments, each with length $a$. In this context, we consider the values at the starting timestamps of each segment as realizations of one random variable and the values at the ending timestamps as realizations of another random variable. Consequently, the coefficient $\beta(\cdot)$ measures the dependence between these two random variables. In what follows, we present the generalization error bound for Transformer-based TSF methods.

\begin{theorem}
\label{theorem:1}
Let $\mathcal{X}$ and $\mathcal{Y}$ denote the input and output spaces for TSF, and let $\mathcal{H}$ be a class of Transformer-based functions $h: \mathcal{X} \rightarrow \mathcal{Y}$. Assume a non-negative, $\xi$-Lipschitz loss function $\ell: \mathcal{Y} \times \mathcal{Y} \rightarrow \mathbb{R}$, bounded above by $M > 0$. Given a training dataset $\mathcal{D} = {(X_i, Y_i)}_{i=1}^m$, with probability at least $1 - \delta$, the following generalization bound holds for any $h \in \mathcal{H}$:
\begin{equation}
    \begin{aligned}
        \mathbb{E}_{(X,Y) \sim \mathcal{X} \times \mathcal{Y}} [\ell(h(X), Y)] \leq \frac{1}{m}\sum_{i=1}^{m} [\ell(h(X_i), Y_i)] + \xi G\left( \left[ \mathfrak{R}_\mu(\mathcal{H}_s^1) \right], \dots, \left[ \mathfrak{R}_\mu(\mathcal{H}_s^L) \right] \right) + M \sqrt{\frac{\log(2/\delta')}{2\mu}},
    \end{aligned}
\end{equation}
where $\delta' = \delta - 4(\mu - 1)\beta(a)$, $\mathfrak{R}_\mu(\cdot)$ is the Rademacher complexity of a function class, $\mathcal{H}_s^i$ is a function class of the $i$-th Transformer layer, and $G(\cdot)$ is an aggregation function over the layer-wise complexities.
\end{theorem}

This theorem provides an upper bound on the generalization error of Transformer-based TSF methods under a $\beta$-mixing distribution \cite{mohri2008rademacher}. The bound consists of three components: the empirical risk over the training dataset, an aggregation of layer-wise Rademacher complexities reflecting the Transformer structure, and a confidence term. Note that the value of $G(\cdot)$ decreases as the Rademacher complexity of any single layer decreases. The detailed proof of Theorem \ref{theorem:1} is provided in Appendix \ref{sec:app_proof1}.

Next, we analyze the relationship between the Rademacher complexity of a single-layer Transformer in Theorem \ref{theorem:1} and the original topological structures preserved in the model. We first formalize several key concepts, including the layer-wise positional topology distortion and the layer-wise semantic topology distortion.

\begin{definition}[Layer-wise positional topology Distortion]
\label{def:pos_distortion}
Let ${\mathcal{G}}$ denote the OPT among the input tokens, $p(k)$ denote the positional encoding (PE) vector of the k-th position, $p_i(k)$ denote the position encoding of the $k$-th token perceived from OPT at layer $i$. The layer-wise positional topology distortion at layer $i$ is defined as:
\begin{equation}
\Delta_{{\mathcal{G}}}^{i} = \frac{1}{N_t^2} \sum_{k=1}^{N_t} \sum_{j=1}^{N_t} \left| \, \|p(k) - p(j)\| - \|p_i(k) - p_i(j)\| \, \right|.
\end{equation}
\end{definition}

\begin{definition}[Layer-wise Semantic Topology Distortion]
\label{def:delta_S}
Fix a similarity function
${\rm{sim}}(\cdot):\mathbb R^{d}\times\mathbb R^{d}\!\to\![0,1]$
(e.g.\ cosine similarity).
For layer $i\in\{0,1,\dots,L\}$, let
$H_k^{\,i}\in\mathbb R^{d_i}$ be the representation of token $k$.
The layer-wise semantic topology distortion at layer $i$ is defined as:
\begin{equation}
      \Delta_{\mathcal{S}}^{i}
  \;=\;
  \frac{1}{N_t^{2}}
  \sum_{k=1}^{N_t}\sum_{j=1}^{N_t}
  \Bigl|
      \operatorname{sim}\!\bigl(H_k^{0},H_j^{0}\bigr)
      -
      \operatorname{sim}\!\bigl(H_k^{i}, H_j^{i}\bigr)
  \Bigr|.
\end{equation}
\end{definition}

Layer-wise positional-topology distortion measures the average absolute error between the pairwise token distances in the original (when $i=0$) and the retained positional topology in the layer $i$; hence $\Delta_{\mathcal{G}}^{i}=0$ if the full model preserves all pairwise positional structures of the input space. Layer-wise semantic topology distortion quantifies how much the pairwise semantic relationships observed in the input space (when $i=0$) are distorted after propagating to layer $i$. In particular, larger values of $\Delta_{\mathcal{S}}^{i}$ indicate greater degradation of the original semantic topology.

\begin{theorem}
\label{theorem:2}
Under the condition of Theorem \ref{theorem:1}, the upper bound of the Rademacher complexity of $l$-th layer of the Transformer is:
\begin{equation}
\label{eq_sgb}
\begin{aligned}
\mathfrak R(\mathcal{H}_s^l)
  \le
  \mathfrak R(\mathcal{H}_s^0)\;
  \prod_{i=1}^{l} L_i\sqrt{1-\rho_i}=
  \mathfrak R(\mathcal{H}_s^0)\;\exp\!\Bigl\{
      \tfrac12\sum_{i=1}^{l}\!\bigl[\ln L_i + \ln(1-\rho_i)\bigr]\Bigr\}
\end{aligned}
\end{equation}
where $\rho_i$ is the extent to which the original topological structure among input tokens is preserved at the $i$-th layer, i.e., $\rho_i=1-\bigl(\alpha\,\Delta_{\mathcal{G}}^{i}+\beta\,\Delta_{\mathcal{S}}^{i}\bigr)\in[0,1]$, $\alpha ,\beta  \in {\mathbb{R}^ + }$, and $L_i$ is the Lipschitz constant of the function class corresponding to the $i$-th Transformer layer.
\end{theorem}

This theorem shows that preserving the original topological structure at each transformer layer can effectively reduce the Rademacher complexity upper bound, thereby improving its generalization ability.
Specifically, it introduces a layer-wise topology preservation score $\rho_i$, which quantifies how well the original positional and semantic structure is maintained. When $\rho_i$ approaches 1, indicating better structural preservation, the corresponding term $\sqrt{1 - \rho_i}$ becomes smaller, reducing the layer’s contribution to overall model complexity. Recursively applying this effect across layers yields a global complexity bound determined by all $\rho_i$. Since Rademacher complexity directly influences the generalization error bound (Theorem \ref{theorem:1}), better structural preservation leads to stronger generalization. The detailed proof is provided in Appendix \ref{sec_appLproof_2}.

Based on the above results, we can conclude that an appropriate preservation of input token topology across layers leads to a tighter generalization bound. These results explain the empirical observations in Section \ref{motivation}, i.e., the benefits of enhancing structure preservation. Considering the differences in topology information perception at different layers of the network and the values of $\alpha$ and $\beta$ vary across different layers, finding an effective way to adaptively preserve the original topological structure among input tokens is important for boosting model performance.

\begin{figure}
    \centering
    \includegraphics[width=0.6\textwidth]{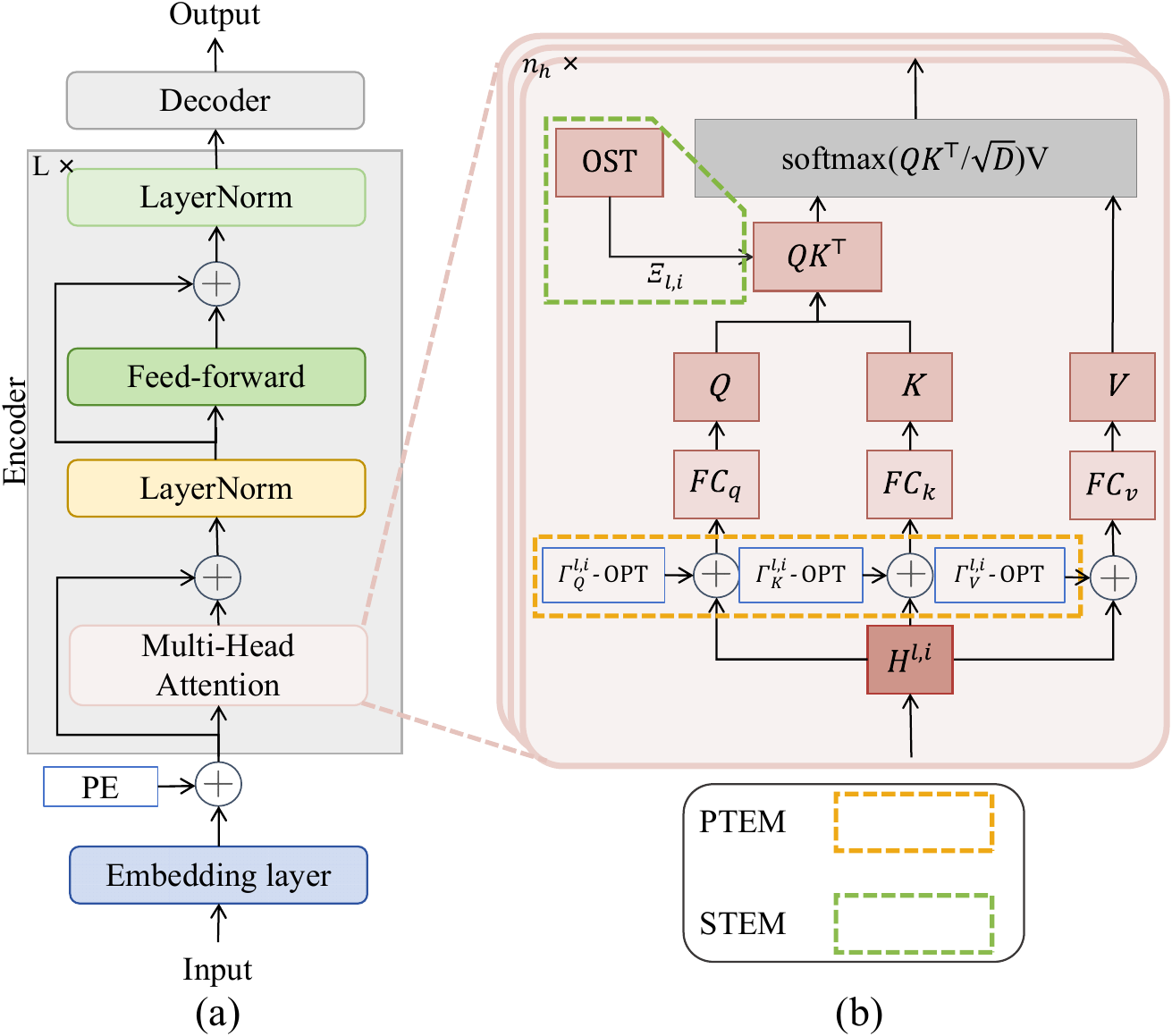}
    \caption{Overview of the Topology Enhancement Method (TEM). (a) Architecture diagram of the current popular Transformer-based TSF methods. (b) Illustration of the integration locations of PTEM and STEM within Transformer-based TSF methods.}
    \label{fig:1}
\end{figure}

\section{Method}

Based on the above analysis, we propose a plug-and-play Topology Enhancement Method (TEM). It aims to adaptively strengthen the preservation of OPT and OST at each Transformer layer to improve generalization performance (demonstrated in Section \ref{SEC_TA}). TEM consists of two components: 1) Positional Topology Enhancement Module (PTEM), which reinforces the preservation of OPT by injecting weighted PE into each Transformer layer (Section \ref{SEC_GNEM}); 2) Semantic Topology Enhancement Module (STEM), which enhances the preservation of OST by adding a weighted similarity matrix of input tokens to each attention map (Section \ref{SEC_SNEM}). To ensure accurate topology injection, we introduce a bi-level optimization process to optimize the proposed TEM (Section \ref{SEC_Loss}). The framework of TEM is shown in Figure \ref{fig:1}, and the corresponding pseudo-code is provided in Section \ref{Pseudocode}.

\subsection{Positional Topology Enhancement Module}
\label{SEC_GNEM}
In this subsection, we introduce the details of the PTEM, i.e., how it injects OPT into each layer of the model.

Let $H^l$ denote the input of the $(l+1)$-th Transformer layer. Let the number of heads be $n_h$. $H^l$ is evenly split into $n_h$ parts, which are used as the inputs for the $n_h$ heads, denoted as $H^{l,i}$ $(i = 1, \dots, n_h)$. Generally, for the $i$-th attention head in the $l$-th layer, $H^{l,i}$ is projected into the Query, Key, and Value matrices through three distinct fully connected (FC) layers. The core idea of PTEM is to inject the OPT into this process. Specifically, before feeding $H^{l,i}$ into the FC layers, PTEM first generates three learnable scalar coefficients, denoted as $\Gamma_Q^{l,i}$, $\Gamma_K^{l,i}$, and $\Gamma_V^{l,i}$, corresponding to the Query, Key, and Value paths, respectively. Then, let $[\cdot]$ be the matrix operation, the modified input for each path is computed as:
\begin{equation}
\label{eq_GNEM}
    H_{j}^{l,i} = H^{l,i} + \Gamma_j^{l,i} \times [\mathcal{G}], \Gamma_j^{l,i} \in \mathbb{R}^+, j \in \{Q,K,V\}.
\end{equation}
 
Finally, the adjusted representations $H_Q^{l,i}$, $H_K^{l,i}$, and $H_V^{l,i}$ are passed through their respective FC layers in the $i$-th attention head of the $l$-th Transformer layer to generate the final Query, Key, and Value matrices. We represent the tensor $\Gamma$ using the index notation $\Gamma = [l, i, j]$, where $l$, $i$, and $j$ denote the Transformer layer index, attention head index, and projection type (Query, Key, or Value), respectively, and $\Gamma[l, i, j]=\Gamma_j^{l,i}$. This tensor can be interpreted as controlling the proportion of the OPT injected into each Transformer layer. In other words, by adjusting the values of elements in $\Gamma$, we can explicitly modulate the extent to which OPT is preserved in each layer of the Transformer. More details are provided in Appendix \ref{method_appe}.

\subsection{Semantic Topology Enhancement Module}
\label{SEC_SNEM}

In this subsection, we introduce the details of the STEM, i.e., how it injects OST into each layer of the model.

In the $i$-th attention head of the $l$-th Transformer layer, a key computational step involves calculating the dot product between the Query and Key matrices, i.e., $QK^{\top}$. Typically, each element in $QK^{\top}$ can be interpreted as the similarity between a pair of tokens, which aligns closely with the definition of the OST. Based on this observation, STEM enhances OST preservation within Transformer layers by injecting OST into $QK^{\top}$ directly, as follows:
\begin{equation}
\label{eq_SNEM}
QK^{\top}= QK^{\top} + \Xi_{l,i} \times [\mathcal{S}^0], \quad \Xi_{l,i} \in \mathbb{R}^+,
\end{equation}
where the $sim(\cdot, \cdot)$ in $\mathcal{S}^0$ is computed directly using the inner product between vectors (i.e., as in $QK^{\top}$), $\Xi_{l,i}$ denotes the injection strength of OST into the $i$-th attention head of the $l$-th Transformer layer. We represent the full OST injection matrix as $\Xi = [l, i]$, where each element $\Xi[l, i] = \Xi_{l,i}$ controls the proportion of OST injected at a specific layer and head. In this formulation, adjusting the values in $\Xi$ enables fine-grained control over the extent to which OST is preserved throughout the network. More details are provided in Appendix \ref{method_appe}.

\subsection{Overall Optimization}
\label{SEC_Loss}

To ensure the accuracy of OPT and OST injection, we propose a bi-level optimization mechanism that jointly optimizes the TSF model ($f_\theta$) and the parameters of PTEM and STEM modules ($\Gamma, \Xi$).

Specifically, during training, a batch (with $N_b$ samples) is denoted as $\mathcal{D}_b=\{X_i,Y_i\}_{i=1}^{N_b}$. The whole objective of TEM is presented as:
\begin{equation} \label{eq_grad_update1}
\mathop {\min }\limits_{\Gamma ,\Xi ,{{f}_\theta }} {\mathcal{L}_{\text{mse}}}({{f}_\theta },\Gamma ,\Xi ,\mathcal{D}_b),
\;\; {\rm{s.t.}} \; \mathop {\min }\limits_{{{f}_\theta }} {\mathcal{L}_{\text{mse}}}({f_\theta },\Gamma ,\Xi ,\mathcal{D}_b),
\end{equation}
where $\mathcal{L}_{\text{mse}}$ represents the loss that used in existing TSF methods, e.g., the MSE loss.

Since Eq.(\ref{eq_grad_update1}) formulates a multi-objective optimization problem, we adopt a bi-level optimization strategy to solve it. Specifically, for the inner-loop optimization, we fix $\Gamma$ and $\Xi$, and update the model parameters $f_\theta$ using gradient descent:
\begin{equation}
\label{eq_grad_update2}
{f_\theta } \leftarrow {f_\theta } - {\eta _1}{\nabla _{{f_\theta }}}{\mathcal{L}_{\text{mse}}}({f_\theta },\Gamma ,\Xi ,\mathcal{D}_b),
\end{equation}
where $\eta_1$ is the learning rate for the inner loop.

For the outer-loop optimization, we proceed based on the updated model from the inner loop. Let $f_\theta^1$ denote the model parameters obtained after the inner-loop update. We then update $\Gamma$ and $\Xi$ via gradient descent as follows:
\begin{equation}
\label{eq_grad_update3}
\{ \Gamma ,\Xi \}   \leftarrow \{ \Gamma ,\Xi \}   - {\eta _2}{\nabla _{\{ \Gamma ,\Xi \}} }{\mathcal{L}_{\text{mse}}}(f_\theta ^1,\Gamma ,\Xi ,\mathcal{D}_b)
\end{equation}
where $\eta_2$ is the learning rate for the outer loop. It is important to note that the loss term $\mathcal{L}_{\text{mse}}(f_\theta^1, \Gamma, \Xi, \mathcal{D}_b)$ is evaluated using the parameters $f_\theta^1$ obtained from the inner-loop optimization. Since $f_\theta^1$ itself is a function of $\Gamma$ and $\Xi$ (as shown in Eq.(\ref{eq_grad_update2})), the gradient $\nabla_{\{ \Gamma ,\Xi \} } \mathcal{L}_{\text{mse}}(f_\theta^1, \Gamma, \Xi, \mathcal{D}_b)$ involves a composite gradient, e.g., $\nabla_{\{ \Gamma ,\Xi \} } \mathcal{L}_{\text{mse}}(f_\theta^1, \Gamma, \Xi, \mathcal{D}_b) = \nabla_{f_\theta} \mathcal{L}_{\text{mse}}(f_\theta^1, \Gamma, \Xi, \mathcal{D}_b) \cdot \nabla_{\{ \Gamma ,\Xi \} } f_\theta^1$.

\textbf{Explanation of the Learning Strategy of $\Gamma$ and $\Xi$}: It is important to emphasize that the learning of $\Gamma$ and $\Xi$ is not performed jointly with the learning of $f_\theta$. Instead, we adopt a bi-level learning strategy in which $\Gamma$ and $\Xi$ are updated after the optimization of $f_\theta$. The rationale behind this design is to accurately model the effect of preserving the original topological structure of the input space in intermediate layers on improving TSF performance. If we were to optimize $\Gamma$, $\Xi$, and $f_\theta$ simultaneously, e.g., by minimizing a joint loss $\min_{\Gamma, \Xi, f_\theta} \mathcal{L}_{\text{mse}}(f_\theta, \Gamma, \Xi, \mathcal{D}_b)$, it would be difficult to determine whether the improvement in the loss is actually due to the updates in $\Gamma$ and $\Xi$, since the loss can already decrease by optimizing $f_\theta$ alone. In contrast, our bi-level strategy (as described in Eq.(\ref{eq_grad_update1})) allows us to isolate the contribution of $\Gamma$ and $\Xi$. That is, after minimizing the loss with respect to $f_\theta$, we further update $\Gamma$ and $\Xi$ to reduce the loss again. This learning process directly reflects the principle that appropriately preserving the topological structure of the original input space in intermediate layers can further enhance TSF performance.

\begin{table*}
\caption{Full \textbf{MSE} results for the TSF task. All baseline models' input length is 96, and the prediction lengths include $\{96,192,336,720\}$. The results of our method are averaged over five random seeds, and the standard deviation across the five runs is reported after the ``$\pm$'' symbol.}
\renewcommand{\arraystretch}{1}
\centering
\resizebox{1\columnwidth}{!}{%
\begin{threeparttable}
\begin{small}
\begin{tabular}{c|c|c|c|c|c|c|c|c|c|c|c|c|c|c}
\toprule
\multicolumn{2}{c}{\multirow{2}{*}{Models}} &
\textbf{PatchTST+TEM} & PatchTST &
\textbf{iTransformer+TEM} & iTransformer &
\textbf{Transformer+TEM} & Transformer &
Crossformer & TimesNet & DLinear & DSformer &
SpareTSF & TIDE & PRReg CI \\[2pt]
\multicolumn{2}{c}{} &
(Ours) & (2023) & (Ours) & (2024) & (Ours) & (2017) &
(2023) & (2023) & (2023) & (2023) & (2024) & (2023) & (2024) \\[2pt]
\multicolumn{2}{c}{Metric} & MSE & MSE & MSE & MSE & MSE & MSE & MSE & MSE & MSE & MSE & MSE & MSE & MSE \\
\toprule
\multirow{5}{*}{\rotatebox{90}{ETTh1}}
 & 96  & 0.404$\pm$0.002 & 0.414 & \textcolor{blue}{0.380}$\pm$0.001 & 0.386 & 0.550$\pm$0.002 & 0.611 & 0.423 & 0.384 & 0.386 & 0.401 & \textcolor{red}{0.374} & 0.479 & 0.409 \\
 & 192 & 0.451$\pm$0.000 & 0.460 & \textcolor{blue}{0.435}$\pm$0.002 & 0.441 & 0.627$\pm$0.000 & 0.697 & 0.471 & 0.436 & 0.437 & 0.441 & \textcolor{red}{0.419} & 0.525 & 0.454 \\
 & 336 & 0.490$\pm$0.001 & 0.501 & \textcolor{blue}{0.478}$\pm$0.000 & 0.487 & 0.693$\pm$0.001 & 0.770 & 0.570 & 0.491 & 0.481 & 0.486 & \textcolor{red}{0.463} & 0.565 & 0.514 \\
 & 720 & \textcolor{blue}{0.488}$\pm$0.001 & 0.500 & 0.495$\pm$0.000 & 0.503 & 0.716$\pm$0.000 & 0.796 & 0.653 & 0.521 & 0.519 & 0.493 & \textcolor{red}{0.477} & 0.594 & 0.530 \\
\cmidrule(lr){2-15}
 & Avg & 0.458 & 0.469 & \textcolor{blue}{0.447} & 0.454 & 0.646 & 0.718 & 0.529 & 0.458 & 0.456 & 0.455 & \textcolor{red}{0.433} & 0.541 & 0.477 \\
\midrule
\multirow{5}{*}{\rotatebox{90}{ETTh2}}
 & 96  & \textcolor{blue}{0.295}$\pm$0.000 & 0.302 & \textcolor{red}{0.291}$\pm$0.002 & 0.297 & 0.423$\pm$0.001 & 0.470 & 0.745 & 0.340 & 0.333 & 0.319 & 0.308 & 0.400 & 0.300 \\
 & 192 & \textcolor{blue}{0.379}$\pm$0.002 & 0.388 & \textcolor{red}{0.376}$\pm$0.001 & 0.380 & 0.541$\pm$0.000 & 0.601 & 0.877 & 0.402 & 0.477 & 0.400 & 0.388 & 0.528 & 0.385 \\
 & 336 & \textcolor{red}{0.419}$\pm$0.001 & \textcolor{blue}{0.426} & \textcolor{red}{0.419}$\pm$0.002 & 0.428 & 0.608$\pm$0.002 & 0.676 & 1.043 & 0.452 & 0.594 & 0.462 & 0.431 & 0.643 & 0.440 \\
 & 720 & \textcolor{blue}{0.424}$\pm$0.003 & 0.431 & \textcolor{red}{0.422}$\pm$0.000 & 0.427 & 0.608$\pm$0.001 & 0.676 & 1.104 & 0.462 & 0.831 & 0.457 & 0.432 & 0.874 & 0.426 \\
\cmidrule(lr){2-15}
 & Avg & \textcolor{blue}{0.379} & 0.387 & \textcolor{red}{0.377} & 0.383 & 0.545 & 0.605 & 0.942 & 0.414 & 0.559 & 0.410 & 0.390 & 0.611 & 0.388 \\
\midrule
\multirow{5}{*}{\rotatebox{90}{ETTm1}}
 & 96  & \textcolor{blue}{0.323}$\pm$0.001 & 0.329 & \textcolor{red}{0.321}$\pm$0.002 & 0.334 & 0.475$\pm$0.003 & 0.528 & 0.404 & 0.338 & 0.345 & 0.342 & 0.349 & 0.364 & 0.344 \\
 & 192 & \textcolor{red}{0.361}$\pm$0.000 & \textcolor{blue}{0.367} & 0.369$\pm$0.000 & 0.377 & 0.536$\pm$0.001 & 0.596 & 0.450 & 0.374 & 0.380 & 0.385 & 0.385 & 0.398 & 0.384 \\
 & 336 & \textcolor{red}{0.390}$\pm$0.002 & \textcolor{blue}{0.399} & 0.413$\pm$0.001 & 0.426 & 0.607$\pm$0.000 & 0.674 & 0.532 & 0.410 & 0.413 & 0.402 & 0.412 & 0.428 & 0.442 \\
 & 720 & \textcolor{red}{0.444}$\pm$0.000 & \textcolor{blue}{0.454} & 0.470$\pm$0.001 & 0.491 & 0.699$\pm$0.002 & 0.777 & 0.666 & 0.478 & 0.474 & 0.472 & 0.474 & 0.487 & 0.479 \\
\cmidrule(lr){2-15}
 & Avg & \textcolor{red}{0.380} & \textcolor{blue}{0.387} & 0.393 & 0.407 & 0.579 & 0.643 & 0.513 & 0.400 & 0.403 & 0.400 & 0.405 & 0.419 & 0.412 \\
\midrule
\multirow{5}{*}{\rotatebox{90}{ETTm2}}
 & 96  & \textcolor{red}{0.168}$\pm$0.000 & 0.175 & \textcolor{blue}{0.174}$\pm$0.001 & 0.180 & 0.257$\pm$0.002 & 0.285 & 0.287 & 0.187 & 0.193 & 0.184 & 0.180 & 0.207 & 0.187 \\
 & 192 & \textcolor{red}{0.233}$\pm$0.001 & \textcolor{blue}{0.241} & 0.244$\pm$0.002 & 0.250 & 0.356$\pm$0.003 & 0.396 & 0.414 & 0.249 & 0.284 & 0.248 & 0.243 & 0.290 & 0.253 \\
 & 336 & \textcolor{red}{0.296}$\pm$0.002 & 0.305 & 0.305$\pm$0.003 & 0.311 & 0.443$\pm$0.000 & 0.492 & 0.597 & 0.321 & 0.369 & 0.335 & \textcolor{blue}{0.302} & 0.377 & 0.320 \\
 & 720 & \textcolor{red}{0.390}$\pm$0.001 & 0.402 & 0.404$\pm$0.000 & 0.412 & 0.587$\pm$0.001 & 0.652 & 1.730 & 0.408 & 0.554 & 0.412 & \textcolor{blue}{0.398} & 0.558 & 0.440 \\
\cmidrule(lr){2-15}
 & Avg & \textcolor{red}{0.272} & \textcolor{blue}{0.281} & 0.282 & 0.288 & 0.410 & 0.456 & 0.757 & 0.291 & 0.350 & 0.295 & \textcolor{blue}{0.281} & 0.358 & 0.300 \\
\midrule
\multirow{5}{*}{\rotatebox{90}{ECL}}
 & 96  & 0.186$\pm$0.001 & 0.195 & \textcolor{red}{0.141}$\pm$0.000 & 0.148 & 0.234$\pm$0.002 & 0.260 & 0.219 & 0.168 & 0.197 & 0.173 & 0.194 & 0.237 & \textcolor{blue}{0.146} \\
 & 192 & 0.192$\pm$0.001 & 0.199 & \textcolor{red}{0.156}$\pm$0.002 & 0.162 & 0.239$\pm$0.001 & 0.266 & 0.231 & 0.184 & 0.196 & 0.183 & 0.197 & 0.236 & \textcolor{blue}{0.160} \\
 & 336 & 0.203$\pm$0.002 & 0.215 & \textcolor{red}{0.169}$\pm$0.001 & 0.178 & 0.252$\pm$0.000 & 0.280 & 0.246 & 0.198 & 0.209 & 0.203 & 0.209 & 0.249 & \textcolor{blue}{0.172} \\
 & 720 & 0.245$\pm$0.000 & 0.256 & \textcolor{red}{0.209}$\pm$0.003 & 0.225 & 0.272$\pm$0.001 & 0.302 & 0.280 & \textcolor{blue}{0.220} & 0.245 & 0.259 & 0.251 & 0.284 & \textcolor{red}{0.209} \\
\cmidrule(lr){2-15}
 & Avg & 0.207 & 0.216 & \textcolor{red}{0.169} & 0.178 & 0.249 & 0.277 & 0.244 & 0.192 & 0.212 & 0.205 & 0.213 & 0.251 & \textcolor{blue}{0.172} \\
\midrule
\multirow{5}{*}{\rotatebox{90}{Traffic}}
 & 96  & 0.524$\pm$0.002 & 0.544 & \textcolor{red}{0.381}$\pm$0.000 & \textcolor{blue}{0.395} & 0.582$\pm$0.001 & 0.647 & 0.522 & 0.593 & 0.650 & 0.529 & 0.543 & 0.805 & 0.403 \\
 & 192 & 0.537$\pm$0.001 & 0.540 & \textcolor{red}{0.412}$\pm$0.002 & 0.417 & 0.584$\pm$0.004 & 0.649 & 0.530 & 0.617 & 0.598 & 0.533 & 0.555 & 0.756 & \textcolor{blue}{0.416} \\
 & 336 & 0.543$\pm$0.000 & 0.551 & \textcolor{red}{0.429}$\pm$0.001 & \textcolor{blue}{0.433} & 0.600$\pm$0.002 & 0.667 & 0.558 & 0.629 & 0.605 & 0.545 & 0.565 & 0.762 & 0.448 \\
 & 720 & 0.571$\pm$0.001 & 0.586 & \textcolor{red}{0.462}$\pm$0.000 & \textcolor{blue}{0.467} & 0.627$\pm$0.001 & 0.697 & 0.589 & 0.640 & 0.645 & 0.583 & 0.639 & 0.719 & 0.471 \\
\cmidrule(lr){2-15}
 & Avg & 0.544 & 0.555 & \textcolor{red}{0.421} & \textcolor{blue}{0.428} & 0.599 & 0.665 & 0.550 & 0.620 & 0.625 & 0.548 & 0.576 & 0.760 & 0.435 \\
\midrule
\multirow{5}{*}{\rotatebox{90}{Weather}}
 & 96  & 0.171$\pm$0.000 & 0.177 & 0.165$\pm$0.002 & 0.174 & 0.356$\pm$0.002 & 0.395 & \textcolor{red}{0.158} & 0.172 & 0.196 & \textcolor{blue}{0.162} & 0.180 & 0.202 & 0.171 \\
 & 192 & \textcolor{blue}{0.209}$\pm$0.002 & 0.225 & 0.210$\pm$0.003 & 0.221 & 0.557$\pm$0.001 & 0.619 & \textcolor{red}{0.206} & 0.219 & 0.237 & 0.211 & 0.226 & 0.242 & 0.219 \\
 & 336 & 0.273$\pm$0.001 & 0.278 & \textcolor{red}{0.266}$\pm$0.001 & 0.278 & 0.620$\pm$0.002 & 0.689 & 0.272 & 0.280 & 0.283 & \textcolor{blue}{0.267} & 0.281 & 0.287 & 0.286 \\
 & 720 & \textcolor{red}{0.342}$\pm$0.001 & 0.354 & 0.350$\pm$0.000 & 0.358 & 0.833$\pm$0.000 & 0.926 & 0.398 & 0.365 & 0.345 & \textcolor{blue}{0.343} & 0.358 & 0.351 & 0.357 \\
\cmidrule(lr){2-15}
 & Avg & 0.249 & 0.259 & \textcolor{blue}{0.248} & 0.258 & 0.591 & 0.657 & 0.259 & 0.259 & 0.265 & \textcolor{red}{0.246} & 0.261 & 0.271 & 0.258 \\
\midrule
\multirow{5}{*}{\rotatebox{90}{Solar-Energy}}
 & 96  & 0.227$\pm$0.000 & 0.234 & \textcolor{red}{0.197}$\pm$0.001 & \textcolor{blue}{0.203} & 0.347$\pm$0.002 & 0.386 & 0.310 & 0.250 & 0.290 & 0.247 & 0.245 & 0.312 & 0.204 \\
 & 192 & 0.257$\pm$0.001 & 0.267 & \textcolor{red}{0.227}$\pm$0.002 & \textcolor{blue}{0.233} & 0.400$\pm$0.001 & 0.444 & 0.734 & 0.296 & 0.320 & 0.288 & 0.255 & 0.339 & 0.249 \\
 & 336 & 0.282$\pm$0.002 & 0.290 & \textcolor{red}{0.242}$\pm$0.001 & \textcolor{blue}{0.248} & 0.424$\pm$0.001 & 0.471 & 0.750 & 0.319 & 0.353 & 0.329 & 0.259 & 0.368 & 0.260 \\
 & 720 & 0.280$\pm$0.004 & 0.289 & \textcolor{red}{0.243}$\pm$0.000 & \textcolor{blue}{0.249} & 0.426$\pm$0.000 & 0.473 & 0.769 & 0.338 & 0.356 & 0.341 & 0.262 & 0.370 & 0.271 \\
\cmidrule(lr){2-15}
 & Avg & 0.262 & 0.270 & \textcolor{red}{0.227} & \textcolor{blue}{0.233} & 0.398 & 0.442 & 0.641 & 0.301 & 0.330 & 0.301 & 0.255 & 0.347 & 0.246 \\
\bottomrule
\end{tabular}
\end{small}
\end{threeparttable}}%
\label{tab:mse_std}
\end{table*}

\begin{table*}
\caption{Full \textbf{MAE} results for the TSF task. All baseline models' input length is 96, and the prediction lengths include $\{96,192,336,720\}$. The results of our method are averaged over five random seeds, and the standard deviation across the five runs is reported after the ``$\pm$'' symbol.}
\renewcommand{\arraystretch}{1}
\centering
\resizebox{1\columnwidth}{!}{%
\begin{threeparttable}
\begin{small}
\begin{tabular}{c|c|c|c|c|c|c|c|c|c|c|c|c|c|c}
\toprule
\multicolumn{2}{c}{\multirow{2}{*}{Models}} &
\textbf{PatchTST+TEM} & PatchTST &
\textbf{iTransformer+TEM} & iTransformer &
\textbf{Transformer+TEM} & Transformer &
Crossformer & TimesNet & DLinear & DSformer &
SpareTSF & TIDE & PRReg CI \\[2pt]
\multicolumn{2}{c}{} &
(Ours) & (2023) & (Ours) & (2024) & (Ours) & (2017) &
(2023) & (2023) & (2023) & (2023) & (2024) & (2023) & (2024) \\[2pt]
\multicolumn{2}{c}{Metric} & MAE & MAE & MAE & MAE & MAE & MAE & MAE & MAE & MAE & MAE & MAE & MAE & MAE \\
\toprule
\multirow{5}{*}{\rotatebox{90}{ETTh1}}
 & 96  & 0.410$\pm$0.000 & 0.419 & \textcolor{blue}{0.400}$\pm$0.001 & 0.405 & 0.485$\pm$0.001 & 0.539 & 0.448 & 0.402 & \textcolor{blue}{0.400} & 0.412 & \textcolor{red}{0.390} & 0.464 & 0.408 \\
 & 192 & 0.436$\pm$0.001 & 0.445 & 0.431$\pm$0.001 & 0.436 & 0.522$\pm$0.001 & 0.580 & 0.474 & \textcolor{blue}{0.429} & 0.432 & 0.435 & \textcolor{red}{0.420} & 0.492 & 0.447 \\
 & 336 & 0.456$\pm$0.002 & 0.466 & \textcolor{blue}{0.450}$\pm$0.001 & 0.458 & 0.548$\pm$0.002 & 0.609 & 0.546 & 0.469 & 0.459 & 0.456 & \textcolor{red}{0.440} & 0.515 & 0.483 \\
 & 720 & 0.477$\pm$0.003 & 0.488 & 0.482$\pm$0.000 & 0.491 & 0.588$\pm$0.001 & 0.653 & 0.621 & 0.500 & 0.516 & \textcolor{blue}{0.475} & \textcolor{red}{0.472} & 0.558 & 0.490 \\
\cmidrule(lr){2-15}
 & Avg & 0.445 & 0.454 & \textcolor{blue}{0.441} & 0.447 & 0.535 & 0.594 & 0.522 & 0.450 & 0.452 & 0.446 & \textcolor{red}{0.431} & 0.507 & 0.457 \\
\midrule
\multirow{5}{*}{\rotatebox{90}{ETTh2}}
 & 96  & \textcolor{red}{0.344}$\pm$0.001 & 0.348 & \textcolor{red}{0.344}$\pm$0.001 & \textcolor{blue}{0.347} & 0.415$\pm$0.001 & 0.461 & 0.584 & 0.374 & 0.387 & 0.359 & 0.354 & 0.440 & 0.363 \\
 & 192 & \textcolor{red}{0.395}$\pm$0.002 & 0.400 & \textcolor{blue}{0.397}$\pm$0.002 & 0.400 & 0.479$\pm$0.001 & 0.532 & 0.656 & 0.414 & 0.476 & 0.410 & 0.398 & 0.509 & 0.414 \\
 & 336 & \textcolor{red}{0.419}$\pm$0.001 & 0.433 & \textcolor{blue}{0.427}$\pm$0.001 & 0.432 & 0.517$\pm$0.002 & 0.575 & 0.731 & 0.452 & 0.541 & 0.451 & 0.443 & 0.571 & 0.452 \\
 & 720 & \textcolor{blue}{0.440}$\pm$0.003 & 0.446 & \textcolor{red}{0.435}$\pm$0.001 & 0.445 & 0.533$\pm$0.000 & 0.592 & 0.763 & 0.468 & 0.657 & 0.463 & 0.447 & 0.679 & 0.466 \\
\cmidrule(lr){2-15}
 & Avg & \textcolor{red}{0.400} & 0.407 & \textcolor{blue}{0.401} & 0.407 & 0.487 & 0.541 & 0.684 & 0.427 & 0.515 & 0.421 & 0.411 & 0.550 & 0.424 \\
\midrule
\multirow{5}{*}{\rotatebox{90}{ETTm1}}
 & 96  & \textcolor{blue}{0.361}$\pm$0.001 & 0.367 & \textcolor{red}{0.359}$\pm$0.001 & 0.368 & 0.440$\pm$0.001 & 0.489 & 0.426 & 0.375 & 0.372 & 0.375 & 0.381 & 0.387 & 0.384 \\
 & 192 & \textcolor{red}{0.379}$\pm$0.001 & \textcolor{blue}{0.385} & 0.387$\pm$0.001 & 0.391 & 0.468$\pm$0.001 & 0.520 & 0.451 & 0.387 & 0.389 & 0.392 & 0.397 & 0.404 & 0.399 \\
 & 336 & \textcolor{red}{0.402}$\pm$0.002 & \textcolor{blue}{0.410} & \textcolor{blue}{0.410}$\pm$0.002 & 0.420 & 0.503$\pm$0.002 & 0.559 & 0.515 & 0.411 & 0.413 & 0.411 & 0.417 & 0.425 & 0.426 \\
 & 720 & \textcolor{red}{0.429}$\pm$0.002 & \textcolor{blue}{0.439} & 0.447$\pm$0.001 & 0.459 & 0.549$\pm$0.003 & 0.610 & 0.589 & 0.450 & 0.453 & 0.448 & 0.450 & 0.461 & 0.455 \\
\cmidrule(lr){2-15}
 & Avg & \textcolor{red}{0.393} & \textcolor{blue}{0.400} & 0.401 & 0.410 & 0.491 & 0.545 & 0.496 & 0.406 & 0.407 & 0.407 & 0.411 & 0.419 & 0.416 \\
\midrule
\multirow{5}{*}{\rotatebox{90}{ETTm2}}
 & 96  & \textcolor{red}{0.253}$\pm$0.001 & 0.259 & \textcolor{blue}{0.258}$\pm$0.000 & 0.264 & 0.316$\pm$0.001 & 0.351 & 0.366 & 0.267 & 0.292 & 0.269 & 0.259 & 0.305 & 0.278 \\
 & 192 & \textcolor{red}{0.296}$\pm$0.000 & \textcolor{blue}{0.302} & 0.303$\pm$0.001 & 0.309 & 0.370$\pm$0.002 & 0.411 & 0.492 & 0.309 & 0.362 & 0.306 & 0.308 & 0.364 & 0.329 \\
 & 336 & \textcolor{red}{0.335}$\pm$0.002 & 0.343 & \textcolor{blue}{0.341}$\pm$0.001 & 0.348 & 0.417$\pm$0.002 & 0.463 & 0.542 & 0.351 & 0.427 & 0.365 & 0.342 & 0.422 & 0.366 \\
 & 720 & \textcolor{red}{0.392}$\pm$0.003 & 0.400 & 0.400$\pm$0.002 & 0.407 & 0.487$\pm$0.003 & 0.541 & 1.042 & 0.403 & 0.522 & 0.409 & \textcolor{blue}{0.398} & 0.524 & 0.428 \\
\cmidrule(lr){2-15}
 & Avg & \textcolor{red}{0.319} & \textcolor{blue}{0.326} & \textcolor{blue}{0.326} & 0.332 & 0.398 & 0.442 & 0.610 & 0.333 & 0.401 & 0.337 & 0.327 & 0.404 & 0.350 \\
\midrule
\multirow{5}{*}{\rotatebox{90}{ECL}}
 & 96  & 0.281$\pm$0.001 & 0.285 & \textcolor{red}{0.235}$\pm$0.001 & 0.240 & 0.322$\pm$0.001 & 0.358 & 0.314 & 0.272 & 0.282 & 0.269 & 0.269 & 0.329 & \textcolor{blue}{0.239} \\
 & 192 & 0.283$\pm$0.001 & 0.289 & \textcolor{red}{0.252}$\pm$0.000 & \textcolor{blue}{0.253} & 0.330$\pm$0.000 & 0.367 & 0.322 & 0.289 & 0.285 & 0.280 & 0.273 & 0.330 & 0.262 \\
 & 336 & 0.291$\pm$0.002 & 0.305 & \textcolor{red}{0.264}$\pm$0.001 & \textcolor{blue}{0.269} & 0.338$\pm$0.002 & 0.375 & 0.337 & 0.300 & 0.301 & 0.297 & 0.287 & 0.344 & 0.276 \\
 & 720 & 0.318$\pm$0.003 & 0.337 & \textcolor{red}{0.305}$\pm$0.002 & 0.317 & 0.348$\pm$0.002 & 0.386 & 0.363 & 0.320 & 0.333 & 0.340 & 0.321 & 0.373 & \textcolor{blue}{0.306} \\
\cmidrule(lr){2-15}
 & Avg & 0.293 & 0.304 & \textcolor{red}{0.264} & \textcolor{blue}{0.270} & 0.335 & 0.372 & 0.334 & 0.295 & 0.300 & 0.297 & 0.288 & 0.344 & 0.271 \\
\midrule
\multirow{5}{*}{\rotatebox{90}{Traffic}}
 & 96  & 0.351$\pm$0.001 & 0.359 & \textcolor{red}{0.263}$\pm$0.001 & \textcolor{blue}{0.268} & 0.321$\pm$0.001 & 0.357 & 0.290 & 0.321 & 0.396 & 0.370 & 0.332 & 0.493 & 0.279 \\
 & 192 & 0.353$\pm$0.001 & 0.354 & \textcolor{red}{0.273}$\pm$0.001 & \textcolor{blue}{0.276} & 0.320$\pm$0.001 & 0.356 & 0.293 & 0.336 & 0.370 & 0.366 & 0.356 & 0.474 & 0.301 \\
 & 336 & 0.355$\pm$0.001 & 0.358 & \textcolor{red}{0.281}$\pm$0.002 & \textcolor{blue}{0.283} & 0.328$\pm$0.002 & 0.364 & 0.305 & 0.336 & 0.373 & 0.370 & 0.360 & 0.477 & 0.305 \\
 & 720 & 0.361$\pm$0.002 & 0.375 & \textcolor{red}{0.300}$\pm$0.001 & \textcolor{blue}{0.302} & 0.338$\pm$0.003 & 0.376 & 0.328 & 0.350 & 0.394 & 0.386 & 0.352 & 0.449 & 0.322 \\
\cmidrule(lr){2-15}
 & Avg & 0.355 & 0.362 & \textcolor{red}{0.279} & \textcolor{blue}{0.282} & 0.327 & 0.363 & 0.304 & 0.336 & 0.383 & 0.373 & 0.350 & 0.473 & 0.302 \\
\midrule
\multirow{5}{*}{\rotatebox{90}{Weather}}
 & 96  & \textcolor{blue}{0.208}$\pm$0.001 & 0.218 & \textcolor{blue}{0.208}$\pm$0.000 & 0.214 & 0.384$\pm$0.002 & 0.427 & 0.230 & 0.220 & 0.255 & \textcolor{red}{0.207} & 0.230 & 0.261 & 0.217 \\
 & 192 & 0.255$\pm$0.001 & 0.259 & \textcolor{red}{0.252}$\pm$0.001 & \textcolor{blue}{0.254} & 0.504$\pm$0.002 & 0.560 & 0.277 & 0.261 & 0.296 & \textcolor{red}{0.252} & 0.271 & 0.298 & 0.270 \\
 & 336 & \textcolor{blue}{0.294}$\pm$0.002 & 0.297 & \textcolor{red}{0.293}$\pm$0.002 & 0.296 & 0.535$\pm$0.002 & 0.594 & 0.335 & 0.306 & 0.335 & \textcolor{blue}{0.294} & 0.311 & 0.335 & 0.314 \\
 & 720 & \textcolor{red}{0.340}$\pm$0.002 & 0.348 & 0.345$\pm$0.001 & 0.349 & 0.639$\pm$0.003 & 0.710 & 0.418 & 0.359 & 0.381 & \textcolor{blue}{0.343} & 0.360 & 0.386 & 0.349 \\
\cmidrule(lr){2-15}
 & Avg & \textcolor{red}{0.274} & 0.281 & \textcolor{blue}{0.275} & 0.279 & 0.515 & 0.572 & 0.315 & 0.287 & 0.317 & \textcolor{red}{0.274} & 0.293 & 0.320 & 0.288 \\
\midrule
\multirow{5}{*}{\rotatebox{90}{Solar-Energy}}
 & 96  & 0.280$\pm$0.001 & 0.286 & \textcolor{red}{0.231}$\pm$0.001 & \textcolor{blue}{0.237} & 0.335$\pm$0.001 & 0.372 & 0.331 & 0.292 & 0.378 & 0.292 & 0.291 & 0.399 & 0.251 \\
 & 192 & 0.302$\pm$0.001 & 0.310 & \textcolor{red}{0.255}$\pm$0.001 & \textcolor{blue}{0.261} & 0.368$\pm$0.002 & 0.409 & 0.725 & 0.318 & 0.398 & 0.320 & 0.294 & 0.416 & 0.272 \\
 & 336 & 0.304$\pm$0.001 & 0.315 & \textcolor{red}{0.266}$\pm$0.002 & \textcolor{blue}{0.273} & 0.386$\pm$0.002 & 0.429 & 0.735 & 0.330 & 0.415 & 0.344 & 0.300 & 0.430 & 0.285 \\
 & 720 & 0.309$\pm$0.002 & 0.317 & \textcolor{red}{0.268}$\pm$0.000 & \textcolor{blue}{0.275} & 0.389$\pm$0.002 & 0.432 & 0.765 & 0.337 & 0.413 & 0.352 & 0.303 & 0.425 & 0.292 \\
\cmidrule(lr){2-15}
 & Avg & 0.299 & 0.307 & \textcolor{red}{0.255} & \textcolor{blue}{0.262} & 0.370 & 0.411 & 0.639 & 0.319 & 0.401 & 0.327 & 0.297 & 0.417 & 0.275 \\
\bottomrule
\end{tabular}
\end{small}
\end{threeparttable}}%
\label{tab:mae_std}
\end{table*}

\section{Experiments}
\label{Experiment}
In this section, we first introduce the experimental setup, followed by a comparison of our method with other approaches on eight benchmark datasets. Second, we conduct ablation studies to evaluate the contribution of each submodule. Thrid, we show the hyperparameter sensitivity of our method. Next, we analysis the computational complexity of our method. Finally, we present a visual comparison between our method and other methods. More experiments are provided in \textbf{Appendix \ref{Empirical_Analysis_Details}}.

\subsection{Experimental Settings}
\textbf{Datasets:} We use eight real-world datasets, including ECL, ETTh2 (h1,h2,m1,m2), Traffic, Weather in \cite{liu2023itransformer}, and Solar-Energy dataset in \cite{LSTNet}. We use the same train-validation-test split as in \cite{liu2023itransformer}. See Appendix \ref{Dataset_Descriptions} for details.

\textbf{Baselines:} We select ten classic time series forecasting (TSF) methods for comparison. These include: 1) Transformer-based TSF methods such as PatchTST \cite{PatchTST}, iTransformer \cite{liu2023itransformer}, Transformer \cite{Transformer}, Crossformer \cite{Crossformer}, DSformer \cite{yu2023dsformer}, and PRReg CI(Transformer) \cite{han2024capacity}; and 2) other widely used TSF methods from different model families, including DLinear \cite{DLinear}, SpareTSF \cite{lin2024sparsetsf}, TIDE \cite{das2023long}, and TimesNet \cite{Timesnet}.

\textbf{Implementation Details:} All models are trained on an NVIDIA 4090 GPU. The evaluation metrics used are Mean Squared Error (MSE) and Mean Absolute Error (MAE). The current mainstream Transformer-based methods primarily handle three types of tokens: temporal tokens, patch tokens, and variable tokens. These tokens are generated using different tokenization strategies, which are detailed in Appendix \ref{Token_Acquisition}. To verify the effectiveness of our method across different types of tokens, we apply the proposed Topology Enhancement Method (TEM) to three methods. They are: Transformer \cite{Transformer}, which uses temporal tokens; PatchTST \cite{PatchTST}, which uses variable tokens; and iTransformer \cite{liu2023itransformer}, which also uses variable tokens. When applying TEM to the above methods, we use the same hyperparameters as reported in their original papers. Additional implementation details are provided in Appendix \ref{method_appe}. For the bi-level optimization, in the inner loop, we adopt the original update strategies and optimizers used by each of the aforementioned methods. In the outer loop, the positional/semantic topology weights are updated using the Adam optimizer with a decaying learning rate. The initial learning rate is set to $1e-3$.

\begin{table}
  \centering
  \caption{Ablation study of PTEM and STEM on \textbf{PatchTST}. 
           Look-back length is 96; prediction lengths $\in\{96,192,336,720\}$. 
           Lower is better; the best result in each row is highlighted in bold.}
  \label{tab:ablation_patchtst}
  \resizebox{0.8\textwidth}{!}{%
  \begin{tabular}{l|c|c|c|c|c}
    \toprule
    \textbf{Dataset} & \textbf{Pred Len} &
    \textbf{PatchTST} & \textbf{+STEM} & \textbf{+PTEM} & 
    \textbf{PatchTST} \\
    & & & & & \textbf{+TEM (PTEM \& STEM)} \\
    \midrule
    \multirow{5}{*}{ETTm1}
        & 96  & 0.329 / 0.367 & 0.327 / 0.365 & 0.324 / 0.362 & \textbf{0.323} / \textbf{0.361} \\
        & 192 & 0.367 / 0.385 & 0.365 / 0.383 & 0.362 / 0.380 & \textbf{0.361} / \textbf{0.379} \\
        & 336 & 0.399 / 0.410 & 0.395 / 0.407 & 0.392 / 0.404 & \textbf{0.390} / \textbf{0.402} \\
        & 720 & 0.454 / 0.439 & 0.450 / 0.435 & 0.446 / 0.431 & \textbf{0.444} / \textbf{0.429} \\
        & Avg & 0.387 / 0.400 & 0.384 / 0.397 & 0.382 / 0.395 & \textbf{0.380} / \textbf{0.393} \\[2pt]
    \cmidrule{1-6}
    \multirow{5}{*}{ECL}
        & 96  & 0.195 / 0.285 & 0.191 / 0.283 & 0.188 / 0.282 & \textbf{0.186} / \textbf{0.281} \\
        & 192 & 0.199 / 0.289 & 0.196 / 0.287 & 0.193 / 0.285 & \textbf{0.192} / \textbf{0.283} \\
        & 336 & 0.215 / 0.305 & 0.210 / 0.300 & 0.206 / 0.296 & \textbf{0.203} / \textbf{0.291} \\
        & 720 & 0.256 / 0.337 & 0.252 / 0.329 & 0.247 / 0.323 & \textbf{0.245} / \textbf{0.318} \\
        & Avg & 0.216 / 0.304 & 0.213 / 0.300 & 0.210 / 0.297 & \textbf{0.207} / \textbf{0.293} \\[2pt]
    \cmidrule{1-6}
    \multirow{5}{*}{Weather}
        & 96  & 0.177 / 0.218 & 0.175 / 0.214 & 0.172 / 0.210 & \textbf{0.171} / \textbf{0.208} \\
        & 192 & 0.225 / 0.259 & 0.219 / 0.257 & 0.212 / 0.256 & \textbf{0.209} / \textbf{0.255} \\
        & 336 & 0.278 / 0.297 & 0.276 / 0.296 & 0.274 / 0.295 & \textbf{0.273} / \textbf{0.294} \\
        & 720 & 0.354 / 0.348 & 0.349 / 0.345 & 0.346 / 0.342 & \textbf{0.342} / \textbf{0.340} \\
        & Avg & 0.259 / 0.281 & 0.255 / 0.278 & 0.252 / 0.275 & \textbf{0.249} / \textbf{0.274} \\
    \bottomrule
  \end{tabular}}
\end{table}

\begin{table}
  \centering
  \caption{Ablation study of PTEM and STEM on \textbf{iTransformer}. 
           Look-back length is 96; prediction lengths $\in\{96,192,336,720\}$. 
           The best result in each row is highlighted in bold.}
  \label{tab:ablation_itransformer}
  \resizebox{0.8\textwidth}{!}{%
  \begin{tabular}{l|c|c|c|c|c}
    \toprule
    \textbf{Dataset} & \textbf{Pred Len} &
    \textbf{iTransformer} & \textbf{+STEM} & \textbf{+PTEM} & 
    \textbf{iTransformer} \\
    & & & & & \textbf{+TEM (PTEM \& STEM)} \\
    \midrule
    \multirow{5}{*}{ETTm1}
        & 96  & 0.334 / 0.368 & 0.329 / 0.364 & 0.324 / 0.361 & \textbf{0.321} / \textbf{0.359} \\
        & 192 & 0.377 / 0.391 & 0.374 / 0.390 & 0.371 / 0.389 & \textbf{0.369} / \textbf{0.387} \\
        & 336 & 0.426 / 0.420 & 0.421 / 0.416 & 0.416 / 0.412 & \textbf{0.413} / \textbf{0.410} \\
        & 720 & 0.491 / 0.459 & 0.483 / 0.454 & 0.474 / 0.449 & \textbf{0.470} / \textbf{0.447} \\
        & Avg & 0.407 / 0.410 & 0.401 / 0.406 & 0.396 / 0.402 & \textbf{0.393} / \textbf{0.401} \\[2pt]
    \cmidrule{1-6}
    \multirow{5}{*}{ECL}
        & 96  & 0.148 / 0.240 & 0.145 / 0.238 & 0.143 / 0.237 & \textbf{0.141} / \textbf{0.235} \\
        & 192 & 0.162 / 0.253 & 0.160 / 0.253 & 0.158 / 0.252 & \textbf{0.156} / \textbf{0.252} \\
        & 336 & 0.178 / 0.269 & 0.174 / 0.267 & 0.170 / 0.265 & \textbf{0.169} / \textbf{0.264} \\
        & 720 & 0.225 / 0.317 & 0.219 / 0.312 & 0.213 / 0.307 & \textbf{0.209} / \textbf{0.305} \\
        & Avg & 0.178 / 0.270 & 0.174 / 0.268 & 0.171 / 0.266 & \textbf{0.169} / \textbf{0.264} \\[2pt]
    \cmidrule{1-6}
    \multirow{5}{*}{Weather}
        & 96  & 0.174 / 0.214 & 0.170 / 0.212 & 0.167 / 0.210 & \textbf{0.165} / \textbf{0.208} \\
        & 192 & 0.221 / 0.254 & 0.217 / 0.253 & 0.212 / 0.252 & \textbf{0.210} / \textbf{0.252} \\
        & 336 & 0.278 / 0.296 & 0.273 / 0.295 & 0.269 / 0.294 & \textbf{0.266} / \textbf{0.293} \\
        & 720 & 0.358 / 0.349 & 0.355 / 0.347 & 0.352 / 0.346 & \textbf{0.350} / \textbf{0.345} \\
        & Avg & 0.258 / 0.279 & 0.254 / 0.277 & 0.250 / 0.276 & \textbf{0.248} / \textbf{0.275} \\
    \bottomrule
  \end{tabular}}
\end{table}

\begin{table}
  \centering
  \caption{Ablation study of PTEM and STEM on the vanilla \textbf{Transformer}. 
           Look-back length is 96; prediction lengths $\in\{96,192,336,720\}$. 
           The best (lowest-error) result in each row is in bold.}
  \label{tab:ablation_transformer}
  \resizebox{0.8\textwidth}{!}{%
  \begin{tabular}{l|c|c|c|c|c}
    \toprule
    \textbf{Dataset} & \textbf{Pred Len} &
    \textbf{Transformer} & \textbf{+STEM} & \textbf{+PTEM} & 
    \textbf{Transformer} \\
    & & & & & \textbf{+TEM (PTEM \& STEM)} \\
    \midrule
    \multirow{5}{*}{ETTm1}
        & 96  & 0.528 / 0.489 & 0.507 / 0.469 & 0.486 / 0.450 & \textbf{0.475} / \textbf{0.440} \\
        & 192 & 0.596 / 0.520 & 0.572 / 0.499 & 0.548 / 0.478 & \textbf{0.536} / \textbf{0.468} \\
        & 336 & 0.674 / 0.559 & 0.647 / 0.537 & 0.620 / 0.514 & \textbf{0.607} / \textbf{0.503} \\
        & 720 & 0.777 / 0.610 & 0.746 / 0.586 & 0.715 / 0.561 & \textbf{0.699} / \textbf{0.549} \\
        & Avg & 0.643 / 0.545 & 0.617 / 0.523 & 0.592 / 0.502 & \textbf{0.579} / \textbf{0.491} \\[2pt]
    \cmidrule{1-6}
    \multirow{5}{*}{ECL}
        & 96  & 0.260 / 0.358 & 0.250 / 0.344 & 0.240 / 0.330 & \textbf{0.234} / \textbf{0.322} \\
        & 192 & 0.266 / 0.367 & 0.255 / 0.353 & 0.244 / 0.340 & \textbf{0.239} / \textbf{0.330} \\
        & 336 & 0.280 / 0.375 & 0.269 / 0.360 & 0.259 / 0.348 & \textbf{0.252} / \textbf{0.338} \\
        & 720 & 0.302 / 0.386 & 0.290 / 0.371 & 0.280 / 0.360 & \textbf{0.272} / \textbf{0.348} \\
        & Avg & 0.277 / 0.372 & 0.266 / 0.357 & 0.255 / 0.344 & \textbf{0.249} / \textbf{0.335} \\[2pt]
    \cmidrule{1-6}
    \multirow{5}{*}{Weather}
        & 96  & 0.395 / 0.427 & 0.379 / 0.410 & 0.363 / 0.394 & \textbf{0.356} / \textbf{0.384} \\
        & 192 & 0.619 / 0.560 & 0.594 / 0.538 & 0.570 / 0.515 & \textbf{0.557} / \textbf{0.504} \\
        & 336 & 0.689 / 0.594 & 0.662 / 0.571 & 0.635 / 0.547 & \textbf{0.620} / \textbf{0.535} \\
        & 720 & 0.926 / 0.710 & 0.889 / 0.682 & 0.852 / 0.653 & \textbf{0.833} / \textbf{0.639} \\
        & Avg & 0.657 / 0.572 & 0.631 / 0.549 & 0.605 / 0.537 & \textbf{0.591} / \textbf{0.515} \\
    \bottomrule
  \end{tabular}}
\end{table}

\begin{figure*}
    \centering
    \subfloat[]{\includegraphics[width=0.25\textwidth]{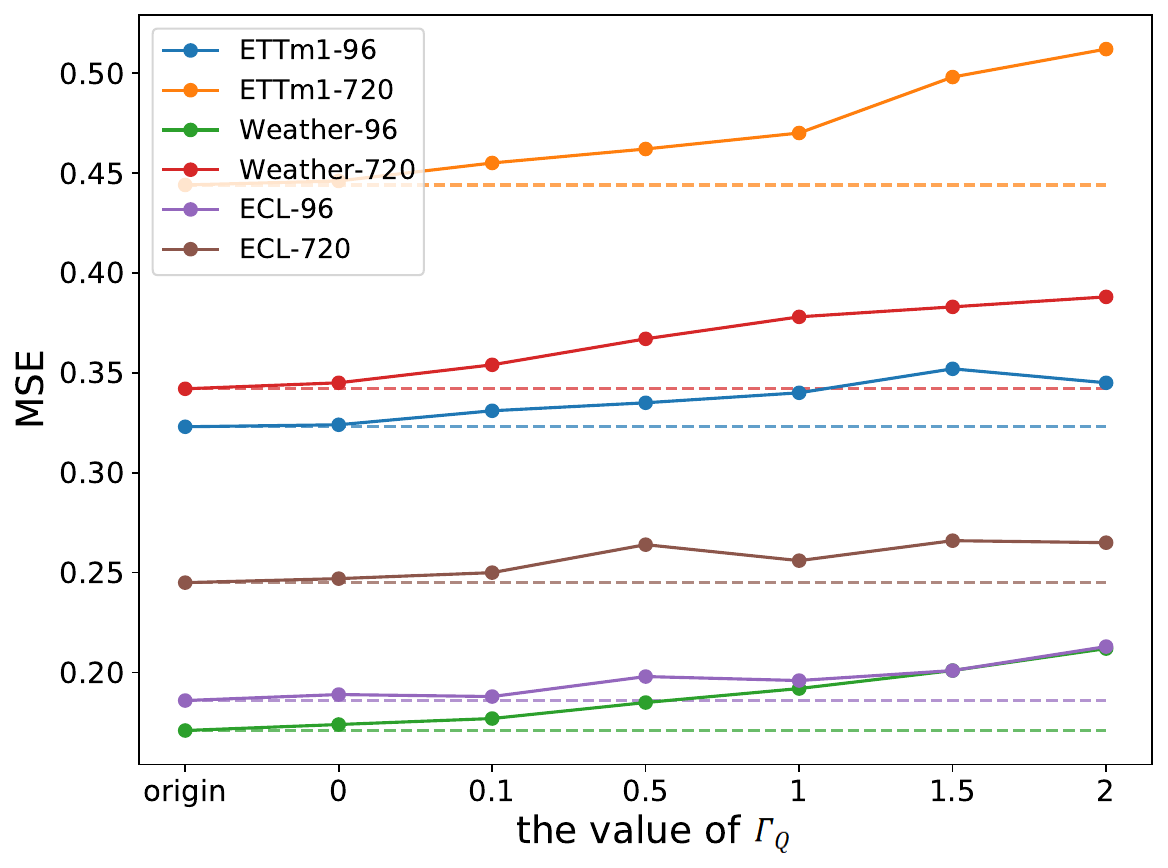}%
    \label{fig_adp_p1}}
    \hfil
    \subfloat[]{\includegraphics[width=0.25\textwidth]{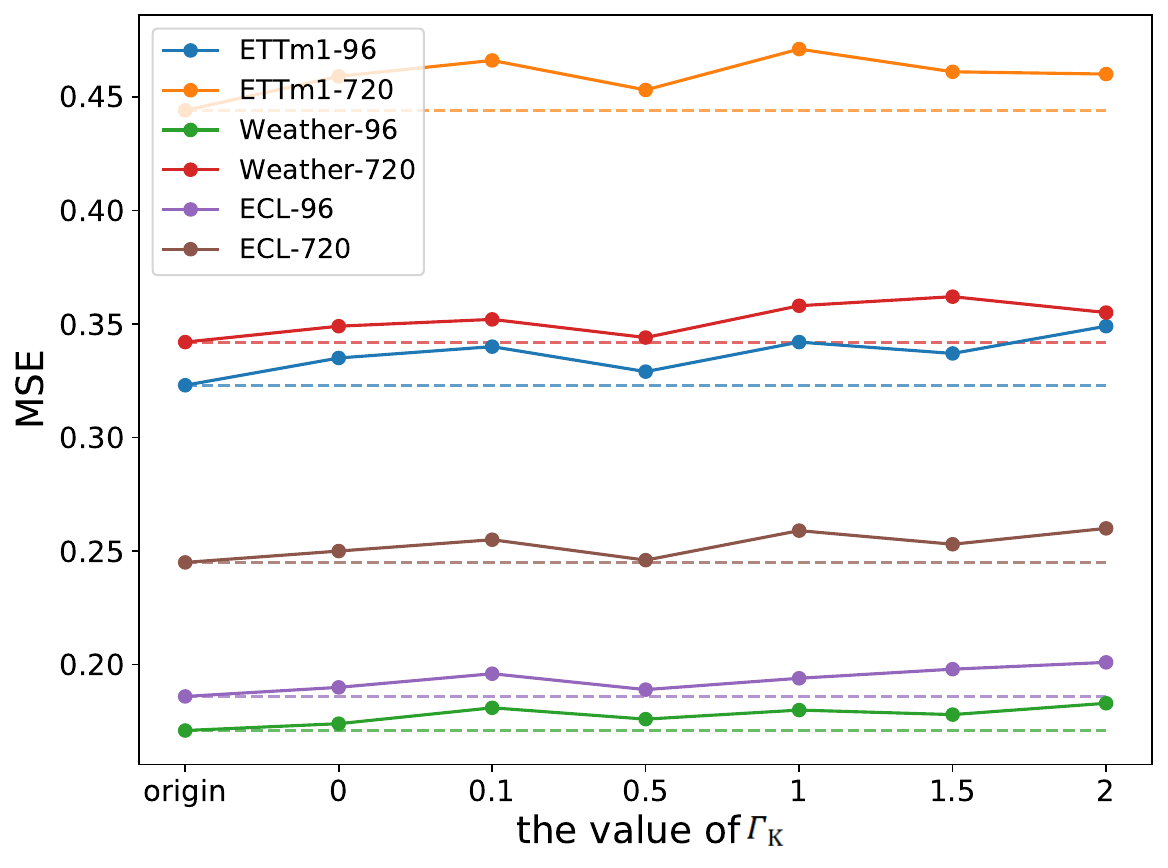}%
    \label{fig_adp_p2}}
    \hfil
    \subfloat[]{\includegraphics[width=0.25\textwidth]{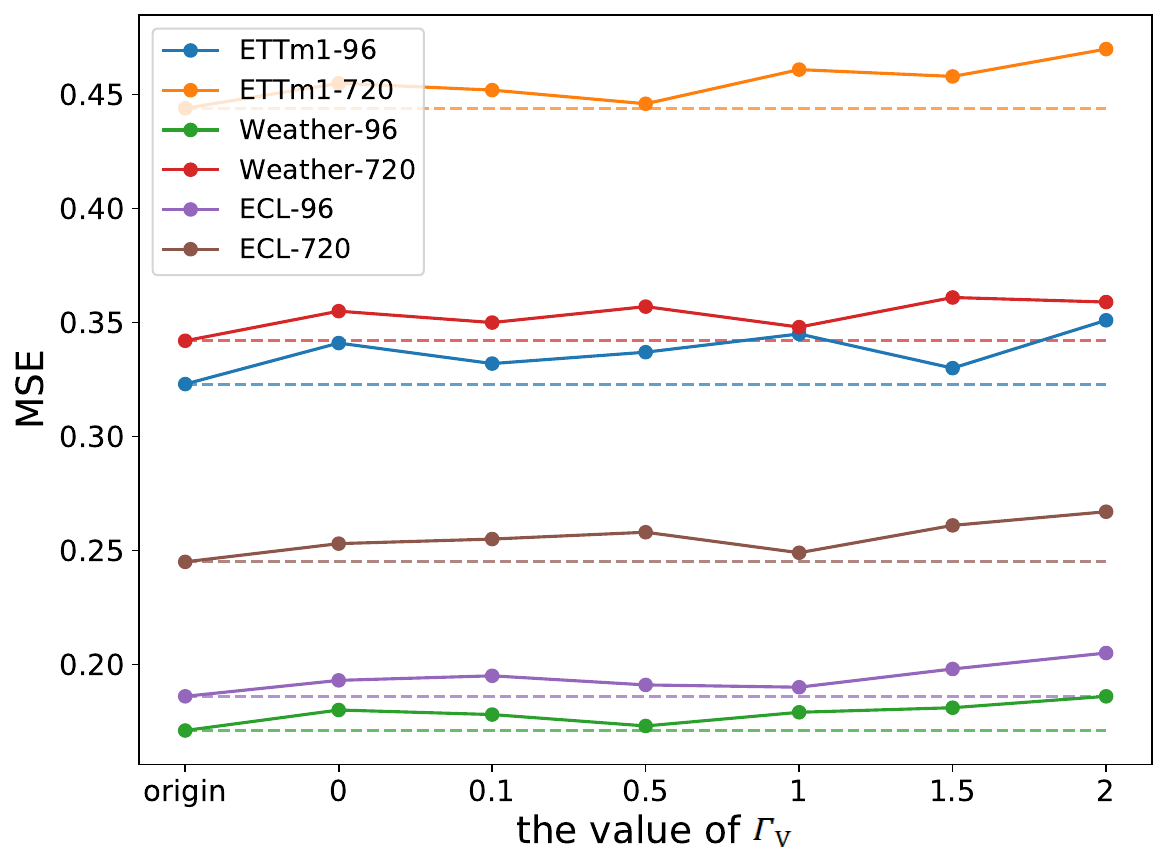}%
    \label{fig_adp_p3}}
    \hfil
    \subfloat[]{\includegraphics[width=0.25\textwidth]{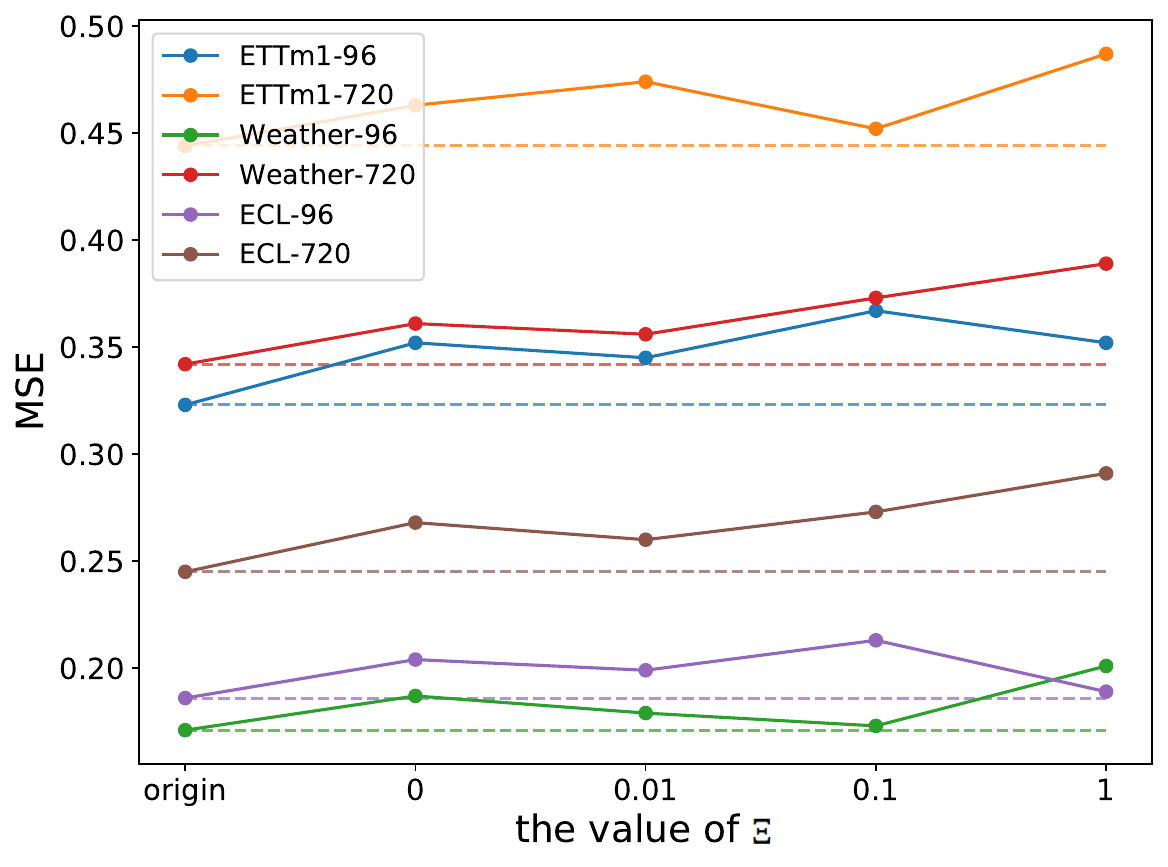}%
    \label{fig_adp_p4}}
    \hfil
    \caption{Illustration of our model’s performance under different fixed values of $\Gamma_Q$/$\Gamma_K$/$\Gamma_V$/$\Xi$ across various datasets (ETTm1, Weather, and ECL) and prediction lengths (96 and 720). The algorithm adopts \textbf{PatchTST+TEM}. The first point on each curve represents the performance of using the adaptive weight. The dashed lines, drawn horizontally through the first points of each curve, are intended to facilitate comparison between the performance of using the adaptive weight and that under other fixed-value settings.}
    \label{fig_adp_p}
\end{figure*}

\begin{figure*}
    \centering
    \subfloat[]{\includegraphics[width=0.25\textwidth]{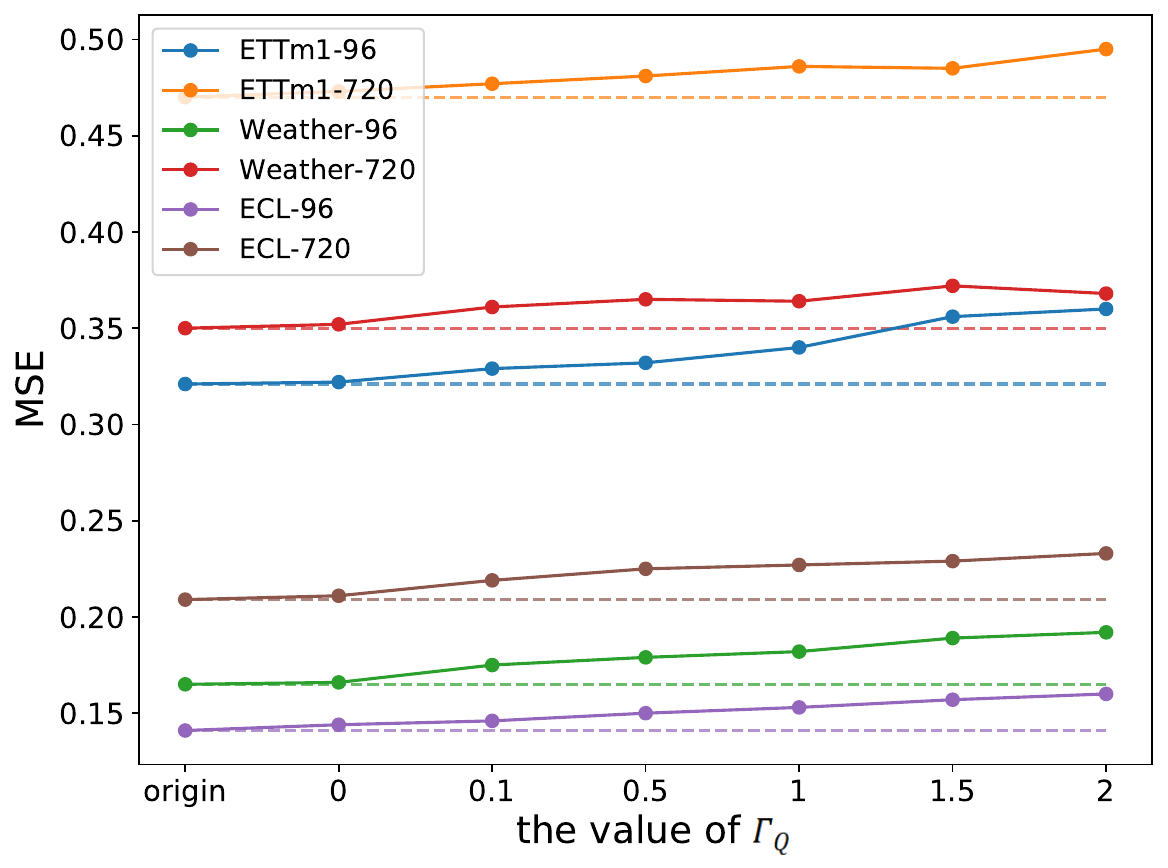}%
    \label{fig_adp_i1}}
    \hfil
    \subfloat[]{\includegraphics[width=0.25\textwidth]{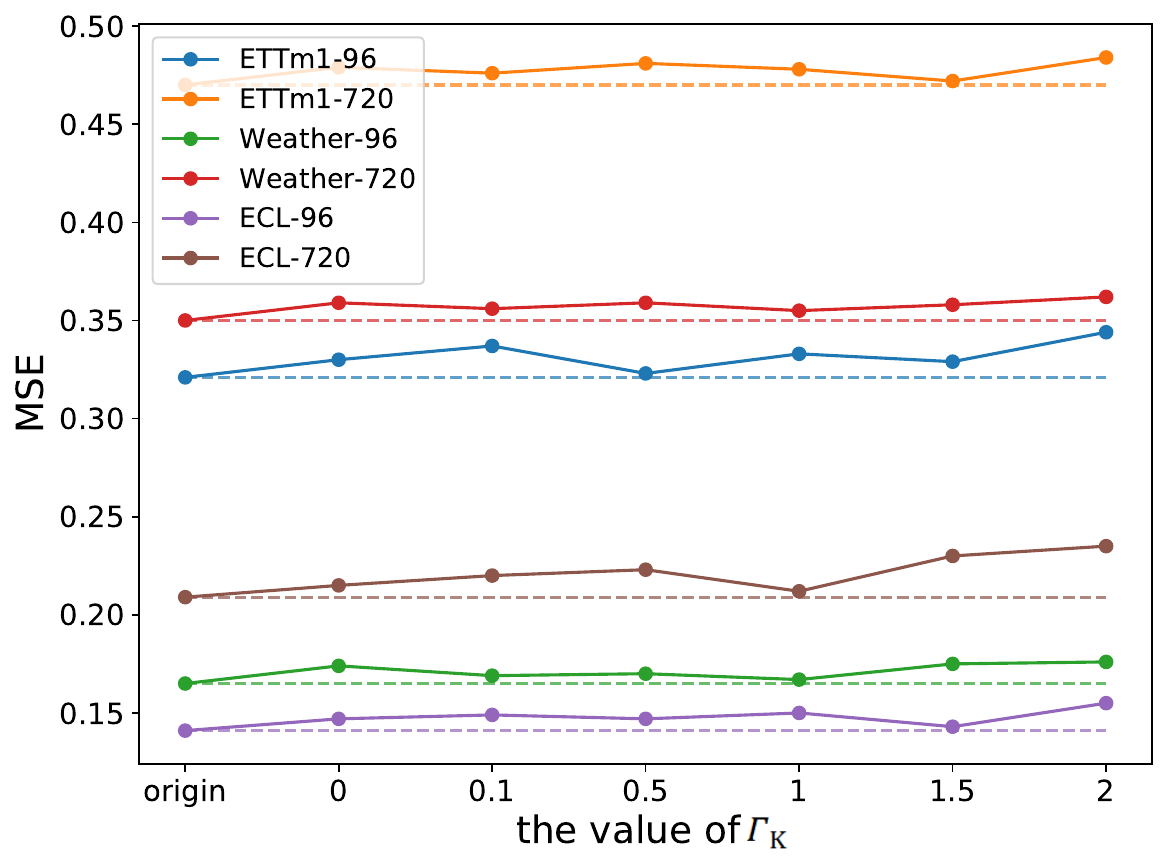}%
    \label{fig_adp_i2}}
    \hfil
    \subfloat[]{\includegraphics[width=0.25\textwidth]{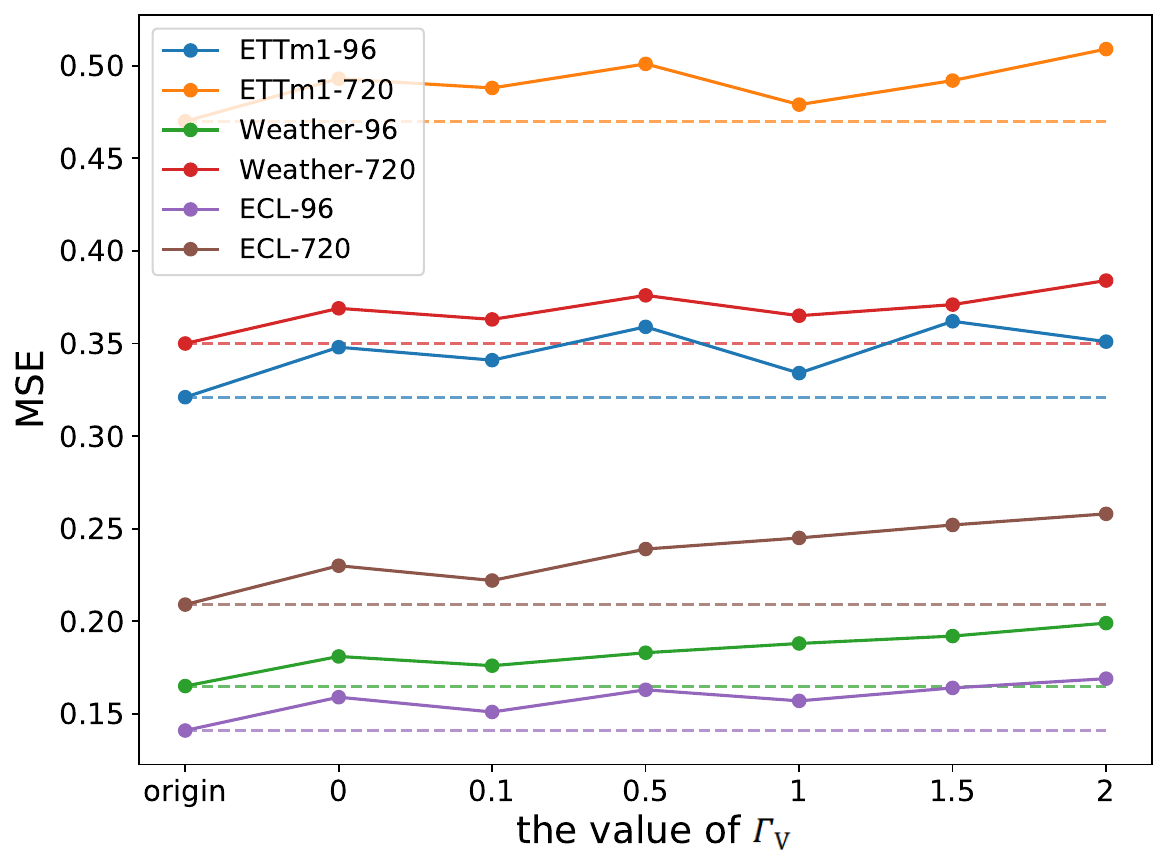}%
    \label{fig_adp_i3}}
    \hfil
    \subfloat[]{\includegraphics[width=0.25\textwidth]{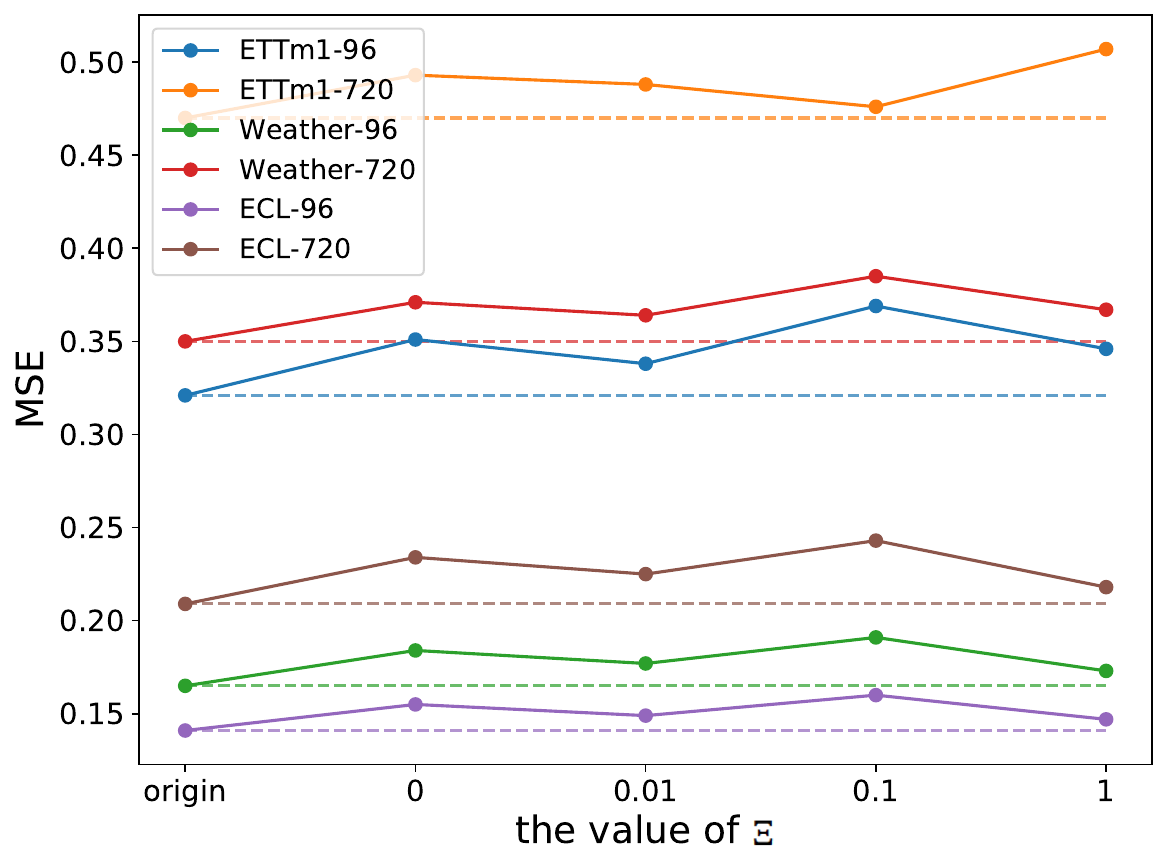}%
    \label{fig_adp_i4}}
    \hfil
    \caption{Illustration of our model’s performance under different fixed values of $\Gamma_Q$/$\Gamma_K$/$\Gamma_V$/$\Xi$ across various datasets (ETTm1, Weather, and ECL) and prediction lengths (96 and 720). The algorithm adopts \textbf{iTransformer+TEM}. The setting is the same as Figure \ref{fig_adp_p}.}
    \label{fig_adp_i}
\end{figure*}

\begin{figure*}
    \centering
    \subfloat[]{\includegraphics[width=0.25\textwidth]{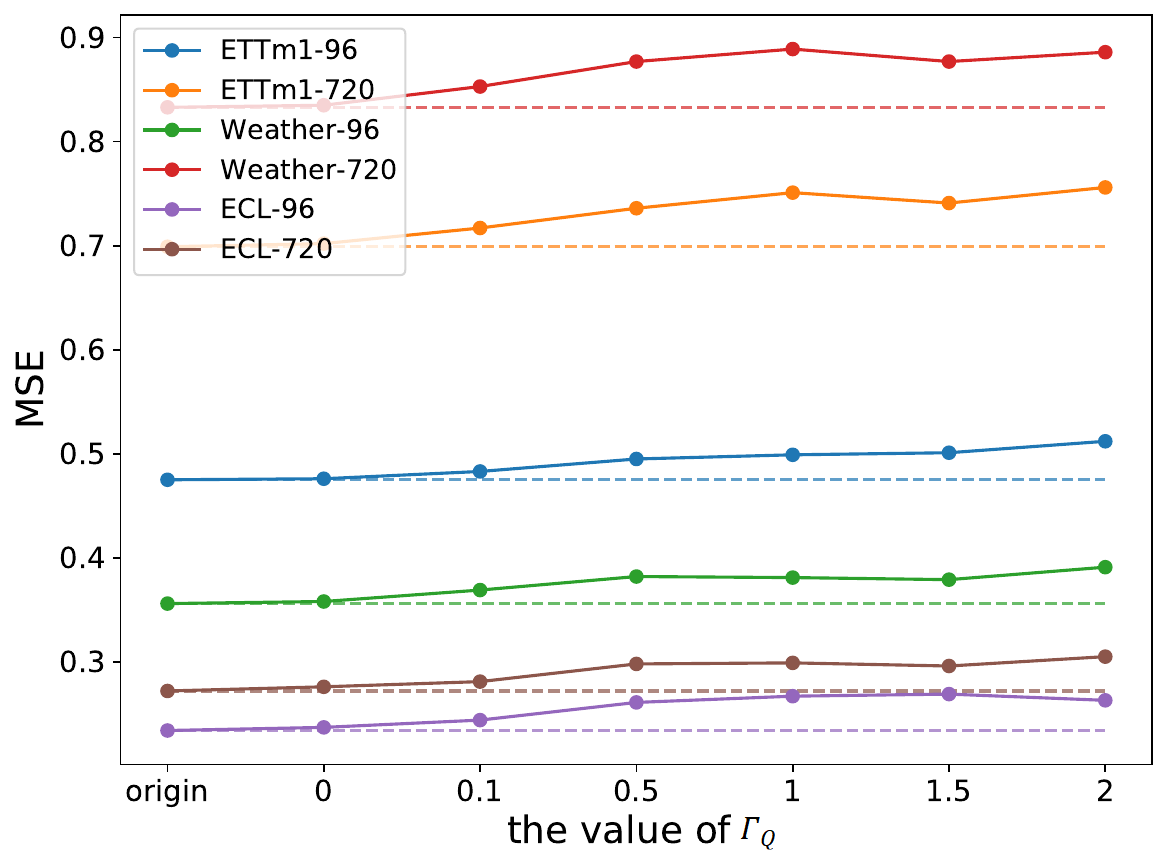}%
    \label{fig_adp_t1}}
    \hfil
    \subfloat[]{\includegraphics[width=0.25\textwidth]{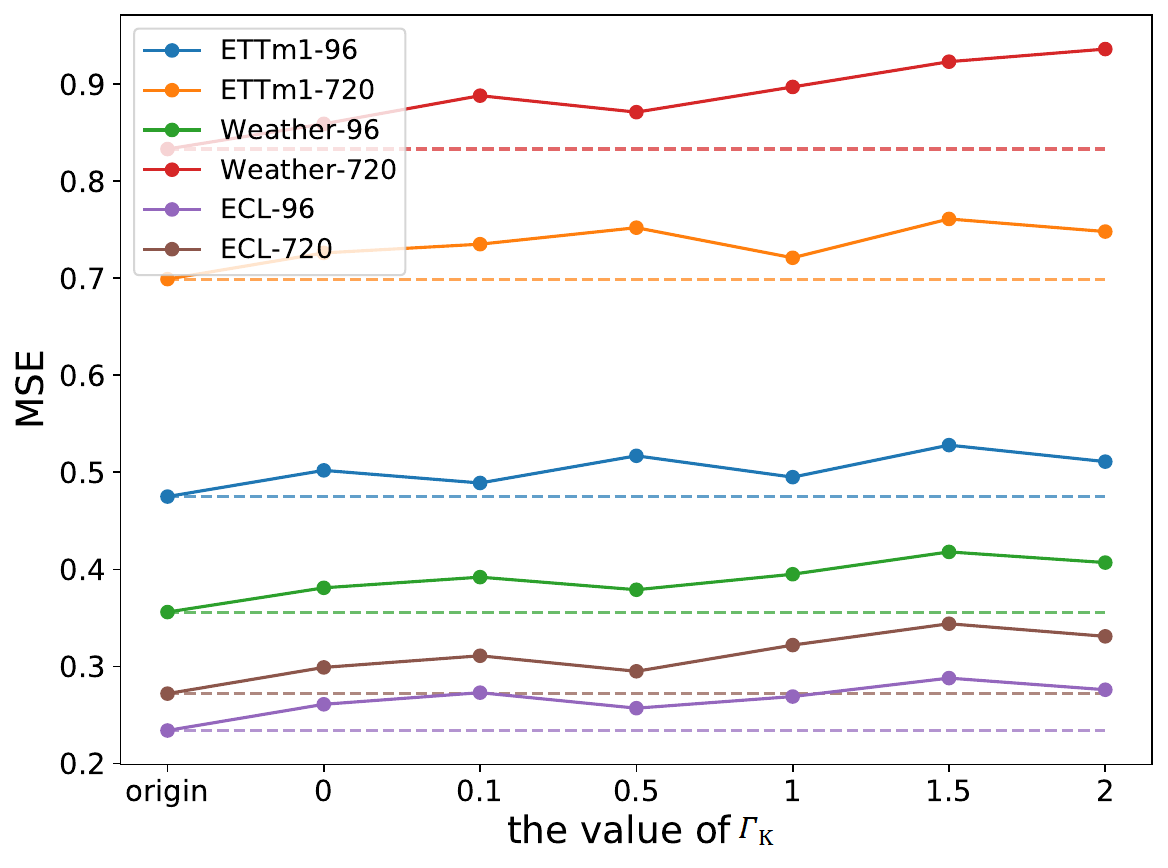}%
    \label{fig_adp_t2}}
    \hfil
    \subfloat[]{\includegraphics[width=0.25\textwidth]{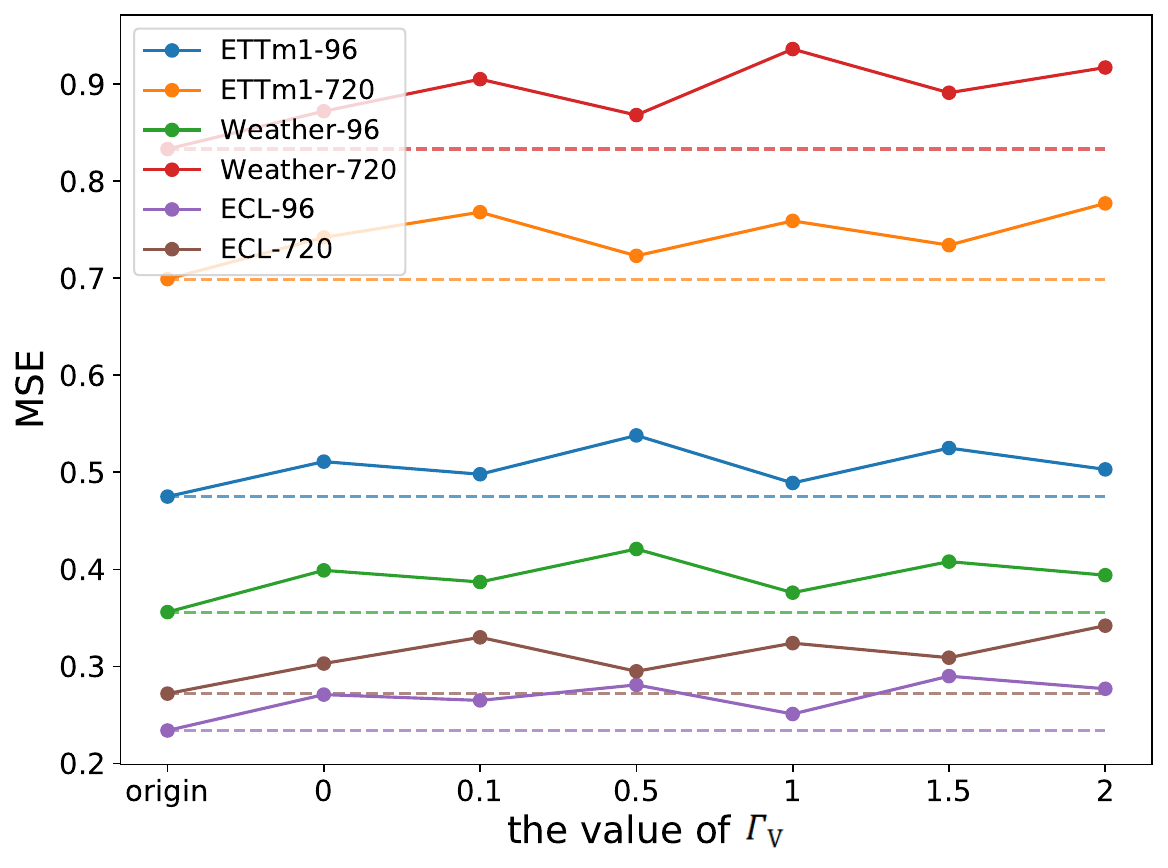}%
    \label{fig_adp_t3}}
    \hfil
    \subfloat[]{\includegraphics[width=0.25\textwidth]{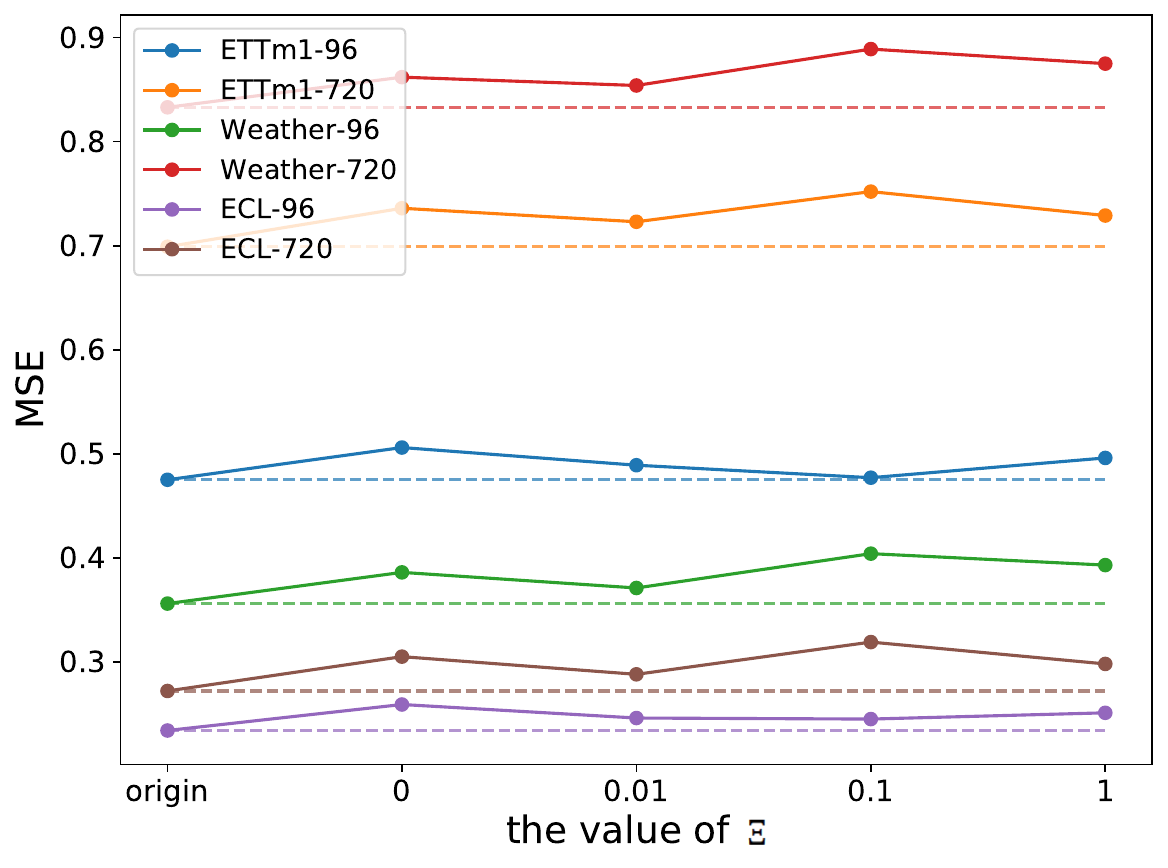}%
    \label{fig_adp_t4}}
    \hfil
    \caption{Illustration of our model’s performance under different fixed values of $\Gamma_Q$/$\Gamma_K$/$\Gamma_V$/$\Xi$ across various datasets (ETTm1, Weather, and ECL) and prediction lengths (96 and 720). The algorithm adopts \textbf{Transformer+TEM}. The setting is the same as Figure \ref{fig_adp_p}. }
    \label{fig_adp_t}
\end{figure*}

\begin{figure*}
    \centering
    \subfloat[]{\includegraphics[width=0.3\textwidth]{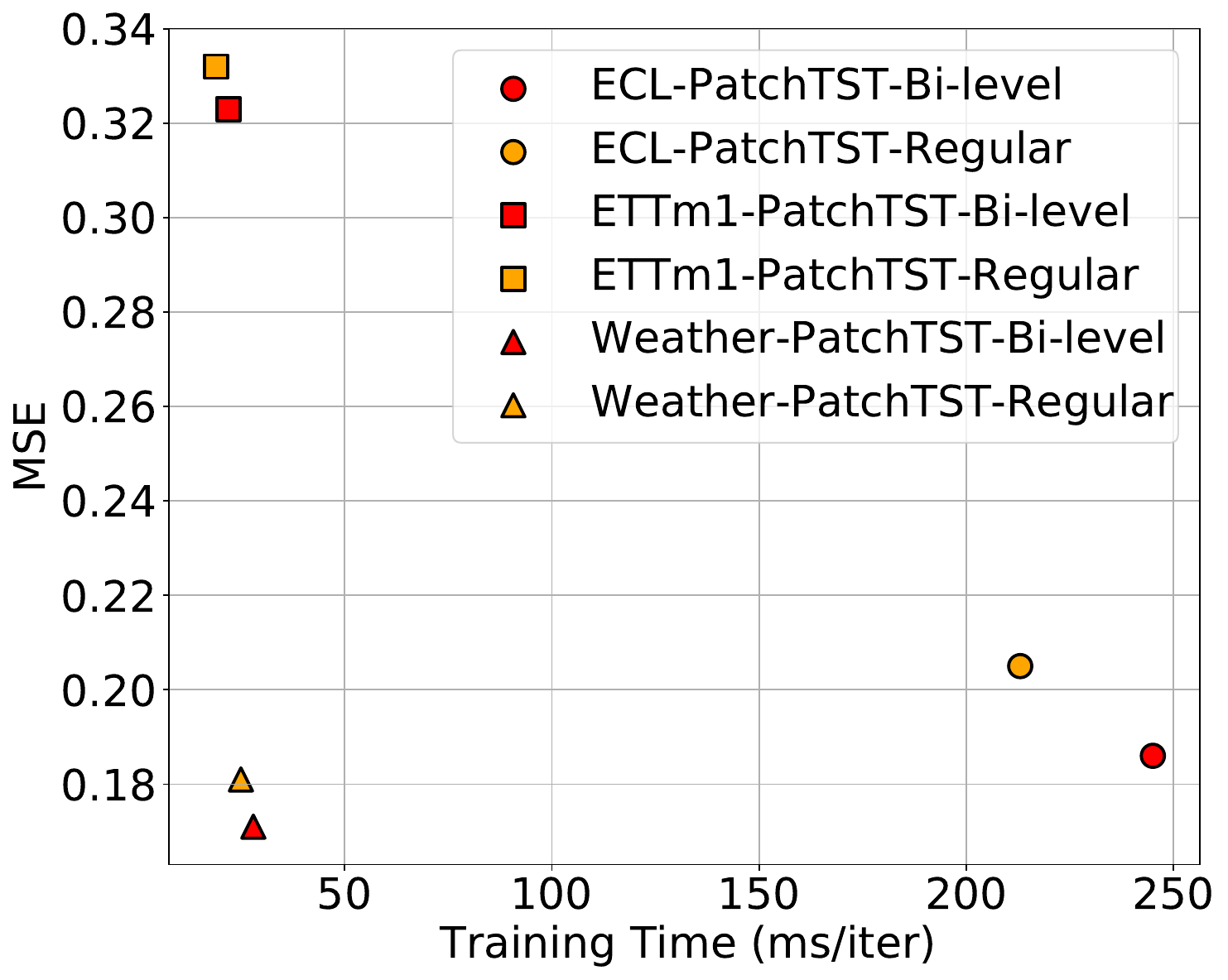}%
    \label{fig_bi_level1}}
    \hfil
    \subfloat[]{\includegraphics[width=0.3\textwidth]{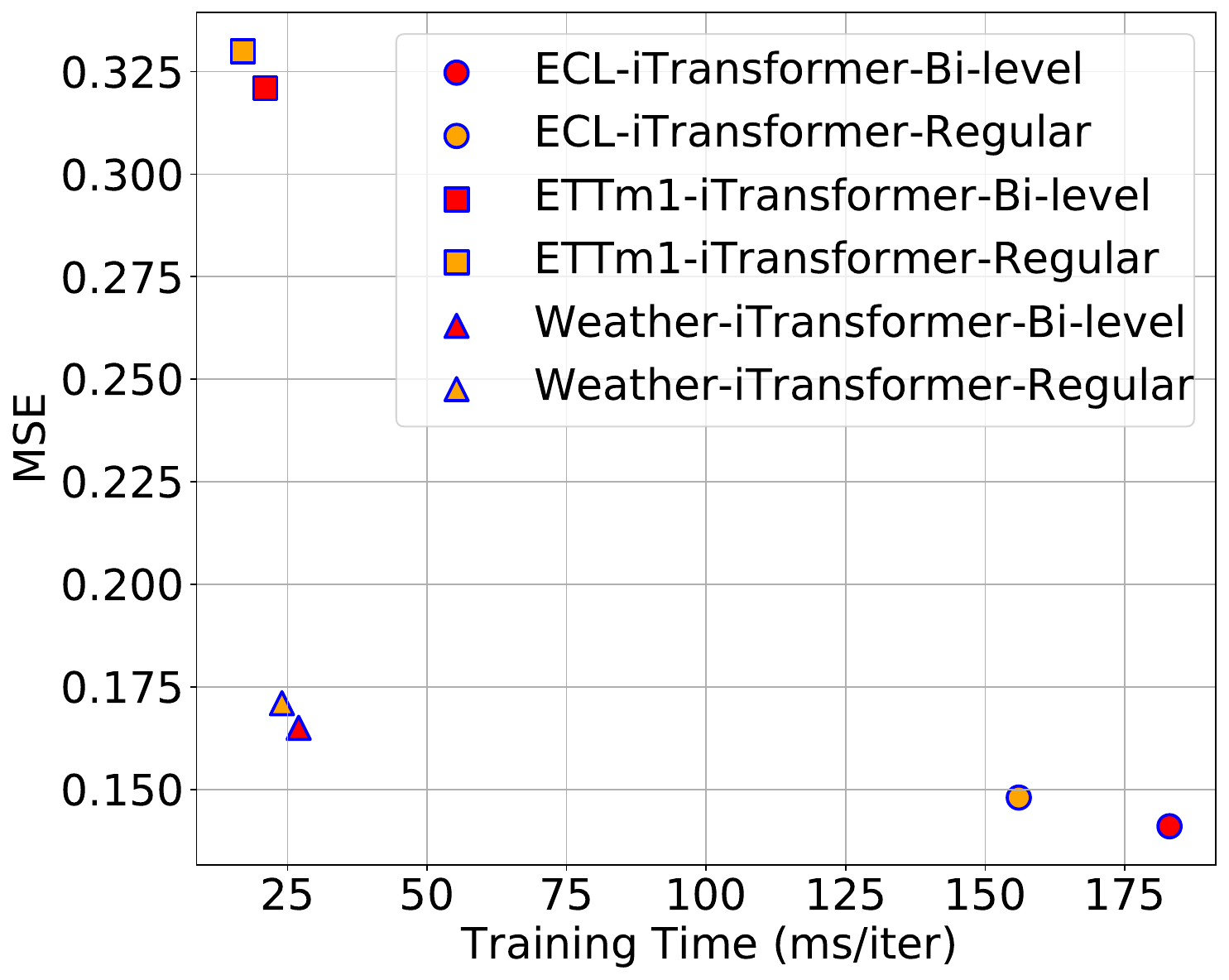}%
    \label{fig_bi_level2}}
    \hfil
    \subfloat[]{\includegraphics[width=0.3\textwidth]{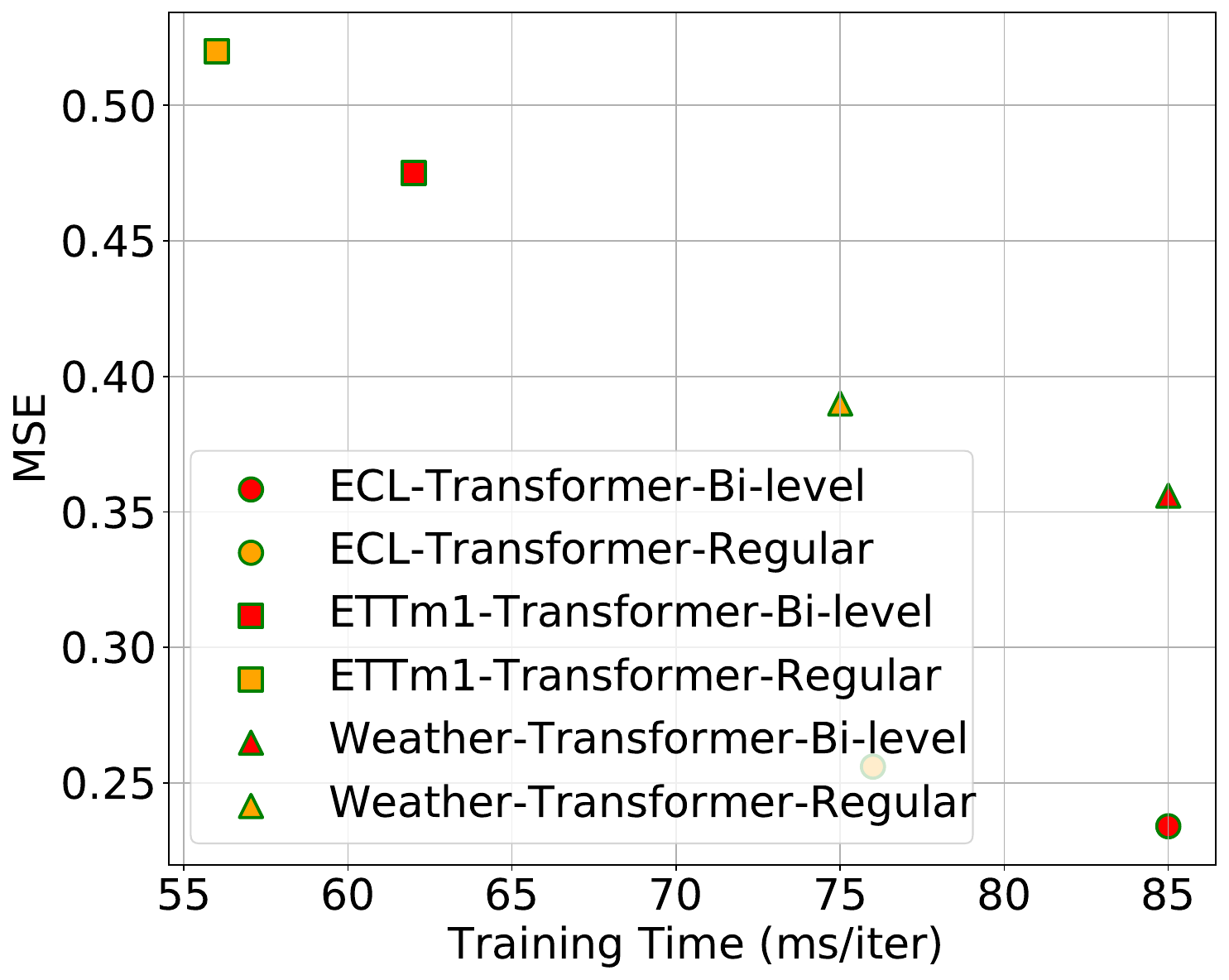}%
    \label{fig_bi_level3}}
    \hfil
    \caption{Comparison of training time and prediction accuracy between bi-level and regular optimization strategies. (a) PatchTST+TEM. (b) iTransformer+TEM. (c) Transformer+TEM. }
    \label{fig_bi_level}
\end{figure*}

\begin{figure*}
    \centering
    \subfloat[]{\includegraphics[width=0.31\textwidth]{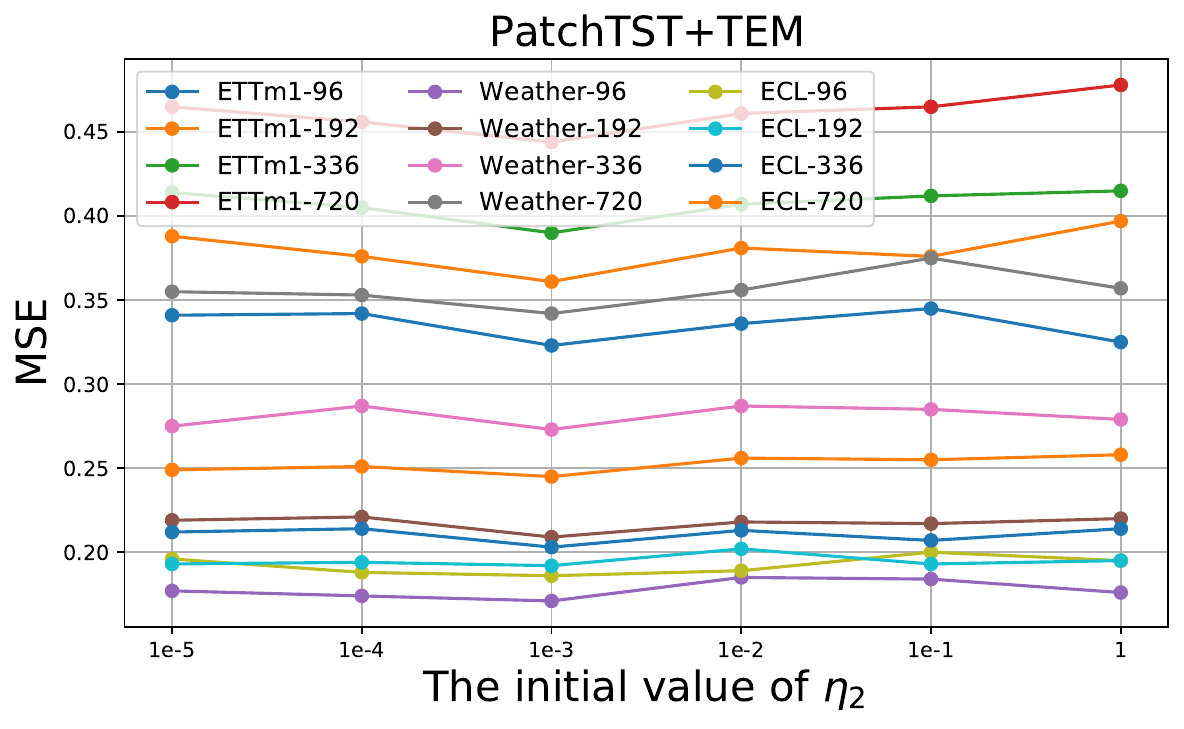}%
    \label{fig_eta_p}}
    \hfil
    \subfloat[]{\includegraphics[width=0.31\textwidth]{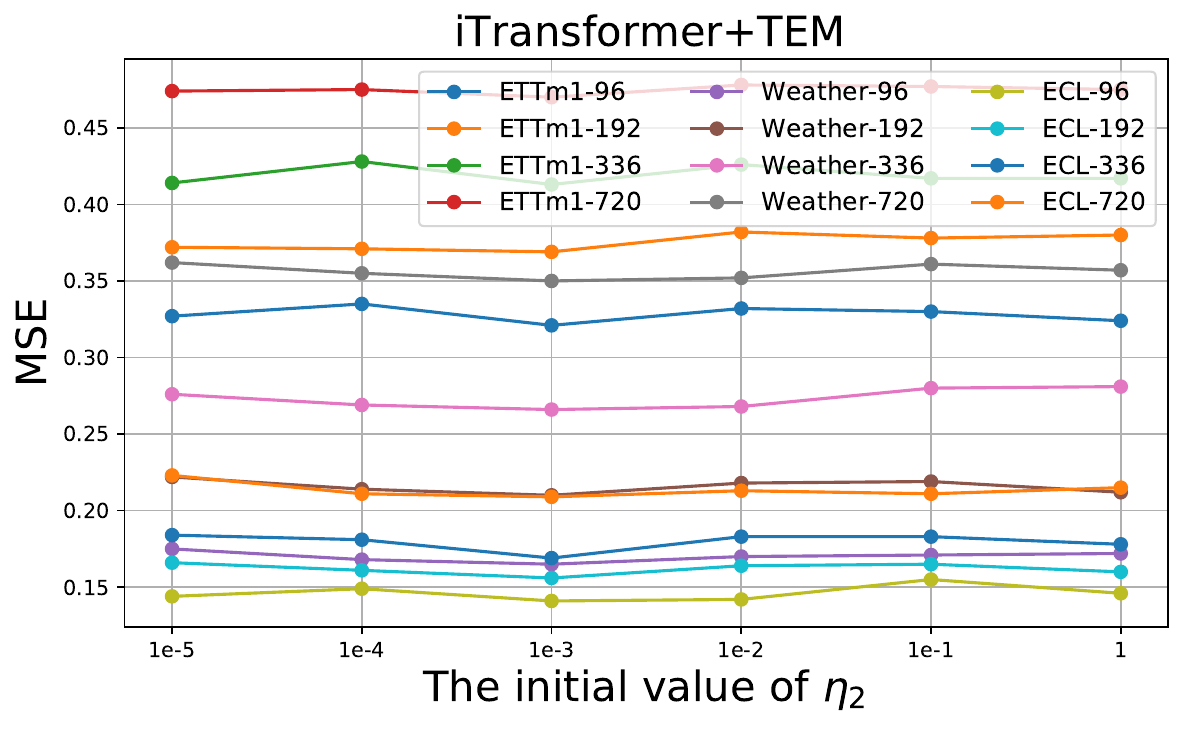}%
    \label{fig_et_i}}
    \hfil
    \subfloat[]{\includegraphics[width=0.31\textwidth]{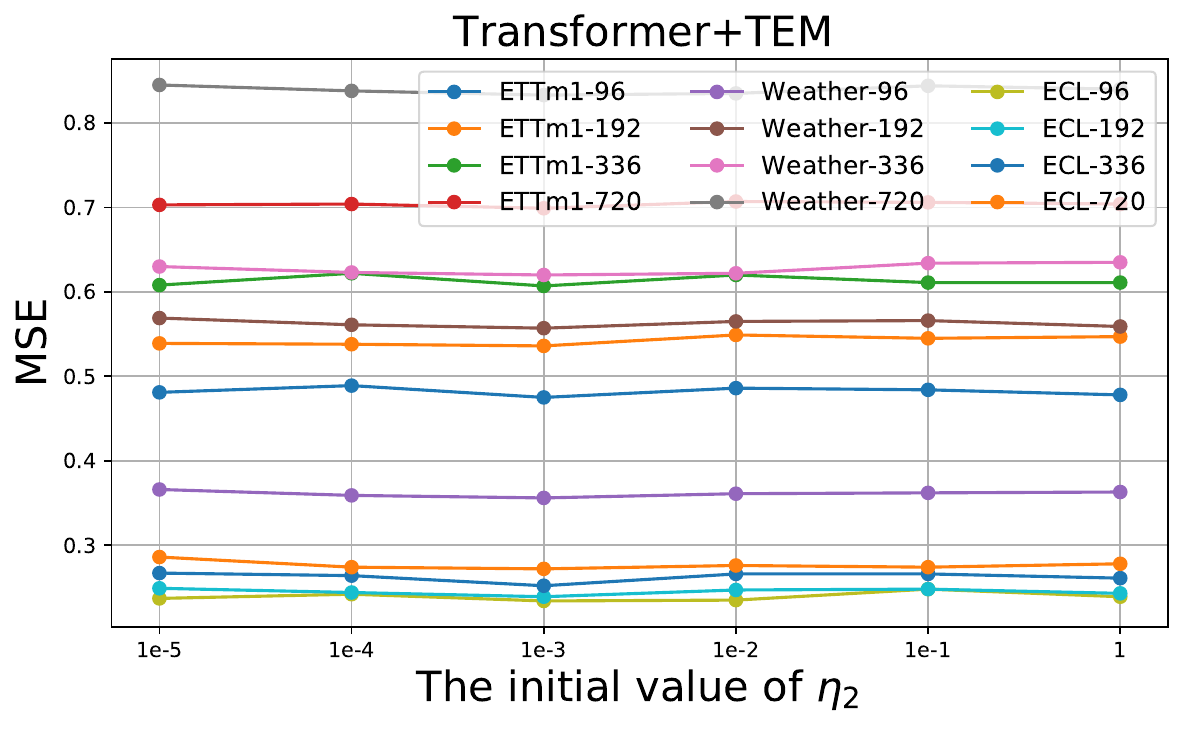}%
    \label{fig_eta_t}}
    \hfil
    \caption{Hyperparameter sensitivity analysis of $\eta_2^{init}$. (a) PatchTST+TEM, (b) iTransformer+TEM, (c) Transformer+TEM.}
    \label{fig_eta}
\end{figure*}

\begin{figure}
    \centering
    \subfloat[]{\includegraphics[width=0.5\textwidth]{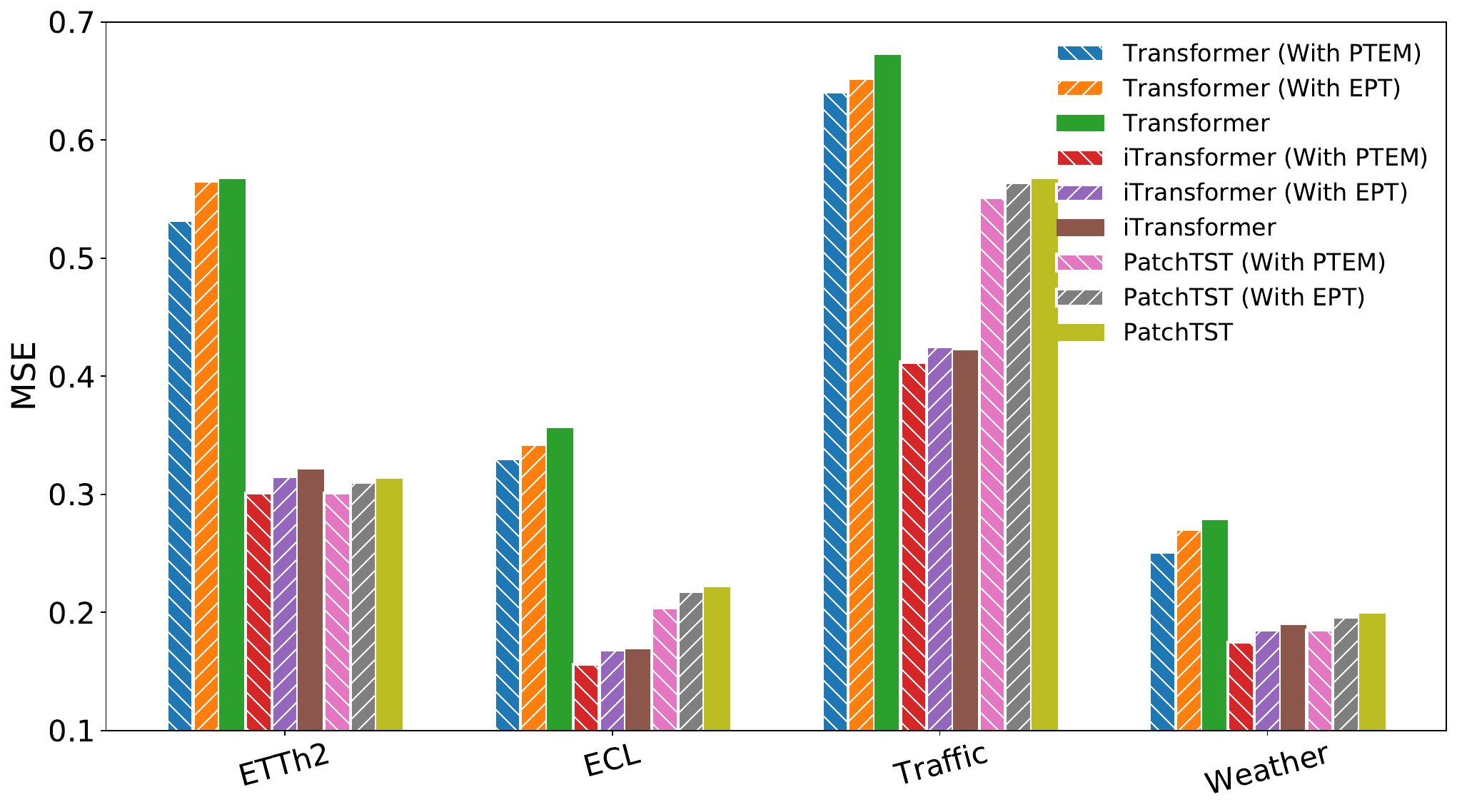}%
    \label{fig_moti_5}}
    \hfil
    \subfloat[]{\includegraphics[width=0.5\textwidth]{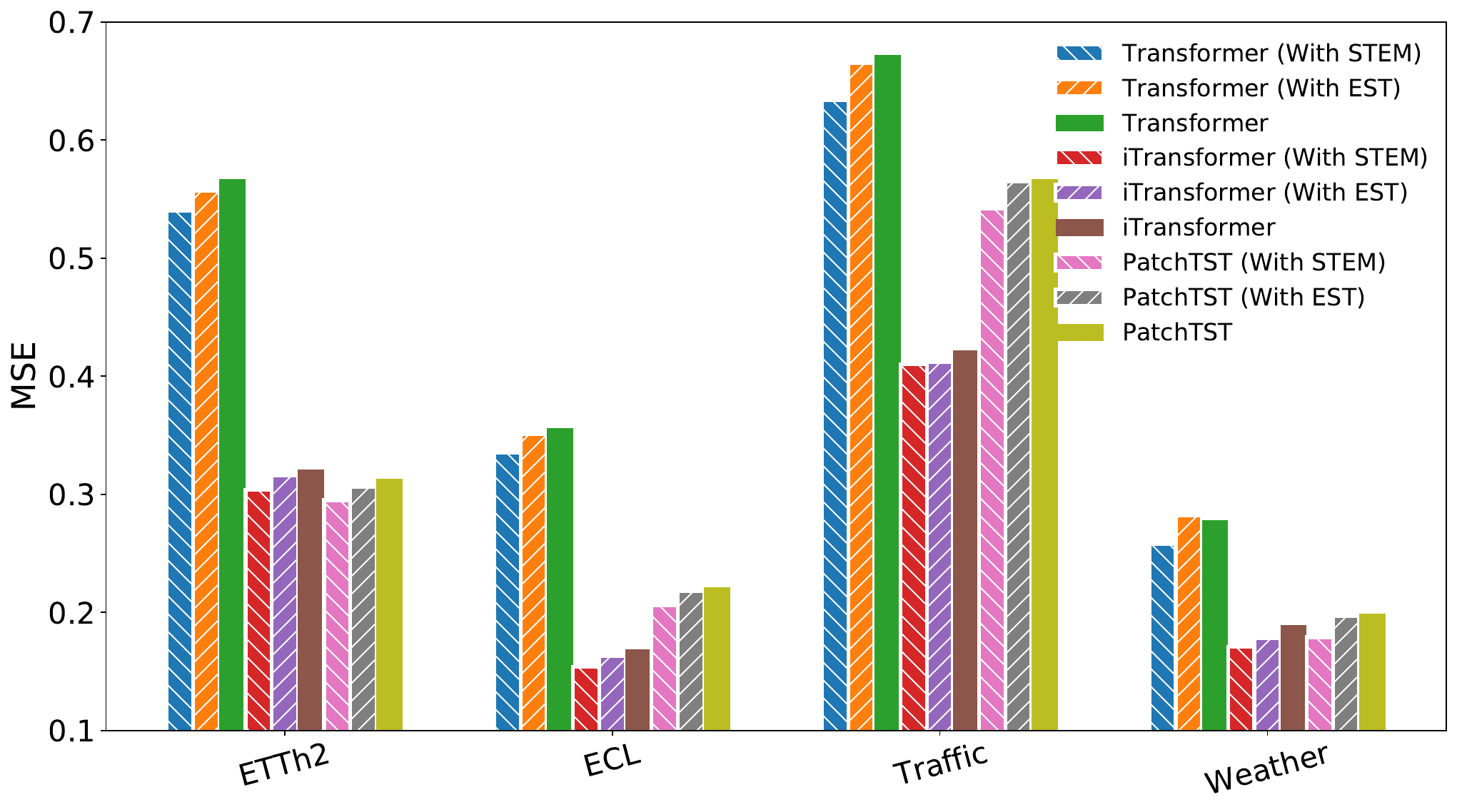}%
    \label{fig_moti_6}}
    \hfil
    \caption{(a) Changes in model performance after applying PTEM to the models used in Figure \ref{fig_moti_3}. (b) Changes in model performance after applying STEM to the models used in Figure \ref{fig_moti_4}.}
    \label{fig_moti_more}
\end{figure}

\begin{figure}
    \centering
    \subfloat[]{\includegraphics[width=0.33\textwidth]{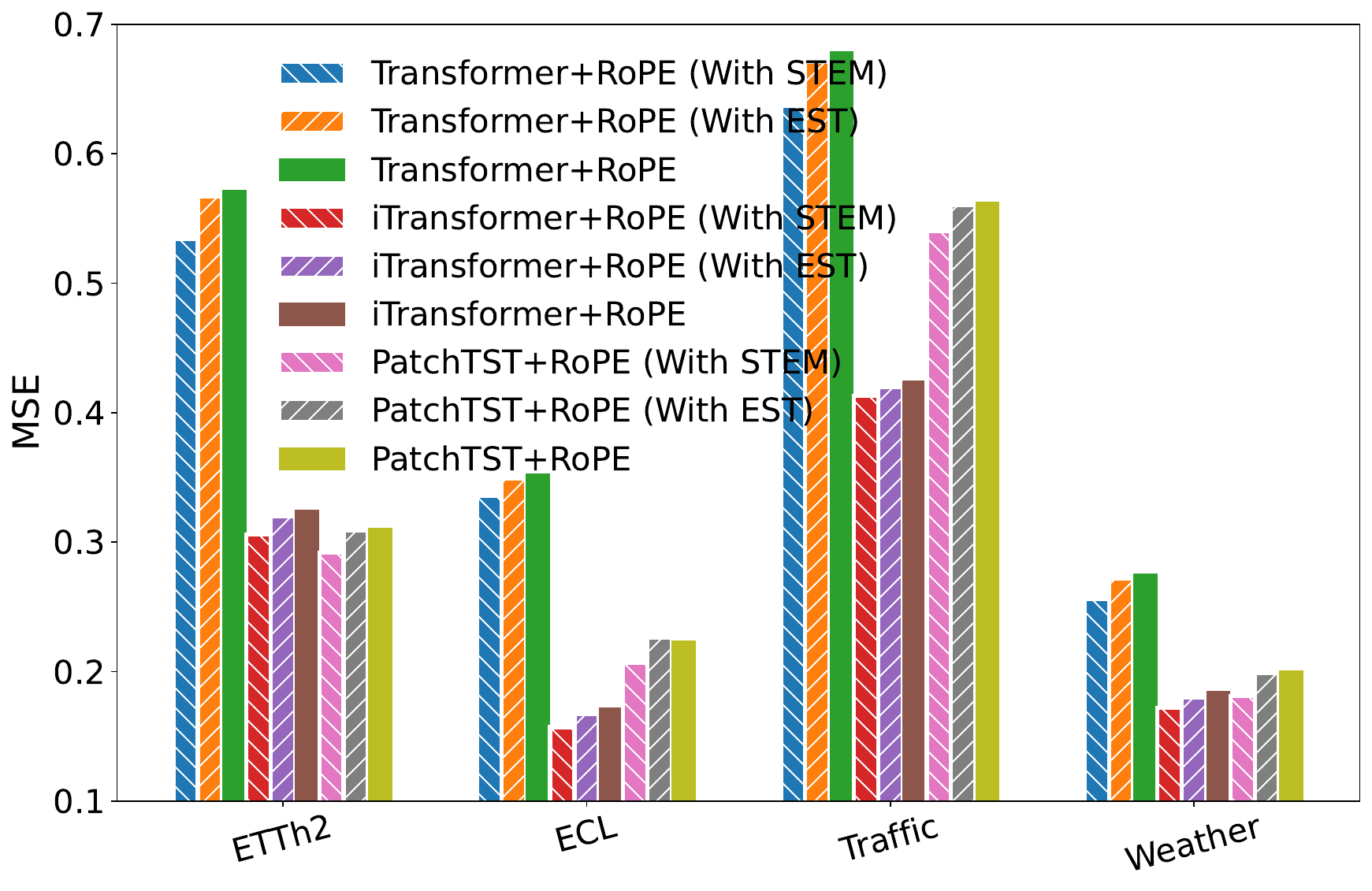}%
    \label{fig_moti_51}}
    \hfil
    \subfloat[]{\includegraphics[width=0.33\textwidth]{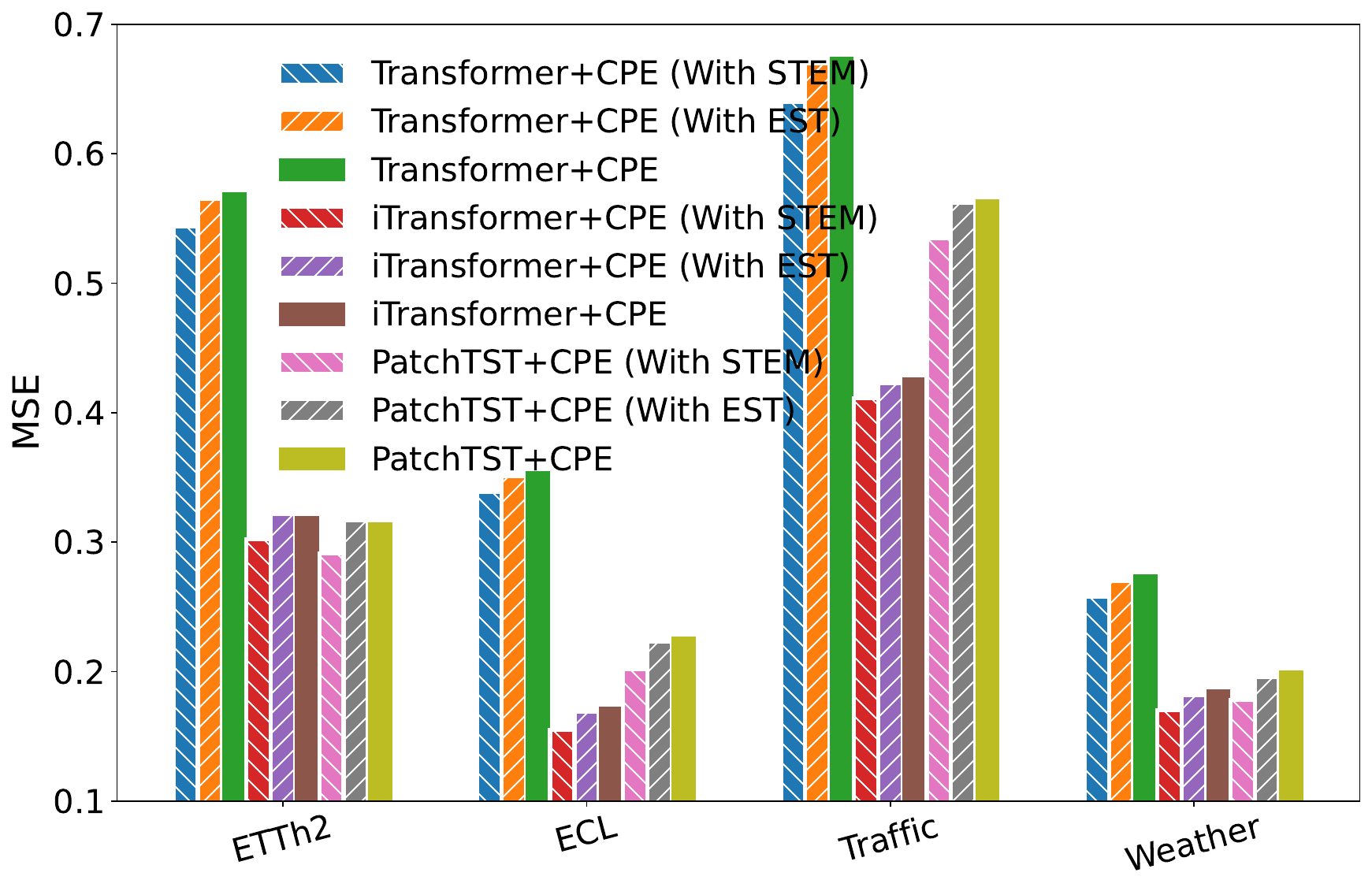}%
    \label{fig_moti_62}}
    \hfil
    \subfloat[]{\includegraphics[width=0.33\textwidth]{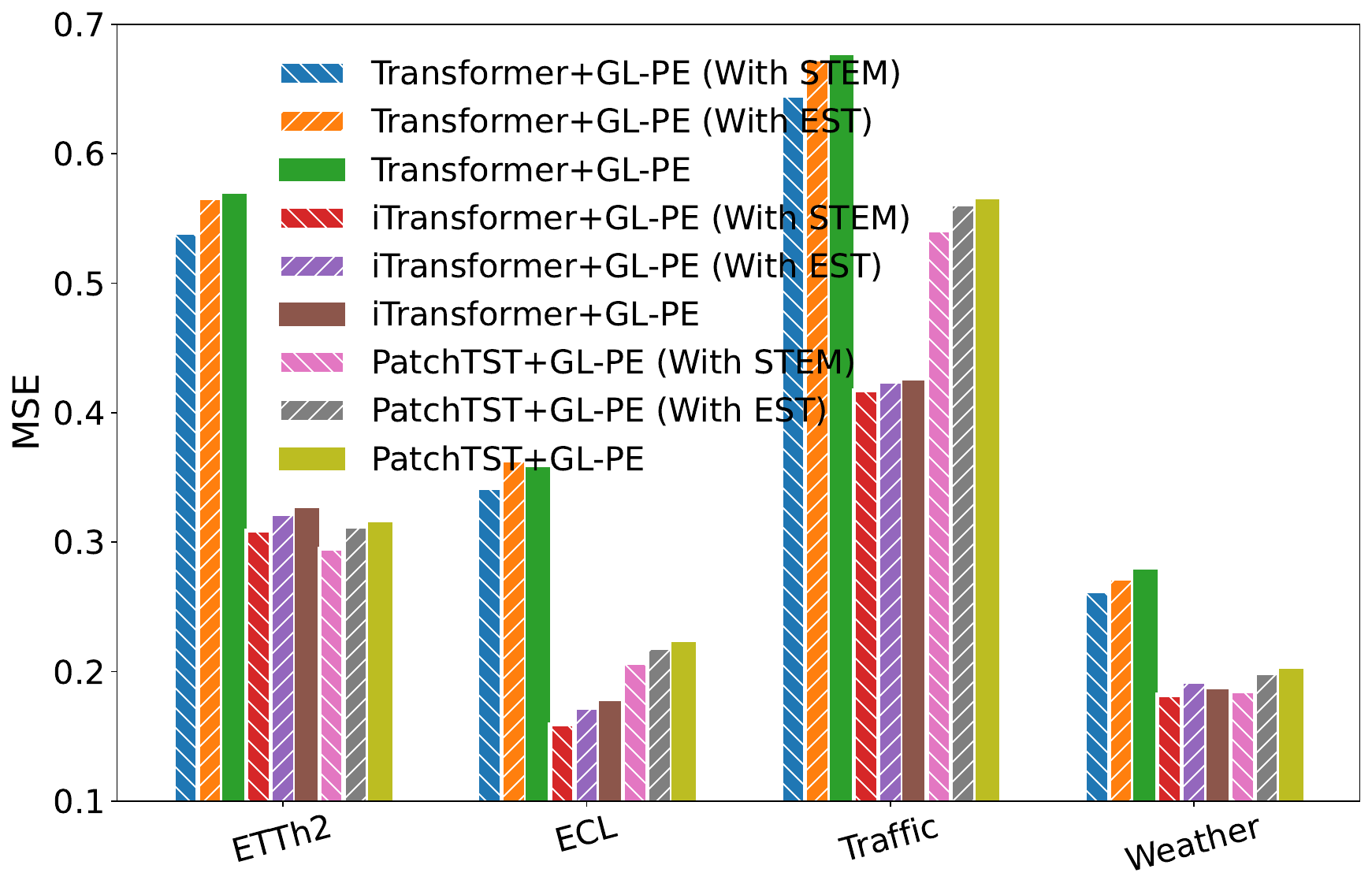}%
    \label{fig_moti_63}}
    \hfil
    \caption{(a) Changes in model performance after applying STEM to the models used in Figure \ref{fig_moti1_4}. (b) Changes in model performance after applying STEM to the models used in Figure \ref{fig_moti1_5}. (c) Changes in model performance after applying STEM to the models used in Figure \ref{fig_moti1_6}.}
    \label{fig_moti_more1}
\end{figure}

\begin{figure*}
  \centering
  \subfloat[]{\includegraphics[width=.46\textwidth]{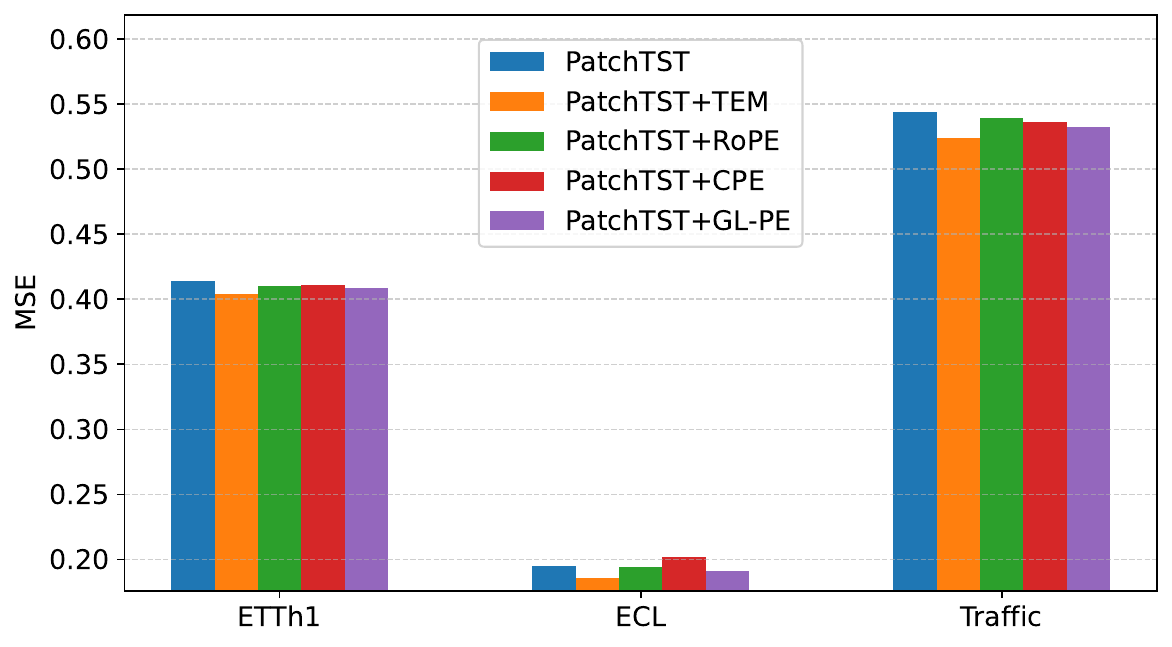}\label{fig_bar_1}}
  \hfill
  \subfloat[]{\includegraphics[width=.46\textwidth]{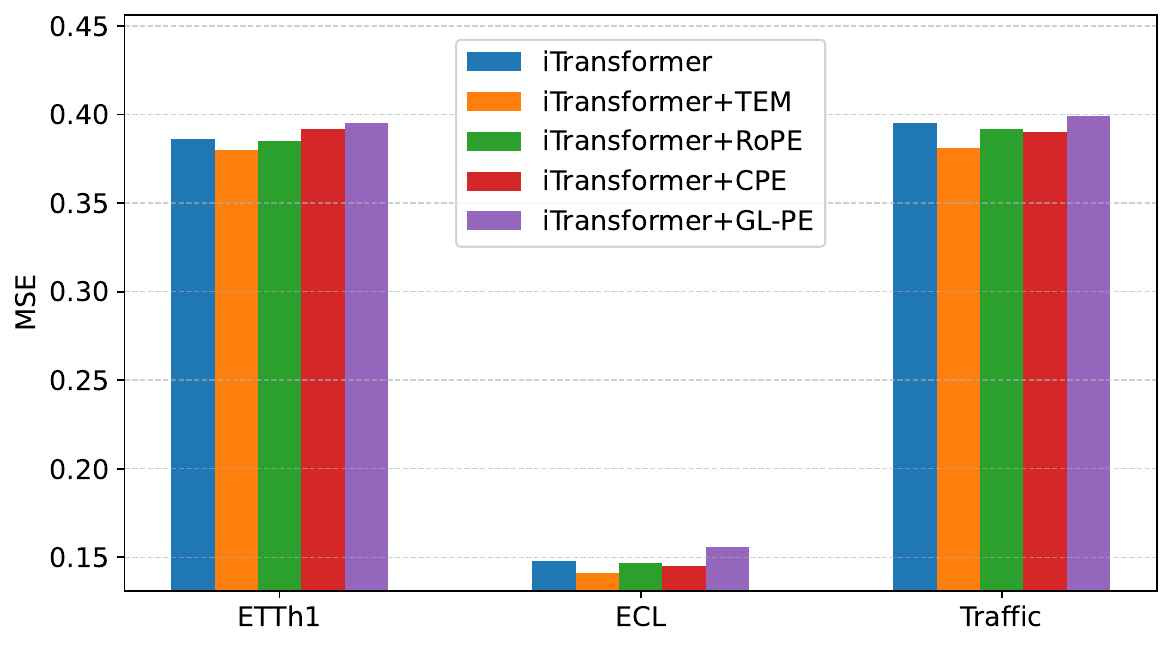}\label{fig_bar_2}}
  \caption{Comparison of the prediction performance of the proposed TEM and other positional encoding methods when applied to the TSF methods.}
  \label{fig_bar}
\end{figure*}

\begin{figure*}
    \centering
    \subfloat[]{\includegraphics[width=0.48\textwidth]{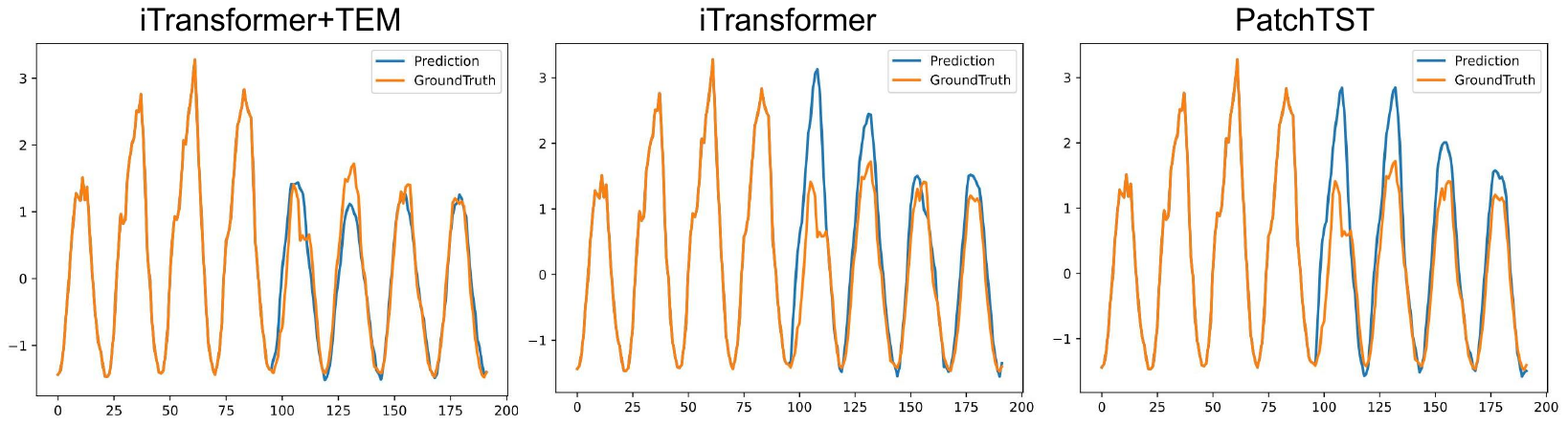}%
    \label{fig:9}}
    \hfil
    \subfloat[]{\includegraphics[width=0.48\textwidth]{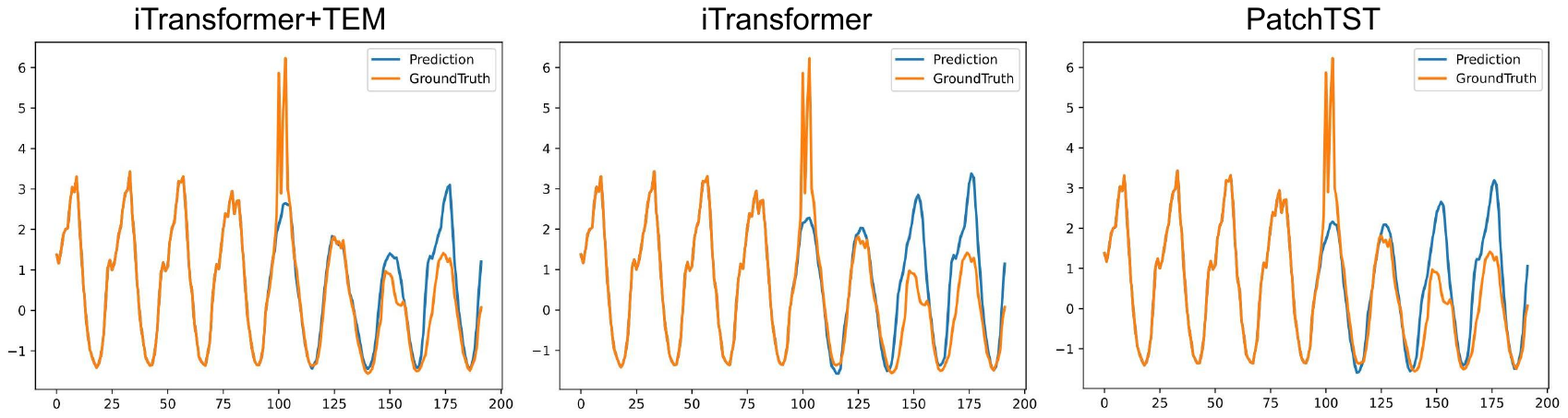}%
    \label{fig:20}}
    \hfil
    \caption{Figures (a) and (b) present two visualization examples of the Traffic dataset on the input-96-predict-96 task.}
    \label{fig_visual1}
\end{figure*}

\begin{figure*}
    \centering
    \subfloat[]{\includegraphics[width=0.48\textwidth]{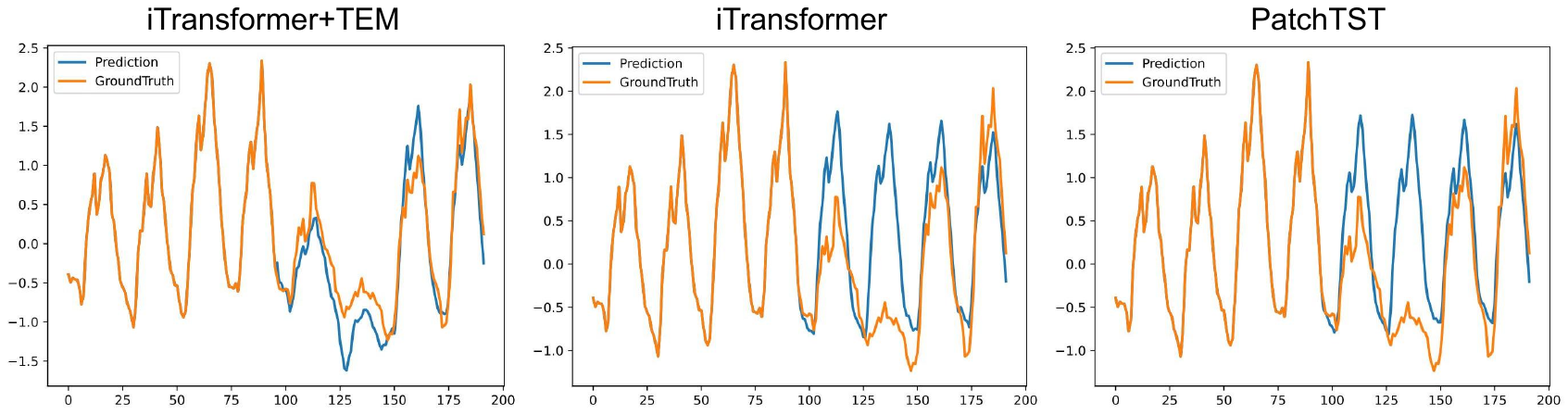}%
    \label{fig:10}}
    \hfil
    \subfloat[]{\includegraphics[width=0.48\textwidth]{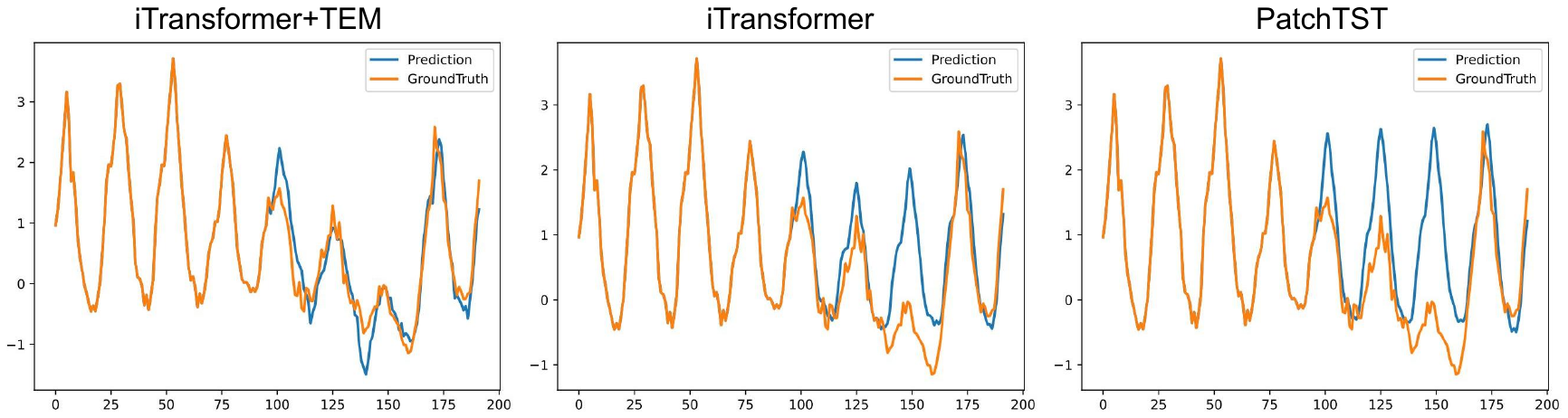}%
    \label{fig:18}}
    \hfil
    \caption{Figures (a) and (b) present two visualization examples of the ECL dataset on the input-96-predict-96 task.}
    \label{fig_visual2}
\end{figure*}

\begin{figure*}
    \centering
    \subfloat[]{\includegraphics[width=0.48\textwidth]{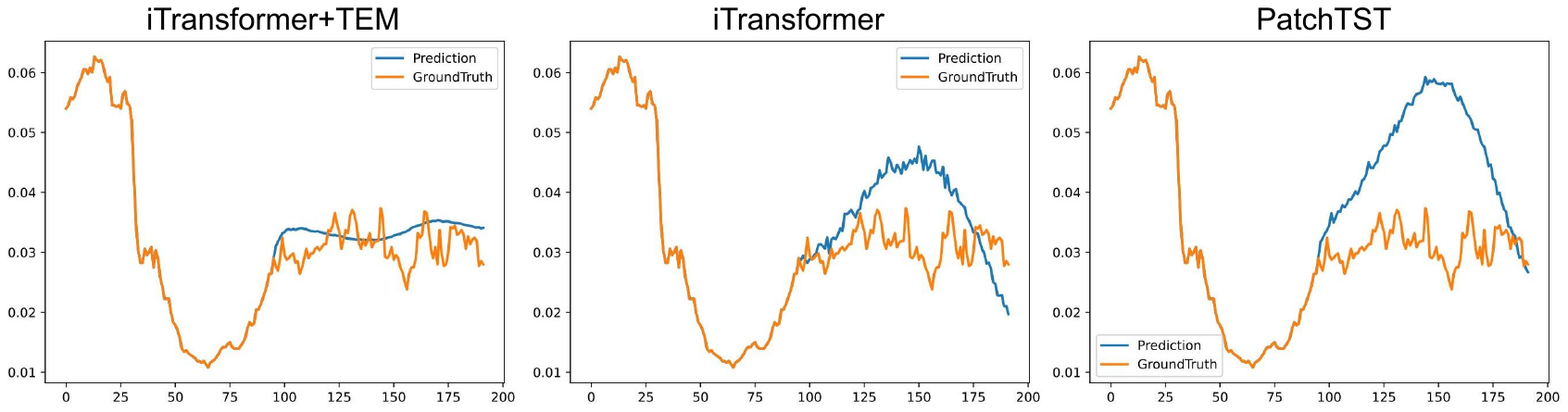}%
    \label{fig:11}}
    \hfil
    \subfloat[]{\includegraphics[width=0.48\textwidth]{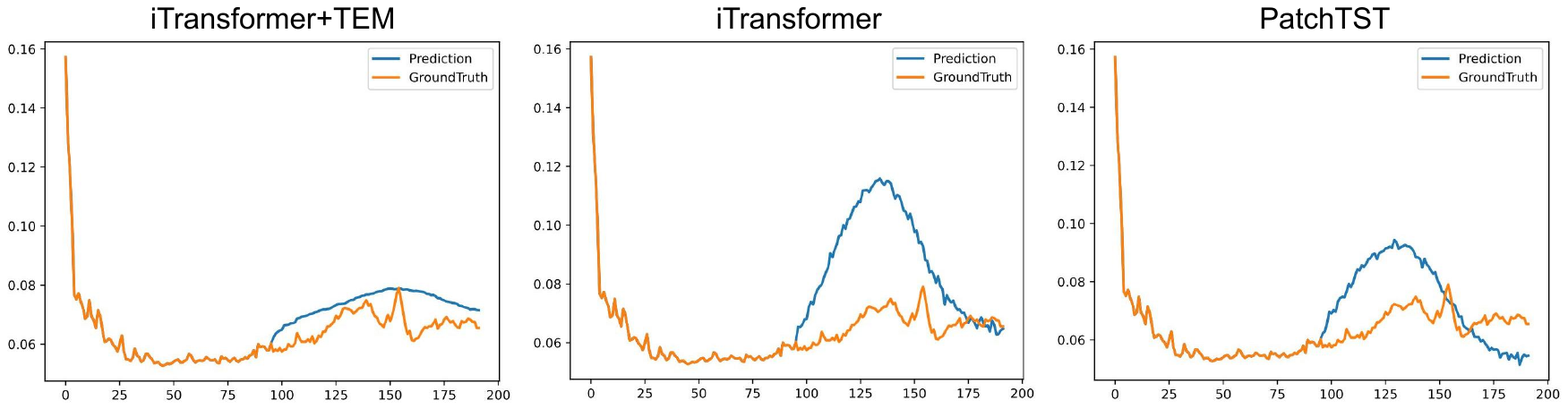}%
    \label{fig:19}}
    \hfil
    \caption{Figures (a) and (b) present two visualization examples of the Weather dataset on the input-96-predict-96 task.}
    \label{fig_visual3}
\end{figure*}

\begin{figure*}
    \centering

    \subfloat[]{\includegraphics[width=0.44\textwidth]{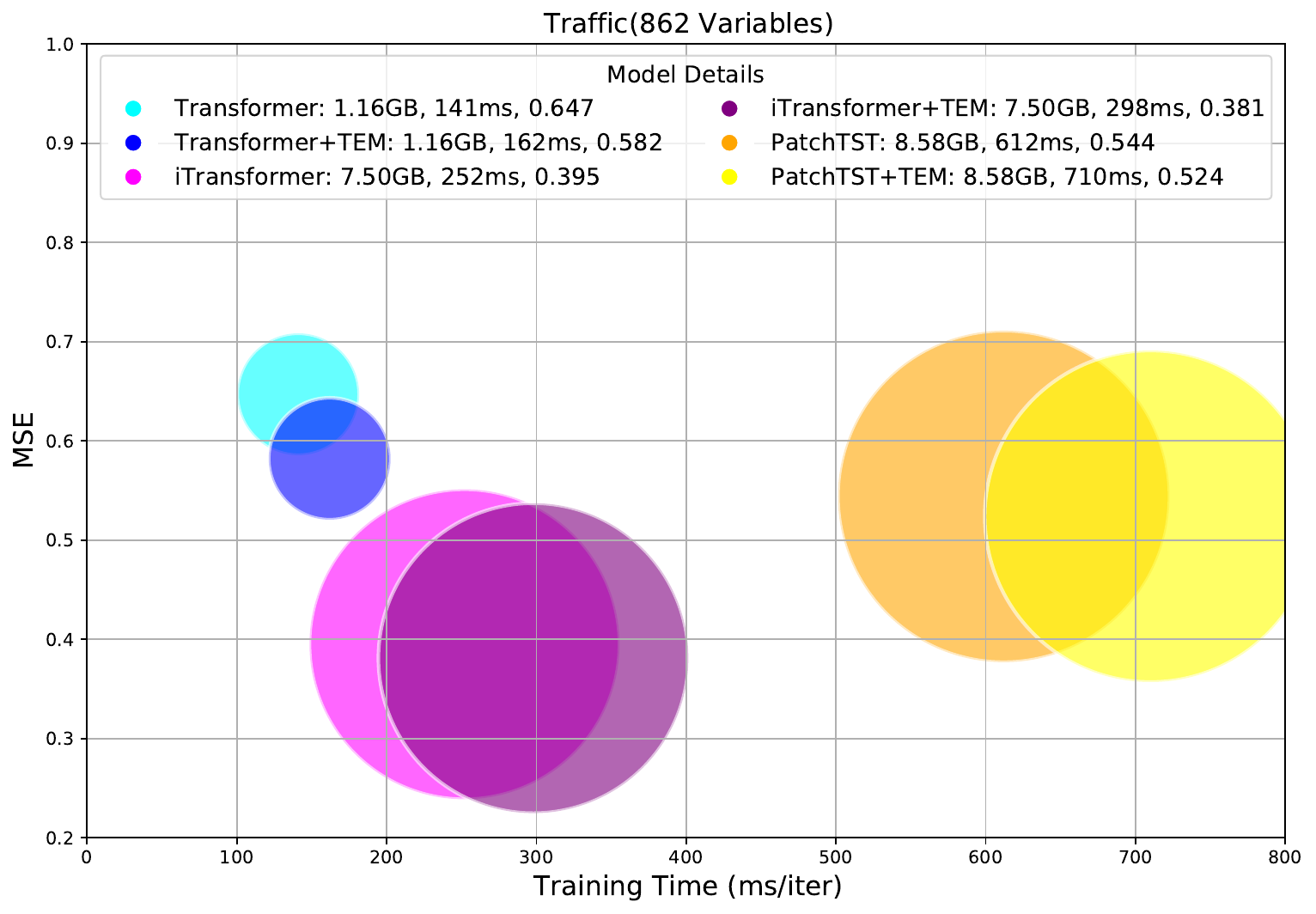}%
    \label{fig_eff_tra}}
    \hfil
    \subfloat[]{\includegraphics[width=0.45\textwidth]{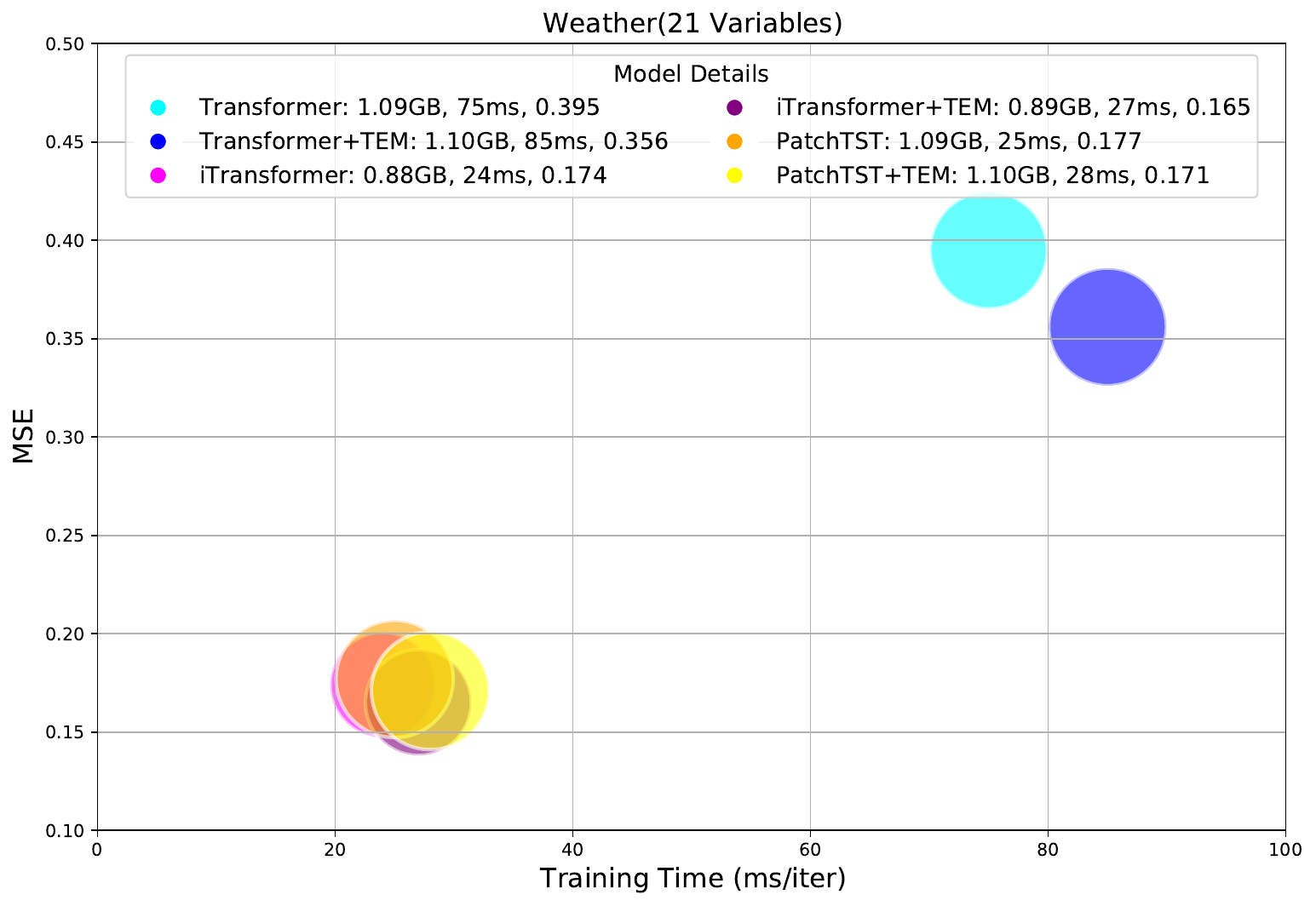}%
    \label{fig_eff_wea}}
    \hfil
    \caption{(a) Computational complexity analysis on the Traffic dataset. (b) Computational complexity analysis on the Weather dataset. Each item in the Model Details of the figure, from left to right, is: model name, GPU memory used, training speed, and forecasting MSE.}
    \label{fig:something2}
\end{figure*}

\subsection{Comparative Experimental Results}
\label{sec_all_res}
Table \ref{tab:mse_std} and Table \ref{tab:mae_std} present the comprehensive prediction results, where the best performance is highlighted in red and the second-best in blue. Lower MSE/MAE values indicate higher prediction accuracy. Compared to other time series forecasting (TSF) models, incorporating our method into various baseline models consistently improves performance. These experimental results demonstrate that our approach is applicable to different types of tokens, thereby validating the effectiveness of the proposed method.

\subsection{Ablation Study}
\label{sec_ablation}
This subsection conducts ablation studies to systematically evaluate the contribution of the PTEM and STEM designs in the proposed method to model performance. We first analyze the performance changes after removing PTEM and STEM from the model to verify the effectiveness of each module. Next, we inject quantitative OPT and OST into each layer of the model and assess their impact on performance, thereby demonstrating the importance of the adaptive injection mechanisms in PTEM and STEM. Finally, we perform ablation studies on the adopted bi-level optimization strategy to verify its rationality and contribution.

\paragraph{Ablation Study of PTEM, STEM}
In this section, we conduct a systematic ablation study on the effects of the two proposed components, PTEM and STEM, across different models. Specifically, we progressively introduce STEM, PTEM, and their combination (referred to as TEM) into three representative models: PatchTST, iTransformer, and the vanilla Transformer. Experiments are performed on the ETTm1, Weather, and ECL datasets, which contain 7, 21, and 321 variables, respectively, representing three typical scenarios with small, medium, and large numbers of variables. The complete experimental results are shown in Tables \ref{tab:ablation_patchtst}, \ref{tab:ablation_itransformer}, and \ref{tab:ablation_transformer}. The results demonstrate that incorporating PTEM and STEM consistently leads to stable and significant performance improvements across all three models. Taking MSE as an example, introducing STEM alone yields an average 2.32\% reduction in error, while PTEM alone brings an average 4.48\% reduction in error. When combining both modules (+TEM), the overall average MSE reduction reaches 5.81\%, achieving the best performance compared to the baseline models. These findings indicate that PTEM and STEM each contribute independently and significantly to model performance, and their joint usage further enhances the model’s ability to capture temporal structures and improve generalization.

\paragraph{Ablation Study of Adaptive Injection Mechanisms}
In Eq.(\ref{eq_GNEM}) and (\ref{eq_SNEM}), we introduce adaptive parameters $\Gamma$ and $\Xi$ to flexibly control the amount of OPT and OST injected into the features. To evaluate the importance of these adaptive mechanisms, we train and evaluate our model by setting $\Gamma$ and $\Xi$ to different fixed values. The experiments are conducted on three datasets: ETTm1, Weather, and ECL. The experiments are also conducted on three model configurations: PatchTST+TEM, iTransformer+TEM, and Transformer+TEM. The experimental results are shown in Figure \ref{fig_adp_p}, \ref{fig_adp_i}, \ref{fig_adp_t}. The results show that injecting a fixed amount of OPT and OST into each layer of the network leads to unstable and inferior performance compared to the adaptive injection strategy. This demonstrates the effectiveness and rationality of the adaptive design in PTEM and STEM.

\paragraph{Ablation Study of Bi-level Optimization Strategy}
To ensure the accuracy of OPT/OST injection, we propose a bi-level optimization mechanism that jointly optimizes the TSF model ($f_\theta$) and the parameters of PTEM and STEM modules ($\Gamma, \Xi$). To verify the rationality of this design, we replace the bi-level optimization with a regular joint optimization strategy, where the TSF model and the topology injection weights are updated simultaneously. We then train and evaluate the model on the ETTm1, Weather, and ECL datasets. In these experiments, both the input and prediction lengths are fixed at 96. The experiments are also conducted on three model configurations: PatchTST+TEM, iTransformer+TEM, and Transformer+TEM. The mean squared errors (MSE) and training times across the datasets are shown in Figure \ref{fig_bi_level}. The experimental results show that the bi-level optimization strategy significantly outperforms the regular joint optimization strategy in terms of prediction performance, while incurring only a slight increase in training time (less than 20\%). In particular, for model configurations with low training costs (e.g., ETTm1-PatchTST and ETTm1-iTransformer), bi-level optimization achieves notable performance improvements with almost no additional time overhead, as illustrated by the box icons in Figures \ref{fig_bi_level1} and \ref{fig_bi_level2}. These results further validate the rationality and effectiveness of the proposed design.

\subsection{Hyperparameter Sensitivity}
\label{Hyperparameter_sense}
The only hyperparameter involved in this study is the initial learning rate of the outer loop ($\eta_2$). All other hyperparameters are inherited from the original settings of other baseline algorithms. In this section, we conduct a hyperparameter sensitivity analysis to determine the optimal value of the initial $\eta_2$. The experiments are carried out on the ETTh1, Weather, and ECL datasets. As shown in Figure \ref{fig_eta}, the optimal value for the initial $\eta_2$ is found to be $1e-3$.

\subsection{Further Experiments Based on Section \ref{motivation}}
\label{Sec_fem}
To demonstrate the superiority of the adaptive PTEM and STEM approaches over the quantitative positional/semantic topology enhancement method used in Section \ref{motivation}, we integrate PTEM and STEM into the same three models (Transformer, PatchTST, iTransformer) used in that section. These models are then trained and evaluated under identical experimental settings. As illustrated in Figure \ref{fig_moti_more} and Figure \ref{fig_moti_more1}, employing either PTEM or STEM consistently yields significant performance gains across all datasets, with improvements exceeding those achieved by +EPT and +EST.

\subsection{Comparison with Other Positional Encodings (PEs)}
\label{compare_PE}
Our approach is related to several studies on PE that likewise reinforce positional topology at deeper layers. For example, \cite{su2024roformer} introduces Rotary Position Embedding (RoPE) to improve Transformer performance on natural language processing tasks; \cite{chu2021conditional} proposes Convolutional Positional Encoding (CPE) to enhance Transformers for vision tasks; and \cite{lv2025toward} presents GL-PE, which strengthens TSF by injecting fixed relative positional encodings at every layer. Note that these methods do not provide fine-grained control over the topological quantities injected into each layer, whereas our method does. In this subsection, we apply these three PEs and our proposed TEM to the same baseline method to compare their effects. In all experiments, the look-back length and the prediction length are both fixed at 96. Results are shown in Figure \ref{fig_bar}. The experiments indicate that when these PEs are directly applied to Transformer-based TSF models, many methods yield limited gains and can even have negative effects in certain cases. We hypothesize that, due to the lack of fine-grained control over the amount of topology injected across layers, these three methods may introduce either too little or too much topological information, thereby weakening the model’s predictive capability. In contrast, TEM employs an adaptive injection mechanism and introduces semantic topology enhancement, resulting in more stable and substantially larger improvements. These results further validate the effectiveness of our approach.

\subsection{Computational Complexity Analysis}
\label{Complexity}
We conduct a comprehensive comparison of several models in terms of forecasting performance, training speed, and GPU memory usage. The experiments are conducted on the Weather and Traffic datasets, with both the input (look-back) length and prediction length set to 96. All models are implemented using their official configurations, with a batch size of 16. All experiments are performed on a single NVIDIA 4090 GPU with 24GB of GPU memory. The results are shown in Figures \ref{fig_eff_tra} and \ref{fig_eff_wea}. From the figures, we can observe the following: integrating our TEM into other Transformer-based methods introduces almost no increase in GPU memory usage, results in a slight increase in training time, and significantly improves forecasting performance.

\subsection{Visualization of Prediction}
\label{visual_res}
To clearly compare the performance of different models, we present prediction visualizations for three representative datasets in Figures \ref{fig_visual1}, \ref{fig_visual2}, and \ref{fig_visual3}. These results are generated using the following methods: our iTransformer+TEM, iTransformer \cite{liu2023itransformer}, and PatchTST \cite{PatchTST}. Among the various models, our method predicts future sequence variations with the highest accuracy and demonstrates superior performance.

\section{Conclusion}
This study empirically and theoretically reveals that as the depth of Transformer architectures increases, the original positional and semantic topological structures among input tokens tend to be progressively weakened, thereby limiting forecasting accuracy in time series tasks. To address this issue, we propose the Topology Enhancement Method (TEM), which introduces two key modules into Transformers: the Positional Topology Enhancement Module (PTEM) and the Semantic Topology Enhancement Module (STEM). These modules respectively preserve the original positional and semantic topology of input tokens. Both modules are equipped with learnable mechanisms and employ a bi-level optimization strategy to adaptively determine the optimal injection weights. As a plug-and-play framework, TEM can be seamlessly integrated into many mainstream Transformer-based models. Extensive experiments on multiple benchmark datasets demonstrate the superior performance of TEM.

\section*{CRediT authorship contribution statement}
\noindent \textbf{Jianqi Zhang}: Conceptualization, Writing – original draft, Experimental analysis; \textbf{Wenwen Qiang}: Conceptualization, Funding acquisition, Supervision, Writing – review \& editing; \textbf{Jingyao Wang}: Theoretical analysis, Writing – original draft, Writing – review \& editing; \textbf{Jiahuan Zhou}: Writing – review \& editing; \textbf{Changwen Zheng}: Funding acquisition, Supervision; \textbf{Hui Xiong}: Writing – review \& editing.

\section*{Declaration of competing interest}
The authors declare that they have no known competing financial interests or personal relationships that could have appeared to influence the work reported in this paper.

\section*{Data Availability}
Our code and dataset are publicly available at: \href{https://github.com/jlu-phyComputer/TEM}{https://github.com/jlu-phyComputer/TEM}.

\section*{Acknowledgements}
This work is supported by the National Natural Science Foundation of China (No. 62506355).

\bibliographystyle{cas-model2-names}

\bibliography{cas-refs}

\begin{thebibliography}{59}
\expandafter\ifx\csname natexlab\endcsname\relax\def\natexlab#1{#1}\fi
\providecommand{\url}[1]{\texttt{#1}}
\providecommand{\href}[2]{#2}
\providecommand{\path}[1]{#1}
\providecommand{\DOIprefix}{doi:}
\providecommand{\ArXivprefix}{arXiv:}
\providecommand{\URLprefix}{URL: }
\providecommand{\Pubmedprefix}{pmid:}
\providecommand{\doi}[1]{\href{http://dx.doi.org/#1}{\path{#1}}}
\providecommand{\Pubmed}[1]{\href{pmid:#1}{\path{#1}}}
\providecommand{\bibinfo}[2]{#2}
\ifx\xfnm\relax \def\xfnm[#1]{\unskip,\space#1}\fi
\bibitem[{Abhishek et~al.(2012)Abhishek, Singh, Ghosh and Anand}]{abhishek2012weather}
\bibinfo{author}{Abhishek, K.}, \bibinfo{author}{Singh, M.}, \bibinfo{author}{Ghosh, S.}, \bibinfo{author}{Anand, A.}, \bibinfo{year}{2012}.
\newblock \bibinfo{title}{Weather forecasting model using artificial neural network}.
\newblock \bibinfo{journal}{Procedia Technology} \bibinfo{volume}{4}, \bibinfo{pages}{311--318}.
\bibitem[{Boussif et~al.(2024)Boussif, Boukachab, Assouline, Massaroli, Yuan, Benabbou and Bengio}]{boussif2024improving}
\bibinfo{author}{Boussif, O.}, \bibinfo{author}{Boukachab, G.}, \bibinfo{author}{Assouline, D.}, \bibinfo{author}{Massaroli, S.}, \bibinfo{author}{Yuan, T.}, \bibinfo{author}{Benabbou, L.}, \bibinfo{author}{Bengio, Y.}, \bibinfo{year}{2024}.
\newblock \bibinfo{title}{Improving* day-ahead* solar irradiance time series forecasting by leveraging spatio-temporal context}.
\newblock \bibinfo{journal}{Advances in Neural Information Processing Systems} \bibinfo{volume}{36}.
\bibitem[{Cai et~al.(2005)Cai, He and Han}]{cai2005document}
\bibinfo{author}{Cai, D.}, \bibinfo{author}{He, X.}, \bibinfo{author}{Han, J.}, \bibinfo{year}{2005}.
\newblock \bibinfo{title}{Document clustering using locality preserving indexing}.
\newblock \bibinfo{journal}{IEEE transactions on knowledge and data engineering} \bibinfo{volume}{17}, \bibinfo{pages}{1624--1637}.
\bibitem[{Chu et~al.(2021)Chu, Tian, Zhang, Wang, Wei, Xia and Shen}]{chu2021conditional}
\bibinfo{author}{Chu, X.}, \bibinfo{author}{Tian, Z.}, \bibinfo{author}{Zhang, B.}, \bibinfo{author}{Wang, X.}, \bibinfo{author}{Wei, X.}, \bibinfo{author}{Xia, H.}, \bibinfo{author}{Shen, C.}, \bibinfo{year}{2021}.
\newblock \bibinfo{title}{Conditional positional encodings for vision transformers}.
\newblock \bibinfo{journal}{arXiv preprint arXiv:2102.10882} .
\bibitem[{Cover(1999)}]{cover1999elements}
\bibinfo{author}{Cover, T.M.}, \bibinfo{year}{1999}.
\newblock \bibinfo{title}{Elements of information theory}.
\newblock \bibinfo{publisher}{John Wiley \& Sons}.
\bibitem[{Dai et~al.(2025)Dai, Wang, Jie, Wang and Ye}]{dai2025vtformer}
\bibinfo{author}{Dai, R.}, \bibinfo{author}{Wang, Z.}, \bibinfo{author}{Jie, J.}, \bibinfo{author}{Wang, W.}, \bibinfo{author}{Ye, Q.}, \bibinfo{year}{2025}.
\newblock \bibinfo{title}{Vtformer: a novel multiscale linear transformer forecaster with variate-temporal dependency for multivariate time series}.
\newblock \bibinfo{journal}{Complex \& Intelligent Systems} \bibinfo{volume}{11}, \bibinfo{pages}{1--19}.
\bibitem[{Das et~al.(2023)Das, Kong, Leach, Sen and Yu}]{das2023long}
\bibinfo{author}{Das, A.}, \bibinfo{author}{Kong, W.}, \bibinfo{author}{Leach, A.}, \bibinfo{author}{Sen, R.}, \bibinfo{author}{Yu, R.}, \bibinfo{year}{2023}.
\newblock \bibinfo{title}{Long-term forecasting with tide: Time-series dense encoder}.
\newblock \bibinfo{journal}{arXiv preprint arXiv:2304.08424} .
\bibitem[{Devlin et~al.(2018)Devlin, Chang, Lee and Toutanova}]{devlin2018bert}
\bibinfo{author}{Devlin, J.}, \bibinfo{author}{Chang, M.W.}, \bibinfo{author}{Lee, K.}, \bibinfo{author}{Toutanova, K.}, \bibinfo{year}{2018}.
\newblock \bibinfo{title}{Bert: Pre-training of deep bidirectional transformers for language understanding}.
\newblock \bibinfo{journal}{arXiv preprint arXiv:1810.04805} .
\bibitem[{Du et~al.(2023)Du, Su and Wei}]{du2023preformer}
\bibinfo{author}{Du, D.}, \bibinfo{author}{Su, B.}, \bibinfo{author}{Wei, Z.}, \bibinfo{year}{2023}.
\newblock \bibinfo{title}{Preformer: predictive transformer with multi-scale segment-wise correlations for long-term time series forecasting}, in: \bibinfo{booktitle}{ICASSP 2023-2023 IEEE International Conference on Acoustics, Speech and Signal Processing (ICASSP)}, \bibinfo{organization}{IEEE}. pp. \bibinfo{pages}{1--5}.
\bibitem[{Fang et~al.(2023)Fang, Qin, Luo, Zhao and Zheng}]{fang2023stwave+}
\bibinfo{author}{Fang, Y.}, \bibinfo{author}{Qin, Y.}, \bibinfo{author}{Luo, H.}, \bibinfo{author}{Zhao, F.}, \bibinfo{author}{Zheng, K.}, \bibinfo{year}{2023}.
\newblock \bibinfo{title}{Stwave+: A multi-scale efficient spectral graph attention network with long-term trends for disentangled traffic flow forecasting}.
\newblock \bibinfo{journal}{IEEE Transactions on Knowledge and Data Engineering} .
\bibitem[{Gangeh et~al.(2017)Gangeh, Zarkoob and Ghodsi}]{gangeh2017fast}
\bibinfo{author}{Gangeh, M.J.}, \bibinfo{author}{Zarkoob, H.}, \bibinfo{author}{Ghodsi, A.}, \bibinfo{year}{2017}.
\newblock \bibinfo{title}{Fast and scalable feature selection for gene expression data using hilbert-schmidt independence criterion}.
\newblock \bibinfo{journal}{IEEE/ACM transactions on computational biology and bioinformatics} \bibinfo{volume}{14}, \bibinfo{pages}{167--181}.
\bibitem[{Golowich et~al.(2018)Golowich, Rakhlin and Shamir}]{golowich2018size}
\bibinfo{author}{Golowich, N.}, \bibinfo{author}{Rakhlin, A.}, \bibinfo{author}{Shamir, O.}, \bibinfo{year}{2018}.
\newblock \bibinfo{title}{Size-independent sample complexity of neural networks}, in: \bibinfo{booktitle}{Conference on Learning Theory (COLT)}, \bibinfo{publisher}{PMLR}. pp. \bibinfo{pages}{297--299}.
\bibitem[{Greenfeld and Shalit(2020)}]{greenfeld2020robust}
\bibinfo{author}{Greenfeld, D.}, \bibinfo{author}{Shalit, U.}, \bibinfo{year}{2020}.
\newblock \bibinfo{title}{Robust learning with the hilbert-schmidt independence criterion}, in: \bibinfo{booktitle}{International Conference on Machine Learning}, \bibinfo{organization}{PMLR}. pp. \bibinfo{pages}{3759--3768}.
\bibitem[{Han et~al.(2024)Han, Ye and Zhan}]{han2024capacity}
\bibinfo{author}{Han, L.}, \bibinfo{author}{Ye, H.J.}, \bibinfo{author}{Zhan, D.C.}, \bibinfo{year}{2024}.
\newblock \bibinfo{title}{The capacity and robustness trade-off: Revisiting the channel independent strategy for multivariate time series forecasting}.
\newblock \bibinfo{journal}{IEEE Transactions on Knowledge and Data Engineering} .
\bibitem[{He and Niyogi(2003)}]{he2003locality}
\bibinfo{author}{He, X.}, \bibinfo{author}{Niyogi, P.}, \bibinfo{year}{2003}.
\newblock \bibinfo{title}{Locality preserving projections}.
\newblock \bibinfo{journal}{Advances in neural information processing systems} \bibinfo{volume}{16}.
\bibitem[{He et~al.(2005)He, Yan, Hu, Niyogi and Zhang}]{he2005face}
\bibinfo{author}{He, X.}, \bibinfo{author}{Yan, S.}, \bibinfo{author}{Hu, Y.}, \bibinfo{author}{Niyogi, P.}, \bibinfo{author}{Zhang, H.J.}, \bibinfo{year}{2005}.
\newblock \bibinfo{title}{Face recognition using laplacianfaces}.
\newblock \bibinfo{journal}{IEEE transactions on pattern analysis and machine intelligence} \bibinfo{volume}{27}, \bibinfo{pages}{328--340}.
\bibitem[{Huang et~al.(2023)Huang, Ma, Dai, Hu and Du}]{huang2023dbaformer}
\bibinfo{author}{Huang, J.}, \bibinfo{author}{Ma, M.}, \bibinfo{author}{Dai, Y.}, \bibinfo{author}{Hu, J.}, \bibinfo{author}{Du, S.}, \bibinfo{year}{2023}.
\newblock \bibinfo{title}{Dbaformer: A double-branch attention transformer for long-term time series forecasting}.
\newblock \bibinfo{journal}{Human-Centric Intelligent Systems} \bibinfo{volume}{3}, \bibinfo{pages}{263--274}.
\bibitem[{Irani and Metsis(2025a)}]{irani2025dywpe}
\bibinfo{author}{Irani, H.}, \bibinfo{author}{Metsis, V.}, \bibinfo{year}{2025}a.
\newblock \bibinfo{title}{Dywpe: Signal-aware dynamic wavelet positional encoding for time series transformers}.
\newblock \bibinfo{journal}{arXiv preprint arXiv:2509.14640} .
\bibitem[{Irani and Metsis(2025b)}]{irani2025positional}
\bibinfo{author}{Irani, H.}, \bibinfo{author}{Metsis, V.}, \bibinfo{year}{2025}b.
\newblock \bibinfo{title}{Positional encoding in transformer-based time series models: a survey}.
\newblock \bibinfo{journal}{arXiv preprint arXiv:2502.12370} .
\bibitem[{Karevan and Suykens(2020)}]{karevan2020transductive}
\bibinfo{author}{Karevan, Z.}, \bibinfo{author}{Suykens, J.A.}, \bibinfo{year}{2020}.
\newblock \bibinfo{title}{Transductive lstm for time-series prediction: An application to weather forecasting}.
\newblock \bibinfo{journal}{Neural Networks} \bibinfo{volume}{125}, \bibinfo{pages}{1--9}.
\bibitem[{Ke et~al.(2020)Ke, He and Liu}]{ke2020rethinking}
\bibinfo{author}{Ke, G.}, \bibinfo{author}{He, D.}, \bibinfo{author}{Liu, T.Y.}, \bibinfo{year}{2020}.
\newblock \bibinfo{title}{Rethinking positional encoding in language pre-training}.
\newblock \bibinfo{journal}{arXiv preprint arXiv:2006.15595} .
\bibitem[{Kingma and Ba(2015)}]{Adam}
\bibinfo{author}{Kingma, D.P.}, \bibinfo{author}{Ba, J.}, \bibinfo{year}{2015}.
\newblock \bibinfo{title}{Adam: {A} method for stochastic optimization}.
\newblock \bibinfo{journal}{ICLR} .
\bibitem[{Lai et~al.(2018)Lai, Chang, Yang and Liu}]{LSTNet}
\bibinfo{author}{Lai, G.}, \bibinfo{author}{Chang, W.C.}, \bibinfo{author}{Yang, Y.}, \bibinfo{author}{Liu, H.}, \bibinfo{year}{2018}.
\newblock \bibinfo{title}{Modeling long-and short-term temporal patterns with deep neural networks}.
\newblock \bibinfo{journal}{SIGIR} .
\bibitem[{Li et~al.(2022)Li, Zhang, Li, Zhang and Zhang}]{li2022dmgan}
\bibinfo{author}{Li, R.}, \bibinfo{author}{Zhang, F.}, \bibinfo{author}{Li, T.}, \bibinfo{author}{Zhang, N.}, \bibinfo{author}{Zhang, T.}, \bibinfo{year}{2022}.
\newblock \bibinfo{title}{Dmgan: Dynamic multi-hop graph attention network for traffic forecasting}.
\newblock \bibinfo{journal}{IEEE Transactions on Knowledge and Data Engineering} .
\bibitem[{Li et~al.(2024)Li, Yu, Li and Zhu}]{li2024functional}
\bibinfo{author}{Li, T.}, \bibinfo{author}{Yu, B.}, \bibinfo{author}{Li, J.}, \bibinfo{author}{Zhu, Z.}, \bibinfo{year}{2024}.
\newblock \bibinfo{title}{Functional relation field: A model-agnostic framework for multivariate time series forecasting}.
\newblock \bibinfo{journal}{Artificial Intelligence} \bibinfo{volume}{334}, \bibinfo{pages}{104158}.
\bibitem[{Liang et~al.(2023)Liang, Zhang, Yuan, Ma, Li and Zhang}]{liang2023does}
\bibinfo{author}{Liang, D.}, \bibinfo{author}{Zhang, H.}, \bibinfo{author}{Yuan, D.}, \bibinfo{author}{Ma, X.}, \bibinfo{author}{Li, D.}, \bibinfo{author}{Zhang, M.}, \bibinfo{year}{2023}.
\newblock \bibinfo{title}{Does long-term series forecasting need complex attention and extra long inputs?}
\newblock \bibinfo{journal}{arXiv preprint arXiv:2306.05035} .
\bibitem[{Lim et~al.(2021)Lim, Ar{\i}k, Loeff and Pfister}]{lim2021temporal}
\bibinfo{author}{Lim, B.}, \bibinfo{author}{Ar{\i}k, S.{\"O}.}, \bibinfo{author}{Loeff, N.}, \bibinfo{author}{Pfister, T.}, \bibinfo{year}{2021}.
\newblock \bibinfo{title}{Temporal fusion transformers for interpretable multi-horizon time series forecasting}.
\newblock \bibinfo{journal}{International Journal of Forecasting} \bibinfo{volume}{37}, \bibinfo{pages}{1748--1764}.
\bibitem[{Lin et~al.(2024)Lin, Lin, Wu, Chen and Yang}]{lin2024sparsetsf}
\bibinfo{author}{Lin, S.}, \bibinfo{author}{Lin, W.}, \bibinfo{author}{Wu, W.}, \bibinfo{author}{Chen, H.}, \bibinfo{author}{Yang, J.}, \bibinfo{year}{2024}.
\newblock \bibinfo{title}{Sparsetsf: Modeling long-term time series forecasting with 1k parameters}.
\newblock \bibinfo{journal}{arXiv preprint arXiv:2405.00946} .
\bibitem[{Liu et~al.(2022)Liu, Zeng, Chen, Xu, Lai, Ma and Xu}]{liu2022scinet}
\bibinfo{author}{Liu, M.}, \bibinfo{author}{Zeng, A.}, \bibinfo{author}{Chen, M.}, \bibinfo{author}{Xu, Z.}, \bibinfo{author}{Lai, Q.}, \bibinfo{author}{Ma, L.}, \bibinfo{author}{Xu, Q.}, \bibinfo{year}{2022}.
\newblock \bibinfo{title}{Scinet: Time series modeling and forecasting with sample convolution and interaction}.
\newblock \bibinfo{journal}{Advances in Neural Information Processing Systems} \bibinfo{volume}{35}, \bibinfo{pages}{5816--5828}.
\bibitem[{Liu et~al.(2023)Liu, Hu, Zhang, Wu, Wang, Ma and Long}]{liu2023itransformer}
\bibinfo{author}{Liu, Y.}, \bibinfo{author}{Hu, T.}, \bibinfo{author}{Zhang, H.}, \bibinfo{author}{Wu, H.}, \bibinfo{author}{Wang, S.}, \bibinfo{author}{Ma, L.}, \bibinfo{author}{Long, M.}, \bibinfo{year}{2023}.
\newblock \bibinfo{title}{itransformer: Inverted transformers are effective for time series forecasting}.
\newblock \bibinfo{journal}{arXiv preprint arXiv:2310.06625} .
\bibitem[{Liu et~al.(2021)Liu, Lin, Cao, Hu, Wei, Zhang, Lin and Guo}]{liu2021swin}
\bibinfo{author}{Liu, Z.}, \bibinfo{author}{Lin, Y.}, \bibinfo{author}{Cao, Y.}, \bibinfo{author}{Hu, H.}, \bibinfo{author}{Wei, Y.}, \bibinfo{author}{Zhang, Z.}, \bibinfo{author}{Lin, S.}, \bibinfo{author}{Guo, B.}, \bibinfo{year}{2021}.
\newblock \bibinfo{title}{Swin transformer: Hierarchical vision transformer using shifted windows}, in: \bibinfo{booktitle}{Proceedings of the IEEE/CVF international conference on computer vision}, pp. \bibinfo{pages}{10012--10022}.
\bibitem[{Lv et~al.(2025)Lv, Wang, Han, Shen, Zheng, Huang and Li}]{lv2025toward}
\bibinfo{author}{Lv, C.}, \bibinfo{author}{Wang, Y.}, \bibinfo{author}{Han, D.}, \bibinfo{author}{Shen, Y.}, \bibinfo{author}{Zheng, X.}, \bibinfo{author}{Huang, X.}, \bibinfo{author}{Li, D.}, \bibinfo{year}{2025}.
\newblock \bibinfo{title}{Toward relative positional encoding in spiking transformers}.
\newblock \bibinfo{journal}{arXiv preprint arXiv:2501.16745} .
\bibitem[{Messou et~al.(2025)Messou, Chen, Liu, Zhang and Yu}]{messou2025tsformer}
\bibinfo{author}{Messou, F.J.A.}, \bibinfo{author}{Chen, J.}, \bibinfo{author}{Liu, T.}, \bibinfo{author}{Zhang, S.}, \bibinfo{author}{Yu, K.}, \bibinfo{year}{2025}.
\newblock \bibinfo{title}{Tsformer: Temporal-aware transformer for multi-horizon forecasting with learnable positional encodings and attention mechanisms}, in: \bibinfo{booktitle}{2025 Sixteenth International Conference on Ubiquitous and Future Networks (ICUFN)}, \bibinfo{organization}{IEEE}. pp. \bibinfo{pages}{600--605}.
\bibitem[{Mohri(2018)}]{mohri2018foundations}
\bibinfo{author}{Mohri, M.}, \bibinfo{year}{2018}.
\newblock \bibinfo{title}{Foundations of machine learning}.
\bibitem[{Mohri and Rostamizadeh(2008)}]{mohri2008rademacher}
\bibinfo{author}{Mohri, M.}, \bibinfo{author}{Rostamizadeh, A.}, \bibinfo{year}{2008}.
\newblock \bibinfo{title}{Rademacher complexity bounds for non-iid processes}.
\newblock \bibinfo{journal}{Advances in neural information processing systems} \bibinfo{volume}{21}.
\bibitem[{Neyshabur et~al.(2018)Neyshabur, Bhojanapalli, McAllester and Srebro}]{neyshabur2018role}
\bibinfo{author}{Neyshabur, B.}, \bibinfo{author}{Bhojanapalli, S.}, \bibinfo{author}{McAllester, D.}, \bibinfo{author}{Srebro, N.}, \bibinfo{year}{2018}.
\newblock \bibinfo{title}{The role of over-parametrization in generalization of neural networks}, in: \bibinfo{booktitle}{International Conference on Learning Representations (ICLR)}.
\bibitem[{Nie et~al.(2023)Nie, Nguyen, Sinthong and Kalagnanam}]{PatchTST}
\bibinfo{author}{Nie, Y.}, \bibinfo{author}{Nguyen, N.H.}, \bibinfo{author}{Sinthong, P.}, \bibinfo{author}{Kalagnanam, J.}, \bibinfo{year}{2023}.
\newblock \bibinfo{title}{A time series is worth 64 words: Long-term forecasting with transformers}.
\newblock \bibinfo{journal}{ICLR} .
\bibitem[{Novo et~al.(2022)Novo, Marocco, Giorgi, Lanzini, Santarelli and Mattiazzo}]{novo2022planning}
\bibinfo{author}{Novo, R.}, \bibinfo{author}{Marocco, P.}, \bibinfo{author}{Giorgi, G.}, \bibinfo{author}{Lanzini, A.}, \bibinfo{author}{Santarelli, M.}, \bibinfo{author}{Mattiazzo, G.}, \bibinfo{year}{2022}.
\newblock \bibinfo{title}{Planning the decarbonisation of energy systems: The importance of applying time series clustering to long-term models}.
\newblock \bibinfo{journal}{Energy Conversion and Management: X} \bibinfo{volume}{15}, \bibinfo{pages}{100274}.
\bibitem[{P{\'e}rez-Suay and Camps-Valls(2018)}]{perez2018sensitivity}
\bibinfo{author}{P{\'e}rez-Suay, A.}, \bibinfo{author}{Camps-Valls, G.}, \bibinfo{year}{2018}.
\newblock \bibinfo{title}{Sensitivity maps of the hilbert--schmidt independence criterion}.
\newblock \bibinfo{journal}{Applied Soft Computing} \bibinfo{volume}{70}, \bibinfo{pages}{1054--1063}.
\bibitem[{Qiang et~al.(2021)Qiang, Li, Zheng, Su and Xiong}]{qiang2021robust}
\bibinfo{author}{Qiang, W.}, \bibinfo{author}{Li, J.}, \bibinfo{author}{Zheng, C.}, \bibinfo{author}{Su, B.}, \bibinfo{author}{Xiong, H.}, \bibinfo{year}{2021}.
\newblock \bibinfo{title}{Robust local preserving and global aligning network for adversarial domain adaptation}.
\newblock \bibinfo{journal}{IEEE Transactions on Knowledge and Data Engineering} \bibinfo{volume}{35}, \bibinfo{pages}{3014--3029}.
\bibitem[{Sbrana et~al.(2020)Sbrana, Rossi and Naldi}]{sbrana2020n}
\bibinfo{author}{Sbrana, A.}, \bibinfo{author}{Rossi, A.L.D.}, \bibinfo{author}{Naldi, M.C.}, \bibinfo{year}{2020}.
\newblock \bibinfo{title}{N-beats-rnn: deep learning for time series forecasting}, in: \bibinfo{booktitle}{2020 19th IEEE International Conference on Machine Learning and Applications (ICMLA)}, \bibinfo{organization}{IEEE}. pp. \bibinfo{pages}{765--768}.
\bibitem[{Shaw et~al.(2018)Shaw, Uszkoreit and Vaswani}]{shaw2018self}
\bibinfo{author}{Shaw, P.}, \bibinfo{author}{Uszkoreit, J.}, \bibinfo{author}{Vaswani, A.}, \bibinfo{year}{2018}.
\newblock \bibinfo{title}{Self-attention with relative position representations}.
\newblock \bibinfo{journal}{arXiv preprint arXiv:1803.02155} .
\bibitem[{Su et~al.(2024)Su, Ahmed, Lu, Pan, Bo and Liu}]{su2024roformer}
\bibinfo{author}{Su, J.}, \bibinfo{author}{Ahmed, M.}, \bibinfo{author}{Lu, Y.}, \bibinfo{author}{Pan, S.}, \bibinfo{author}{Bo, W.}, \bibinfo{author}{Liu, Y.}, \bibinfo{year}{2024}.
\newblock \bibinfo{title}{Roformer: Enhanced transformer with rotary position embedding}.
\newblock \bibinfo{journal}{Neurocomputing} \bibinfo{volume}{568}, \bibinfo{pages}{127063}.
\bibitem[{Tang and Zhang(2023)}]{tang2023infomaxformer}
\bibinfo{author}{Tang, P.}, \bibinfo{author}{Zhang, X.}, \bibinfo{year}{2023}.
\newblock \bibinfo{title}{Infomaxformer: Maximum entropy transformer for long time-series forecasting problem}.
\newblock \bibinfo{journal}{arXiv preprint arXiv:2301.01772} .
\bibitem[{Vaswani et~al.(2017)Vaswani, Shazeer, Parmar, Uszkoreit, Jones, Gomez, Kaiser and Polosukhin}]{Transformer}
\bibinfo{author}{Vaswani, A.}, \bibinfo{author}{Shazeer, N.}, \bibinfo{author}{Parmar, N.}, \bibinfo{author}{Uszkoreit, J.}, \bibinfo{author}{Jones, L.}, \bibinfo{author}{Gomez, A.N.}, \bibinfo{author}{Kaiser, L.}, \bibinfo{author}{Polosukhin, I.}, \bibinfo{year}{2017}.
\newblock \bibinfo{title}{Attention is all you need}.
\newblock \bibinfo{journal}{NeurIPS} .
\bibitem[{Wang et~al.(2022)Wang, Chen, Fan, Zhang, Cai and Song}]{wang2022st}
\bibinfo{author}{Wang, H.}, \bibinfo{author}{Chen, J.}, \bibinfo{author}{Fan, Z.}, \bibinfo{author}{Zhang, Z.}, \bibinfo{author}{Cai, Z.}, \bibinfo{author}{Song, X.}, \bibinfo{year}{2022}.
\newblock \bibinfo{title}{St-expertnet: A deep expert framework for traffic prediction}.
\newblock \bibinfo{journal}{IEEE Transactions on Knowledge and Data Engineering} .
\bibitem[{Wang et~al.(2021)Wang, Dai and Liu}]{wang2021learning}
\bibinfo{author}{Wang, T.}, \bibinfo{author}{Dai, X.}, \bibinfo{author}{Liu, Y.}, \bibinfo{year}{2021}.
\newblock \bibinfo{title}{Learning with hilbert--schmidt independence criterion: A review and new perspectives}.
\newblock \bibinfo{journal}{Knowledge-based systems} \bibinfo{volume}{234}, \bibinfo{pages}{107567}.
\bibitem[{Wang et~al.(2020)Wang, Liu and Yuan}]{wang2020stacked}
\bibinfo{author}{Wang, Y.}, \bibinfo{author}{Liu, C.}, \bibinfo{author}{Yuan, X.}, \bibinfo{year}{2020}.
\newblock \bibinfo{title}{Stacked locality preserving autoencoder for feature extraction and its application for industrial process data modeling}.
\newblock \bibinfo{journal}{Chemometrics and Intelligent Laboratory Systems} \bibinfo{volume}{203}, \bibinfo{pages}{104086}.
\bibitem[{Wu et~al.(2023)Wu, Hu, Liu, Zhou, Wang and Long}]{Timesnet}
\bibinfo{author}{Wu, H.}, \bibinfo{author}{Hu, T.}, \bibinfo{author}{Liu, Y.}, \bibinfo{author}{Zhou, H.}, \bibinfo{author}{Wang, J.}, \bibinfo{author}{Long, M.}, \bibinfo{year}{2023}.
\newblock \bibinfo{title}{Timesnet: Temporal 2d-variation modeling for general time series analysis}.
\newblock \bibinfo{journal}{ICLR} .
\bibitem[{Wu et~al.(2021)Wu, Xu, Wang and Long}]{Autoformer}
\bibinfo{author}{Wu, H.}, \bibinfo{author}{Xu, J.}, \bibinfo{author}{Wang, J.}, \bibinfo{author}{Long, M.}, \bibinfo{year}{2021}.
\newblock \bibinfo{title}{Autoformer: Decomposition transformers with {Auto-Correlation} for long-term series forecasting}.
\newblock \bibinfo{journal}{NeurIPS} .
\bibitem[{Wu et~al.(2020)Wu, Green, Ben and O'Banion}]{wu2020deep}
\bibinfo{author}{Wu, N.}, \bibinfo{author}{Green, B.}, \bibinfo{author}{Ben, X.}, \bibinfo{author}{O'Banion, S.}, \bibinfo{year}{2020}.
\newblock \bibinfo{title}{Deep transformer models for time series forecasting: The influenza prevalence case}.
\newblock \bibinfo{journal}{arXiv preprint arXiv:2001.08317} .
\bibitem[{Yu et~al.(2023)Yu, Wang, Shao, Sun, Wu and Xu}]{yu2023dsformer}
\bibinfo{author}{Yu, C.}, \bibinfo{author}{Wang, F.}, \bibinfo{author}{Shao, Z.}, \bibinfo{author}{Sun, T.}, \bibinfo{author}{Wu, L.}, \bibinfo{author}{Xu, Y.}, \bibinfo{year}{2023}.
\newblock \bibinfo{title}{Dsformer: a double sampling transformer for multivariate time series long-term prediction}, in: \bibinfo{booktitle}{Proceedings of the 32nd ACM International Conference on Information and Knowledge Management}, pp. \bibinfo{pages}{3062--3072}.
\bibitem[{Zeng et~al.(2023)Zeng, Chen, Zhang and Xu}]{DLinear}
\bibinfo{author}{Zeng, A.}, \bibinfo{author}{Chen, M.}, \bibinfo{author}{Zhang, L.}, \bibinfo{author}{Xu, Q.}, \bibinfo{year}{2023}.
\newblock \bibinfo{title}{Are transformers effective for time series forecasting?}
\newblock \bibinfo{journal}{AAAI} .
\bibitem[{Zhang et~al.(2022a)Zhang, Qiang, Zhang, Chen and Jing}]{zhang2022unified}
\bibinfo{author}{Zhang, H.}, \bibinfo{author}{Qiang, W.}, \bibinfo{author}{Zhang, J.}, \bibinfo{author}{Chen, Y.}, \bibinfo{author}{Jing, L.}, \bibinfo{year}{2022}a.
\newblock \bibinfo{title}{Unified feature extraction framework based on contrastive learning}.
\newblock \bibinfo{journal}{Knowledge-Based Systems} \bibinfo{volume}{258}, \bibinfo{pages}{110028}.
\bibitem[{Zhang et~al.(2022b)Zhang, Zhao, Qiang, Chen and Jing}]{zhang2022feature}
\bibinfo{author}{Zhang, H.}, \bibinfo{author}{Zhao, S.}, \bibinfo{author}{Qiang, W.}, \bibinfo{author}{Chen, Y.}, \bibinfo{author}{Jing, L.}, \bibinfo{year}{2022}b.
\newblock \bibinfo{title}{Feature extraction framework based on contrastive learning with adaptive positive and negative samples}.
\newblock \bibinfo{journal}{Neural Networks} \bibinfo{volume}{156}, \bibinfo{pages}{244--257}.
\bibitem[{Zhang and Yan(2023)}]{Crossformer}
\bibinfo{author}{Zhang, Y.}, \bibinfo{author}{Yan, J.}, \bibinfo{year}{2023}.
\newblock \bibinfo{title}{Crossformer: Transformer utilizing cross-dimension dependency for multivariate time series forecasting}.
\newblock \bibinfo{journal}{ICLR} .
\bibitem[{Zhao et~al.(2023)Zhao, Ma, Zhou, Ye, Sun and Qian}]{zhao2023gcformer}
\bibinfo{author}{Zhao, Y.}, \bibinfo{author}{Ma, Z.}, \bibinfo{author}{Zhou, T.}, \bibinfo{author}{Ye, M.}, \bibinfo{author}{Sun, L.}, \bibinfo{author}{Qian, Y.}, \bibinfo{year}{2023}.
\newblock \bibinfo{title}{Gcformer: An efficient solution for accurate and scalable long-term multivariate time series forecasting}, in: \bibinfo{booktitle}{Proceedings of the 32nd ACM International Conference on Information and Knowledge Management}, pp. \bibinfo{pages}{3464--3473}.
\bibitem[{Zhou et~al.(2023)Zhou, Li, Zhang, Zhang, Yan and Xiong}]{zhou2023expanding}
\bibinfo{author}{Zhou, H.}, \bibinfo{author}{Li, J.}, \bibinfo{author}{Zhang, S.}, \bibinfo{author}{Zhang, S.}, \bibinfo{author}{Yan, M.}, \bibinfo{author}{Xiong, H.}, \bibinfo{year}{2023}.
\newblock \bibinfo{title}{Expanding the prediction capacity in long sequence time-series forecasting}.
\newblock \bibinfo{journal}{Artificial Intelligence} \bibinfo{volume}{318}, \bibinfo{pages}{103886}.
\bibitem[{Zhou et~al.(2022)Zhou, Ma, Wen, Wang, Sun and Jin}]{fedformer}
\bibinfo{author}{Zhou, T.}, \bibinfo{author}{Ma, Z.}, \bibinfo{author}{Wen, Q.}, \bibinfo{author}{Wang, X.}, \bibinfo{author}{Sun, L.}, \bibinfo{author}{Jin, R.}, \bibinfo{year}{2022}.
\newblock \bibinfo{title}{{FEDformer}: Frequency enhanced decomposed transformer for long-term series forecasting}.
\newblock \bibinfo{journal}{ICML} .

\end{thebibliography}




\clearpage 





\clearpage
\newpage
\appendix
\section{Appendix Overview}
This supplementary material provides results for additional experiments and details to reproduce our results that could not be included in the paper submission due to space limitations.

\begin{itemize}
    \item \textbf{Appendix \ref{sec:app_notation}} provides the Table which summary all definitions of the notations from the main text. 
    \item \textbf{Appendix \ref{Token_Acquisition}} provides three different tokenization strategy. 
    \item \textbf{Appendix \ref{Sec_why}} explains why adding the similarity matrix of input tokens to the $QK^{\top}$ can enhance the model's ability to preserve OST.
    \item \textbf{Appendix \ref{Dataset_Descriptions}} describes the dataset used in our paper. 

    \item \textbf{Appendix \ref{method_appe}} provides more details about our method. 
    \item \textbf{Appendix \ref{Pseudocode}} provides the pseudo-code. 
    \item \textbf{Appendix \ref{sec:app_theore}} provides proofs and further theoretical analysis of the theory in the text.
    
    \item \textbf{Appendix \ref{Empirical_Analysis_Details}} provides more details about our Experiments. 

    \item \textbf{Appendix \ref{Sec_further}} describes how TEM is integrated into our Concise Dual-branch Transformer Framework (CDTF) and outlines the network architecture and implementation. Extensive multi-dataset comparisons and ablation studies further demonstrate TEM's marked effectiveness and broad applicability within the dual-branch model.
\end{itemize}

Note that before we illustrate the details and analysis, we provide a brief summary about all the experiments conducted in this paper, as shown in Table \ref{tab:app}.

\begin{table*}
    \centering
    \caption{Illustration of the experiments conducted in this work. Note that all experimental results are obtained after five rounds of experiments.}
    \begin{tabular}{p{0.3\textwidth}|p{0.15\textwidth}|p{0.2\textwidth}}
    \toprule
        \textbf{Experiments} & \textbf{Location} & \textbf{Results}\\
    \midrule    
        Comparative experimental results  & Section \ref{sec_all_res} and Appendix \ref{full_res} & Table \ref{tab:mse_std}, Table \ref{tab:mae_std} and Table \ref{tab:full_baseline_results}\\
    \midrule    
        Ablation Study about PTEM, STEM   & Section \ref{sec_ablation} & Tables \ref{tab:ablation_patchtst}, Tables\ref{tab:ablation_itransformer}, and Tables \ref{tab:ablation_transformer}\\
    \midrule    
        Ablation Study about Adaptive Injection Mechanisms   & Section \ref{sec_ablation} & Figure \ref{fig_adp_p}, Figure \ref{fig_adp_i} and Figure \ref{fig_adp_t}\\
    \midrule    
        Ablation Study of Bi-level Optimization Strategy & Section \ref{sec_ablation} & Figure \ref{fig_bi_level1} and Figure \ref{fig_bi_level2}\\

    \midrule    
        Hyperparameter sensitivity & Section \ref{Hyperparameter_sense} & Figure \ref{fig_eta}\\
    \midrule    
        Computational complexity analysis & Section \ref{Complexity} & Figure \ref{fig_eff_tra} and Figure \ref{fig_eff_wea}\\
    \midrule

        Visualization of Prediction & Section \ref{visual_res} & Figure \ref{fig_visual1}, Figure \ref{fig_visual2}, and Figure \ref{fig_visual3}\\
    \midrule    

    The further experiments based on Section \ref{motivation} & Section \ref{Sec_fem} & Figure \ref{fig_moti_more} and Figure \ref{fig_moti_more1}\\

    \midrule 

   Comparison with Other Positional Encodings (PEs) & Section \ref{compare_PE} & Figure \ref{fig_bar}\\

    \midrule 
    
        The Importance of Position Encoding for Variable Tokens & Appendix \ref{import_PE_V_tokens} & Table \ref{tab:vt_PE}\\

    \midrule    
        Comparative experimental results of our Concise Dual-branch Transformer Framework with TEM & Appendix \ref{app_d_res} & Table \ref{tab:app_mse_results} and Table \ref{tab:app_mae_results}\\

    \midrule    
        Ablation study of our Concise Dual-branch Transformer Framework with TEM & Appendix \ref{app_d_res} & Table \ref{tab:ablation_main_text}\\

    \bottomrule
    \end{tabular}
    \label{tab:app}
\end{table*}

\section{Table of Notations}
\label{sec:app_notation}
We list the definitions of all notations in Table \ref{tab_notation-a} and \ref{tab_notation-b}.

\begin{table*}
    \centering
    \caption{The definitions of notations (Part I).}
    \label{tab_notation-a}
    \resizebox{0.7\linewidth}{!}{
    \begin{tabular}{c|c}
    \toprule
    Notations & Definition \\
    \midrule
    \multicolumn{2}{c}{Notations of Data} \\
    \midrule
    $X=\{\mathbf{x}_1,\ldots,\mathbf{x}_T\}\in\mathbb{R}^{T\times N}$ & The input series\\
    $Y=\{\mathbf{x}_{T+1},\ldots,\mathbf{x}_{T+S}\}\in\mathbb{R}^{S\times N}$ &  The output series \\
    $\mathbf{x}_i$ & The series data of the $i$-th time point  \\
    $T$ & The look-back length \\
    $N$ &  The number of variables in the input/output series\\
    $S$ &  The prediction length\\
    \midrule
    \multicolumn{2}{c}{Notations of Topological Relationships} \\
    \midrule
    $\mathcal{G}$ & Positional Topology\\
    $\mathcal{P} = \{1,\cdots,N_t\}$ & The index set of tokens' positions\\
    $N_t$ & The number of the token in the token series\\
    $\Upsilon$ & The class of functions used to generate positional encodings (PE)\\
    $p$ & The functions used to generate positional encodings (PE)\\
    $p(k)$ & The positional encoding (PE) vector of the $k$-th position\\
    $H^l_i$ & the $i$-th intermediate-layer token feature\\
    $H^0$ & the input tokens \\ 
    $\mathcal{S}^l$ & The \textit{semantic topology} at the $l$-th layer\\
    $sim(\cdot)$ & The similarity function\\
    $\Theta$ & The original local structure among input tokens\\ 
    OPT & Original Positional Topology\\
    OST,$\mathcal{S}^0$ & Original Semantic Topology\\
    \bottomrule
    \end{tabular}
    }
\end{table*}

\begin{table*}
    \centering
    \caption{The definitions of notations (Part II).}
    \label{tab_notation-b}
    \resizebox{0.7\linewidth}{!}{
    \begin{tabular}{c|c}
    \toprule
    Notations & Definition \\
    \midrule
    \multicolumn{2}{c}{Notations of Model} \\
    \midrule
    $f_\theta$ & The TSF (time series forecasting) model with parameters $\theta$ \\
    $\Gamma$ & The learnable coefficients for injecting OPT into each Transformer layer \\
    $\Gamma_j^{l,i}$ & The injection strength of OPT into path $j$ (Query, Key, or Value) in head $i$ of layer $l$ \\
    $\Gamma[l,i,j]$ & Tensor notation representing $\Gamma_j^{l,i}$ \\
    $\Xi$ & The learnable coefficients for injecting OST into each attention head \\
    $\Xi_{l,i}$ & The injection strength of OST into the $i$-th head of the $l$-th layer \\
    $\Xi[l,i]$ & Tensor notation representing $\Xi_{l,i}$ \\
    $H^l$ & The input to the $(l+1)$-th Transformer layer \\
    $H_j^{l,i}$ & The modified input for path $j$ in head $i$ of layer $l+1$ \\
    $QK^{\text{T}}$ & The dot-product similarity matrix between Query and Key \\
    $\mathcal{L}_{\text{mse}}$ & The mean squared error loss used in TSF \\
    $\mathcal{D}_b=\{X_i,Y_i\}_{i=1}^{N_b}$ & A mini-batch of $N_b$ samples \\
    $f_\theta^1$ & The updated model parameters after inner-loop optimization \\
    $\eta_1$ & The learning rate for the inner loop \\
    $\eta_2$ & The learning rate for the outer loop \\
    $\nabla_{f_\theta} \mathcal{L}_{\text{mse}}$ & Gradient of the loss with respect to model parameters \\
    $\nabla_{\{\Gamma, \Xi\}} \mathcal{L}_{\text{mse}}$ & Gradient of the loss with respect to topology injection parameters \\
    \midrule
    \multicolumn{2}{c}{Notations of Theory} \\
    \midrule
    $\mathcal{D} = \{(X_i, Y_i)\}_{i=1}^m$ & The training set \\
    $\beta(\cdot)$ & The mixing coefficient of the $\beta$-mixing distribution\\
    $\mu$&The number of mutually independent segments\\
    $b$ & The length of each mutually independent segment\\
    $\mathcal{X}$ & The input spaces for TSF\\
    $\mathcal{Y}$ & The output spaces for TSF\\
    $\mathcal{H}$ & The function class of Transformer-based functions \\
    $h: \mathcal{X} \rightarrow \mathcal{Y}$  &The function of Transformer-based functions\\
    $\mathcal{H}_s^i$ & A function class of the $i$-th Transformer layer \\
    $\ell: \mathcal{Y} \times \mathcal{Y}$& A non-negative, $\xi$-Lipschitz loss function\\
    $M$ & The up bounded of $\ell$\\
    $\delta$& The confidence level \\
    $\mathfrak{R}_\mu(\cdot)$ & The Rademacher complexity\\
    $G(\cdot)$ & An aggregation function over the layer-wise complexities\\
    $p_i(k)$ & The position encoding of the $k$-th token perceived from OPT at layer $i$.\\
    $H_k^{\,i}\in\mathbb R^{d_i}$ & The representation of the $k$-th token of the output tokes of layer $i$ \\
    $d_i$ & The dimension of tokens of layer $i$  \\
    $\Delta_\mathcal{G}^{i}$ & The positional topology distortion at layer $i$ \\
    $\Delta_\mathcal{S}^{i}$ & The semantic topology distortion at layer $i$ \\
    $\mathfrak{R}(\mathcal{H}_s^l)$ & The Rademacher complexity of the $l$-th Transformer layer \\
    $\mathfrak{R}(\mathcal{H}_s^0)$ & The Rademacher complexity of the input layer \\
    $L_i$ & The Lipschitz constant of the $i$-th Transformer layer \\
    $\rho_i$ & The topology preservation ratio at layer $i$ \\
    $\alpha$ & The weight for positional topology distortion \\
    $\beta$ & The weight for semantic topology distortion \\
    \bottomrule
    \end{tabular}
    }
\end{table*}

\section{Token Acquisition}
\label{Token_Acquisition}
\begin{figure*}
    \centering
    \subfloat[]{\includegraphics[width=0.3\textwidth]{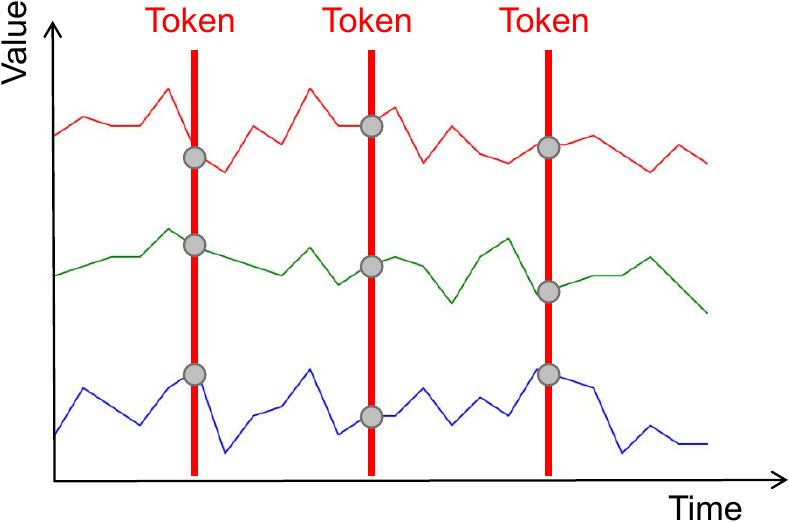}%
    \label{fig:position1}}
    \subfloat[]{\includegraphics[width=0.3\textwidth]{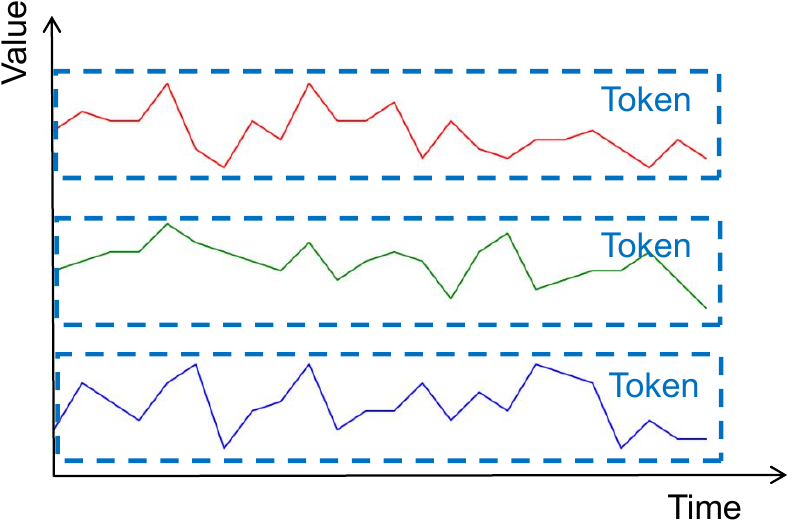}%
    \label{fig:position2}}
    \subfloat[]{\includegraphics[width=0.3\textwidth]{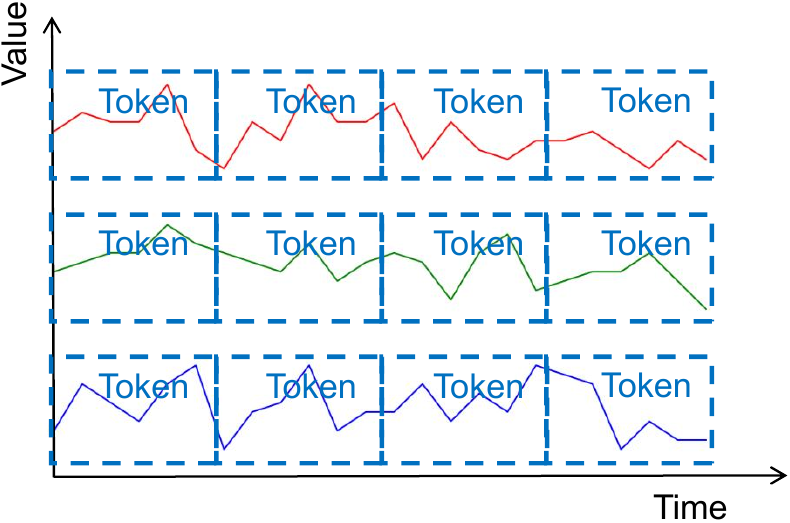}%
    \label{fig:position3}}
    \caption{Visualization of the generation of two types of tokens. (a) temporal tokens, (b) variable tokens, (c) patch token.  
    }
    \label{fig:position_get}
\end{figure*}

The mainstream Transformer-based TSF methods primarily use three types of tokens: 1) Temporal tokens, which include all variables at the same timestamp; 2) Variable tokens, which contain all input time points of a specific variable; 3) Patch tokens, which consist of single variables from adjacent temporal regions. Temporal tokens are primarily used by earlier methods such as Transformer \cite{Transformer}, Informer \cite{zhou2023expanding}. iTransformer \cite{liu2023itransformer} first proposes variable tokens. PatchTST \cite{PatchTST} first proposes patch tokens. We provide a visual representation of the generation processes for these three types of tokens in Figure \ref{fig:position_get} to help readers better understand them. Let the input series be $X \in \mathbb{R}^{T \times N}$, where $T$ is the number of time points and $N$ is the number of variables. Then, $X$ can be converted into $T$ temporal tokens or $N$ variable tokens. If the length of each patch is $l$, $X$ can be converted into $N*T/l$ patch tokens.

\section{Why Adding Similarity Matrix to \texorpdfstring{$QK^{\top}$}{QK-top} Enhances Preservation of OST}
\label{Sec_why}
This section explains why adding the similarity matrix of input tokens to the $QK^{\top}$ can enhance the model's ability to preserve OST.

First, the model's ability to preserve OST can be represented by the dependency between the similarity matrix of the input tokens and that of the output tokens at each layer. A stronger dependency indicates a better preservation ability. In the self-attention mechanism, the new token is computed as follows:
\begin{equation}
    \label{eq_why1}
    V^{new} =  softmax(\frac{QK^{\top}}{\sqrt{D}})V,
\end{equation}
where $Q, K, V \in \mathbb{R}^{N_t\times D } $ are the Query, Key, and Value matrix, respectively. $V^{new}\in \mathbb{R}^{N_t\times D }$ is the new token series generated by the attention layer. Let $\varpi \in \mathbb{R}^{N_t\times N_t }$ be $softmax(\frac{QK^{\top}}{\sqrt{D}})$. We further transform Eq.(\ref{eq_why1}) as:
\begin{equation}
    \label{eq_why2}
    V^{\text{new}}_{i} = \sum_{j=1}^{N_t} \varpi_{i,j} V_j, \quad i = 1, \dots, N_t, 
\end{equation}
where $V_i \in \mathbb{R}^{1\times D }(i=1,...,N_t)$ is the $i$-th token of $V$. $\varpi_{i,j}$ is the attention score between $V_i$ and $V_j$. Let the input token series be $H^0 \in \mathbb{R}^{N_t\times D_{init} }$. $D_{init}$ is the dimension of the input tokens. $H^{0}_i \in \mathbb{R}^{1\times D_{init} }(i=1,...,N_t)$ is the $i$-th token of $H^0$. The similarity matrix is calculated by:
\begin{equation}
    \label{eq_why3}
    \varpi^{init} = H^0[H^0]^\top
\end{equation}
In the main text, we enhance the OST by adding $\varpi^{init}$ into $QK^{\top}$. If the similarity of $H^0_{k_1}$ and $H^0_{k_2}$ is relatively high, the value of $\varpi^{init}_{k_1,k_2}$ is relatively high. In this case, when we add $\varpi^{init}$ into $QK^{\top}$, the $\varpi_{k_1,k_2}$ is increased. According to Eq.(\ref{eq_why2}), the generated $ V^{\text{new}}_{k_1}$  will contain more information from $ V^{\text{new}}_{k_2} $, thereby increasing their similarity. Therefore, adding the similarity matrix of the initial tokens to $QV^{\top}$ will result in initially similar tokens having higher similarity after being processed by the Transformer layer. As a result, the dependency between the similarity matrix of the new tokens and that of the input tokens becomes stronger, which enhances the model’s ability to preserve OST.

\begin{table*}
\centering
\caption{Detailed dataset descriptions. \textbf{Dim} denotes the variate number of each dataset. \textbf{Dataset Size} represents the number of time points in the (Train, Validation, Test) splits, respectively. \textbf{Prediction Length} denotes the future time points to be predicted, with four settings included per dataset. \textbf{Frequency} indicates the sampling interval of time points.}
\label{tab:dataset}
\resizebox{0.8\linewidth}{!}{
\begin{tabular}{l|c|c|c|c|c}
\toprule
\textbf{Dataset} & \textbf{Dim} & \textbf{Prediction Length} & \textbf{Dataset Size} & \textbf{Frequency} & \textbf{Information} \\
\midrule
ETTh1, ETTh2      & 7   & \{96, 192, 336, 720\}     & (8545, 2881, 2881)     & Hourly     & Electricity \\
ETTm1, ETTm2      & 7   & \{96, 192, 336, 720\}     & (34465, 11521, 11521)  & 15min      & Electricity \\
Weather           & 21  & \{96, 192, 336, 720\}     & (36792, 5271, 10540)   & 10min      & Weather \\
ECL               & 321 & \{96, 192, 336, 720\}     & (18317, 2633, 5261)    & Hourly     & Electricity \\
Traffic           & 862 & \{96, 192, 336, 720\}     & (12185, 1757, 3509)    & Hourly     & Transportation \\
Solar-Energy      & 137 & \{96, 192, 336, 720\}     & (36601, 5161, 10417)   & 10min      & Energy \\
PEMS03            & 358 & \{12, 24, 48, 96\}        & (15617, 5135, 5135)    & 5min       & Transportation \\
PEMS04            & 307 & \{12, 24, 48, 96\}        & (10172, 3375, 281)     & 5min       & Transportation \\
PEMS07            & 883 & \{12, 24, 48, 96\}        & (16911, 5622, 468)     & 5min       & Transportation \\
PEMS08            & 170 & \{12, 24, 48, 96\}        & (10690, 3548, 265)     & 5min       & Transportation \\
\bottomrule
\end{tabular}
}
\end{table*}

\section{Dataset Descriptions}
\label{Dataset_Descriptions}
In this paper, we performed tests using twelve real-world datasets. These include: 1) ETT \cite{zhou2023expanding}, which encompasses seven variables related to electricity transformers, gathered from July 2016 to July 2018. There are four subsets where ETTh1 and ETTh2 are recorded every hour, and ETTm1 and ETTm2 are recorded every 15 minutes. 2) Weather \cite{Autoformer}, covering 21 meteorological variables, was recorded at 10-minute intervals in 2020 by the Max Planck Institute for Biogeochemistry's Weather Station. 3) ECL \cite{Autoformer}, detailing the hourly electrical usage of 321 customers. 4) Traffic \cite{Autoformer}, which compiles data on the hourly occupancy rates of roads, monitored by 862 sensors in the San Francisco Bay area's freeways, spanning from January 2015 to December 2016. 5) Solar-Energy \cite{LSTNet}, documenting the solar energy output from 137 photovoltaic (PV) plants in 2006, with data points every 10 minutes. 6) PEMS contains the public traffic network data in California collected by 5-minute windows. We use the same four public subsets (PEMS03, PEMS04, PEMS07, PEMS08) adopted in SCINet \cite{liu2022scinet}. 

We follow the same data processing and train-validation-test set split protocol used in iTransformer \cite{liu2023itransformer}, where the train, validation, and test datasets are strictly divided according to chronological order to make sure there are no data leakage issues. As for the forecasting settings, the lookback length for datasets ETT, Weather, ECL, Solar-Energy, and Traffic is set to 96, while their prediction length varies in $\{96, 192, 336, 720\}$. For PEMS, the lookback length is set to 96, and their prediction length varies in $\{12, 24, 36, 48\}$ which is the same as SCINet \cite{liu2022scinet}. The details of the datasets are provided in Table \ref{tab:dataset}.

\section{Method Details}
\label{method_appe}
In this section, we present the methodological details. First, we describe the details of the Positional Topology Enhancement Module (PTEM) and the Semantic Topology Enhancement Module (STEM). Then, we provide the specific details of how our method is applied to Transformer \cite{Transformer}, iTransformer \cite{liu2023itransformer}, and PatchTST \cite{PatchTST}.

\subsection{More Details of Positional Topology Enhancement Module (PTEM)}
According to Eq.(\ref{eq_GNEM}), PTEM adds the weighted OPT to the input of each Transformer layer. As Definition \ref{eq_Original_Local} and Definition \ref{def:GNI}, OPT essentially represents the positional information of tokens. Since the Transformer models token positions using the positional embedding (PE), OPT is essentially a form of PE. In time series forecasting (TSF), common positional encodings include sinusoidal PE or convolutional PE. For temporal tokens and patch tokens, this paper adopts sinusoidal PE. For variable tokens, each token represents a specific input variable, and permuting the positions of tokens in the variable token sequence does not alter the semantic meaning of the input—i.e., it exhibits permutation invariance. Therefore, convolutional PE, which also possesses permutation invariance, is more suitable for variable tokens.

\subsection{More Details of Semantic Topology Enhancement Module (STEM)}
According to Eq.(\ref{eq_SNEM}), STEM adds the weighted OST to $QK^{\top}$ of each Transformer layer. As Definition \ref{eq_Original_Local} and Definition \ref{def:SNI}, OST essentially represents the similarity of input tokens. We use a similarity matrix of input tokens whose shape is the same as $QK^{\top}$ to represent OST. The similarity matrix is calculated by:
\begin{equation}
    \label{eq_why3_1}
    \varpi^{init} = H^0[H^0]^\top, 
\end{equation}
where $H^0$ is the input tokens. 

\subsection{Applying PTEM and STEM to Transformer}
Transformer \cite{Transformer} uses temporal tokens to process time series forecasting. It employs multiple Transformer layers as the encoder and multiple Transformer layers with cross-attention mechanisms as the decoder. The encoder structure is consistent with the encoder shown in Figure \ref{fig:1}(a). When applying PTEM and STEM to the Transformer model, we integrate them only into the encoder. The method of applying PTEM and STEM to the Transformer encoder can be referred to in Figure \ref{fig:1}(b). The PE used is the sinusoidal PE.

\subsection{Applying PTEM and STEM to iTransformer}
iTransformer \cite{liu2023itransformer} uses variable tokens to handle time series forecasting tasks. It employs multiple Transformer layers as the encoder and uses a linear layer as the decoder. The encoder structure is consistent with the one shown in Figure \ref{fig:1}(a). When applying PTEM and STEM to the iTransformer model, we integrate them only into the encoder. The method of applying PTEM and STEM to the iTransformer encoder can be referred to in Figure \ref{fig:1}(b). The PE used is the convolutional PE.

\subsection{Applying PTEM and STEM to PatchTST}
PatchTST \cite{PatchTST} uses patch tokens to handle time series forecasting tasks. It also employs multiple Transformer layers as the encoder and uses a linear layer as the decoder. The encoder structure is consistent with the one shown in Figure \ref{fig:1}(a). When applying PTEM and STEM to the PatchTST model, we integrate them only into the encoder. The method of applying PTEM and STEM to the PatchTST encoder can be referred to in Figure \ref{fig:1}(b). The PE used is the sinusoidal PE.

\section{Pseudo-code}
\label{Pseudocode}
To facilitate the reader's understanding of the proposed algorithm, we present the pseudo-code of our method (Algorithm \ref{algo:T2B-PE}). The pseudo-code is illustrated using a single-branch method as an example.
\begin{algorithm}[htbp]
  \caption{TEM - Overall Architecture.}\label{algo:T2B-PE}
  \begin{algorithmic}[1]
    \REQUIRE Input time series $X\in\mathbb{R}^{T\times N}$; input Length $T$; predicted length $S$; variates number $N$; token dimension $D$; the number of attention layers $L$; the number of head $n_h$; the tokenization strategy $f_{token}$; the embedding layer $f_{Embed}$; the layer normalization (LN); the Feed-Forward Network (FFN); the linear layer $\mathbf{FC}_{q}$ to generate $Q$; the linear layer $\mathbf{FC}_{k}$ to generate $K$; the linear layer $\mathbf{FC}_{v}$ to generate $V$. 
    \STATE $H^0=f_{token}(X)$
    \STATE OST=$H^0 [H^0]^{\top}$
    \STATE $H^{0}=f_{Embed}(H^0)+\text{PE}$
    \STATE OPT = PE
    \FOR{$l = 0$ to $L-1$}
        \STATE Split $H^l$ to $H^{l,i} (i=1,...,n_t)$
        \FOR{$i = 1$ to $n_h$}
            \STATE $Q = \mathbf{FC}_{q}(H^{l,i}+\Gamma^{l,i}_Q\times \text{OPT})$
            \STATE $K = \mathbf{FC}_{k}(H^{l,i}+\Gamma^{l,i}_K\times \text{OPT})$
            \STATE $V = \mathbf{FC}_{v}(H^{l,i}+\Gamma^{l,i}_V\times \text{OPT})$
            \STATE $A_{init} = QK^\top + \Xi_{l,i}\times \text{OST}$
            \STATE $H^{l,i}_{new} = softmax(\frac{A_{init}}{\sqrt{D}})V$
            \STATE $H^{l,i} = \texttt{LN}\big(H^{l,i} + H^{l,i}_{new}\big)$
        \ENDFOR
        \STATE Fuse $H^{l,i} (i=1,...,n_t)$ to $H^{l}$
        \STATE $H^{l+1} = \texttt{LN}\big(H^l + \texttt{FFN}(H^l)\big)$
    \ENDFOR

    \STATE $\hat{Y} = \texttt{Decoder}(H_{L})$

    \STATE return $\hat{Y}$
  \end{algorithmic}
\end{algorithm}

\section{Theoretical Analysis}
\label{sec:app_theore}
\subsection{Proof of Theorem \ref{theorem:1}}
\label{sec:app_proof1}

\begin{proof}
We begin by invoking the generalization bound for stationary $\beta$-mixing sequences (Theorem 2 in \cite{mohri2008rademacher}). Let
\[
\mathcal{F} := \{(x, y) \mapsto \ell(h(x), y) \mid h \in \mathcal{H} \}
\]
be the composite loss function class. Assume the training dataset $\mathcal{D} = \{(X_i, Y_i)\}_{i=1}^{m}$ is generated from a stationary $\beta$-mixing process with coefficient $\beta(\cdot)$.

Let $m = 2\mu a$ for integers $\mu$ and $a$. Partition the dataset into $2\mu$ blocks of size $a$, and define
\[
\widetilde{\mathcal{D}} := \{ (\widetilde{X}_i, \widetilde{Y}_i) \}_{i=1}^{\mu}
\]
by selecting every other block to approximate independence. Then, with probability at least $1 - \delta$, we have
\begin{equation}\label{eq:mixing-bound}
\begin{aligned}
\mathbb{E}_{(X,Y)}[\ell(h(X), Y)]
&\le \frac{1}{\mu} \sum_{i=1}^{\mu} \ell(h(\widetilde{X}_i), \widetilde{Y}_i)
\quad + \mathfrak{R}_\mu^{\widetilde{\mathcal{D}}}(\mathcal{F})
\quad + M \sqrt{\frac{\log(2/\delta')}{2\mu}}.
\end{aligned}
\end{equation}

where $\delta' = \delta - 4(\mu - 1)\beta(a)$, and $\mathfrak{R}_\mu^{\widetilde{\mathcal{D}}}(\mathcal{F})$ is the empirical Rademacher complexity over $\mu$ independent blocks. Assume the loss function $\ell$ is $\xi$-Lipschitz in its first argument, i.e.,
\[
|\ell(y_1, y) - \ell(y_2, y)| \le \xi \cdot \|y_1 - y_2\|,\quad \forall y_1, y_2, y \in \mathcal{Y}.
\]
Let $f_h(x, y) := \ell(h(x), y)$ for $h \in \mathcal{H}$. Then, by the contraction lemma (Lemma 5.7 in \cite{mohri2018foundations}), we obtain
\[
\mathfrak{R}_\mu^{\widetilde{\mathcal{D}}}(\mathcal{F})
\le \xi \cdot \mathfrak{R}_\mu^{\widetilde{\mathcal{D}}}(\mathcal{H}'),
\]
where $\mathcal{H}' := \{ x \mapsto h(x) \in \mathcal{Y} \mid h \in \mathcal{H} \}$. Now consider that each $h \in \mathcal{H}$ is composed of $L$ Transformer layers. Let $\mathcal{H}_s^l$ be the hypothesis class of the $l$-th layer. Then
\[
h = f^{(L)} \circ f^{(L-1)} \circ \dots \circ f^{(1)},
\quad \text{with } f^{(l)} \in \mathcal{H}_s^l.
\]
By norm-based chaining arguments (see \cite{golowich2018size, neyshabur2018role}), the Rademacher complexity of $\mathcal{H}'$ can be bounded by a function $G$ over the layer-wise complexities:
\[
\mathfrak{R}_\mu^{\widetilde{\mathcal{D}}}(\mathcal{H}')
\le G\left( \left[ \mathfrak{R}_\mu(\mathcal{H}_s^1) \right], \dots, \left[ \mathfrak{R}_\mu(\mathcal{H}_s^L) \right] \right).
\]
Substituting into Eq.(\ref{eq:mixing-bound}), we have
\begin{equation}\label{eq:bound-with-G}
    \begin{aligned}
    \mathbb{E}_{(X,Y)}[\ell(h(X), Y)]
    &\le \frac{1}{\mu} \sum_{i=1}^{\mu} \ell(h(\widetilde{X}_i), \widetilde{Y}_i)
    \quad + \xi \cdot G\!\left( \left[ \mathfrak{R}_\mu(\mathcal{H}_s^1) \right], \dots, \left[ \mathfrak{R}_\mu(\mathcal{H}_s^L) \right] \right)
    \quad + M \sqrt{\frac{\log(2/\delta')}{2\mu}}.
    \end{aligned}
\end{equation}

Finally, since $\widetilde{\mathcal{D}}$ is sampled from $\mathcal{D}$ and blocks are representative under proper mixing, the empirical average can be approximated by the full dataset:
\[
\frac{1}{\mu} \sum_{i=1}^{\mu} \ell(h(\widetilde{X}_i), \widetilde{Y}_i)
\approx \frac{1}{m} \sum_{i=1}^{m} \ell(h(X_i), Y_i).
\]
Substituting into Eq.\ref{eq:bound-with-G} gives the desired bound:
\begin{align*}
\mathbb{E}_{(X,Y)}[\ell(h(X), Y)]
&\le \frac{1}{m} \sum_{i=1}^{m} \ell(h(X_i), Y_i) \quad + \xi \cdot G\left( \left[ \mathfrak{R}_\mu(\mathcal{H}_s^1) \right], \dots, \left[ \mathfrak{R}_\mu(\mathcal{H}_s^L) \right] \right) \quad + M \sqrt{\frac{\log(2/\delta')}{2\mu}}.
\end{align*}
\end{proof}

\subsection{Proof of Theorem \ref{theorem:2}}
\label{sec_appLproof_2}

\begin{proof}

Let $N_t$ denote the number of input tokens and  
$d$ the hidden dimension of each token embedding.  
For layer indices $l\in\{0,\dots,L\}$ write
$\mathcal T_k^{i}\in\mathbb{R}^{d}$ for the representation of token $k$  
after the $i$-th Transformer layer.

Firstly, define the input subspace
\begin{equation}
      \mathcal M_0
  :=
  \operatorname{span}\bigl\{
        \mathcal T_1^{0},\dots,\mathcal T_{N_t}^{0}
  \bigr\},
\end{equation}
and let $P_0$ be the orthogonal projector onto $\mathcal M_0$.  
For every $i\ge1$ and token $k$ decompose
\begin{equation}
\label{eq:decomp}
  \mathcal T_k^{i}
  =
  V_k^{i} + U_k^{i},
\end{equation}
with $V_k^{i} := P_0\mathcal T_k^{i} \in \mathcal M_0$ and $U_k^{i} := (I-P_0)\mathcal T_k^{i} \perp \mathcal M_0$.
The drift coefficient is
\begin{equation}
\label{eq:kappa}
  \kappa_i
  :=
  \frac{\displaystyle
        \sum_{k=1}^{N_t}\!\|U_k^{i}\|_2^{2}}
       {\displaystyle
        \sum_{k=1}^{N_t}\!\|\mathcal T_k^{i}\|_2^{2}}
  =
  1-\rho_i
  \in[0,1],
\end{equation}
where $\rho_i$ is the topology-preservation ratio.

Secondly, let
$\mathcal{D}=\{z_1,\dots,z_m\}$ be a fixed set of samples with $m$, and
$z_k:=\mathcal T_{j(k)}^{i-1}$.  
Let $\sigma_1,\dots,\sigma_m$ be i.i.d. Rademacher variables.
Define
\begin{equation}
\label{eq:R-def}
  \mathfrak R\bigl(\mathcal{H}_s^{\,i}\bigr)
  :=
  \frac{1}{m}\,
  \mathbb{E}_{\sigma}
  \bigl[
    \sup_{f\in\mathcal{H}_s^{\,i}}
    \sum_{k=1}^{m}
    \sigma_k\,f(z_k)
  \bigr].
\end{equation}

Thirdly, set $z_k' := V_{j(k)}^{i}$.  Split
\begin{equation}
\label{eq:split}
  \sum_{k=1}^{m}\sigma_k\,f(z_k)
  =
  \sum_{k=1}^{m}\sigma_k\,f(z_k')
  +
  \sum_{k=1}^{m}\sigma_k\!
    \bigl[f(z_k)-f(z_k')\bigr].
\end{equation}
Aligned part:
\begin{equation}
\label{eq:aligned}
  \frac{1}{m}
  \sup_{f\in\mathcal{H}_s^{\,i}}
  \sum_{k=1}^{m}\sigma_k\,f(z_k')
  \le
  L_i\,
  \mathfrak R\bigl(\mathcal{H}_s^{\,i-1}\bigr).
\end{equation}
Orthogonal part uses $\bigl|f(z_k)-f(z_k')\bigr|\le L_i\|U_{j(k)}^{i}\|_2$:
\begin{equation}
\label{eq:orth}
  \frac{1}{m}
  \sup_{f\in\mathcal{H}_s^{\,i}}
  \sum_{k=1}^{m}\bigl|f(z_k)-f(z_k')\bigr|
  \le
  L_i \sqrt{\kappa_i}\,C_i,
\end{equation}
where
\begin{equation}
\label{eq:C-def}
  C_i
  :=
  \sqrt{\frac{1}{m}
        \sum_{k=1}^{m}
        \|\mathcal T_{j(k)}^{i}\|_2^{2}}.
\end{equation}
Khintchine’s inequality gives
\begin{equation}
\label{eq:C-bound}
  C_i
  \le
  2\,
  \mathfrak R\bigl(\mathcal{H}_s^{\,i-1}\bigr),
  \qquad
  m\ge 1.
\end{equation}
Combining Eq.\eqref{eq:aligned}–\eqref{eq:C-bound} produces
\begin{equation}
\label{eq:single}
  \mathfrak R\bigl(\mathcal{H}_s^{\,i}\bigr)
  \le
  L_i\sqrt{1-\rho_i}\,
  \mathfrak R\bigl(\mathcal{H}_s^{\,i-1}\bigr).
\end{equation}

Fourthly, iterate Eq.\eqref{eq:single} from $i=1$ to $l$, yields:
\begin{equation}
\label{eq:product}
  \mathfrak R\bigl(\mathcal{H}_s^{\,l}\bigr)
  \le
  \mathfrak R\bigl(\mathcal{H}_s^{\,0}\bigr)
  \prod_{i=1}^{l}
  L_i \sqrt{1-\rho_i}.
\end{equation}
Because
\begin{equation}
      \sqrt{1-\rho_i}
  =
  \exp\!\bigl\{\tfrac12\ln(1-\rho_i)\bigr\},
\end{equation}
the bound can be written as
\begin{equation}
\label{eq:final_exp}
  \mathfrak R\bigl(\mathcal{H}_s^{\,l}\bigr)
  \le
  \mathfrak R\bigl(\mathcal{H}_s^{\,0}\bigr)
  \exp\!\Bigl\{
      \tfrac12
      \sum_{i=1}^{l}
      \bigl[\ln L_i + \ln(1-\rho_i)\bigr]
  \Bigr\}.
\end{equation}
This is exactly Eq.\eqref{eq_sgb}, completing the proof.
\end{proof}

\section{Empirical Analyses}
\label{Empirical_Analysis_Details}
\subsection{Weakening of Preservation Positional/Semantic Topology of the Input Tokens from an Information-Theoretic Perspective}
\label{Sec_weak}
From an information-theoretic perspective \cite{cover1999elements}, the raw information contained in the input progressively weakens as network depth increases. This occurs because each layer in the network applies nonlinear transformations and compression to the input, resulting in the loss of information from the original input. Moreover, because positional/semantic topology of the input tokens is typically injected only once at the input layer, this information is gradually diluted or overridden in subsequent layers. Consequently, the model’s preservation of positional/semantic topology of the input tokens weakens with the deepening of network layers.

\subsection{Hilbert-Schmidt Independence Criterion (HSIC)}
\label{sec:app_hsic}
The Hilbert-Schmidt Independence Criterion (HSIC) is a kernel–based, non-parametric measure of statistical dependence between two variables \cite{gangeh2017fast,wang2021learning,perez2018sensitivity}. By embedding the variables into reproducing-kernel Hilbert spaces (RKHSs) and evaluating the Hilbert–Schmidt norm of their cross–covariance operator, HSIC captures both linear and nonlinear relationships without requiring explicit density estimation.

HSIC (Hilbert-Schmidt Independence Criterion) theoretically requires the input to consist of two random variables. However, sinusoidal position encoding (PE) does not appear to meet this requirement, as it is deterministically generated by a fixed formula. In practice, though, HSIC relies solely on observed sample pairs $(x_i, y_i)$, and does not require these samples to be drawn from a truly random sampling process (see \cite{perez2018sensitivity}). Specifically, given a set of token representations $H \in \mathbb{R}^{N_t \times D}$ and their corresponding deterministic sinusoidal position encodings $PE \in \mathbb{R}^{N_t \times D}$, we treat $H$ and $PE$ as collections of $N_t$ variables. Each token and its position encoding at the same index form a sample pair $(H^i, PE^i)$, where $i = 1, \dots, N_t$. HSIC focuses on the dependency patterns among these sample pairs rather than on their generative process (see \cite{perez2018sensitivity}). Therefore, using HSIC to verify whether token representations contain the positional information encoded by the sinusoidal PE remains reasonable and valid. Similarly, when evaluating the dependency between learnable position encodings (PE) and token features, or between the similarity matrix of input tokens and that of intermediate layer token representations, we can construct sample pairs based on corresponding positions or corresponding pairs. These sample pairs naturally satisfy the structural requirements for HSIC estimation, making it a suitable tool for assessing whether the model captures certain structural information or adjacency relations explicitly or implicitly.

\paragraph{Computation.}
Given two variables X $ \in \mathbb{R}^{N_t \times D}$ and Y $ \in \mathbb{R}^{N_t \times D}$, they can be decomposed to sample set: $\{(x_i,y_i)\}_{i=1}^{N_t}$, where $x_i,y_i \in \mathbb{R}^{1 \times D}$. For a sample $(x_i,y_i)$, define the kernel matrices
$K^k \in \mathbb{R}^{N_t \times N_t} (K^k_{ij}=\mathfrak{k}^k(x_i,x_j))$ and $L^k \in \mathbb{R}^{N_t \times N_t} (L^k_{ij}=\mathfrak{l}^k(y_i,y_j))$,
and the centering matrix $H_{center}=I_{N_t}-\tfrac{1}{N_t}\mathbf{1}\mathbf{1}^{\!\top}$. $\mathfrak{k}^k(x_i,x_j)$ and $\mathfrak{l}^k(y_i,y_j)$ are the Gaussian kernel functions which are expressed as:
\begin{equation}
\begin{aligned}
\mathfrak{k}^k(\mathbf{x}_i,\mathbf{x}_j) = \exp\!\left(-\frac{\lVert\mathbf{x}_i - \mathbf{x}_j\rVert_2^{2}}{2\sigma_x^{2}}\right), \mathfrak{l}^k(\mathbf{y}_i,\mathbf{y}_j) = \exp\!\left(-\frac{\lVert\mathbf{y}_i - \mathbf{y}_j\rVert_2^{2}}{2\sigma_y^{2}}\right)
\end{aligned}
\end{equation}
where $\sigma_x, \sigma_y$ are computed by the median heuristic. Specifically, first, given the sample matrix $X\!\in\!\mathbb{R}^{N\times D}$, evaluate the squared Euclidean distance for every unordered pair of samples
\begin{equation}
d_{ij} \;=\;\lVert \mathbf{x}_i-\mathbf{x}_j\rVert_{2}^{2}, \qquad i<j .
\end{equation}
Second, collect all non‑zero $d_{ij}$ values and take their median, denoted by $v_m$.  Finally, set the bandwidth as
\begin{equation}
\sigma_x \;=\;\sqrt{\tfrac{v_m}{2}}.
\end{equation}
$\sigma_y$ is obtained in the same manner. After getting the above items, the HSIC of two variables is computed as follows:
\begin{equation}
\label{eq:hsic_emp}
\mathrm{HSIC}(X,Y)
=\frac{1}{(N_t-1)^2}\,\mathrm{tr}\!\bigl(K^kH_{center}L^kH_{center}\bigr),
\end{equation}
where $\mathrm{tr}(\cdot)$ denotes the matrix trace.  
Equation~\eqref{eq:hsic_emp} involves only matrix multiplications and traces, making HSIC straightforward to implement and scalable with modern hardware.

\begin{table*}
\caption{Full results for the TSF task. The input length for all baseline models is set to 96, and the prediction lengths include $ \{ 12,24,48,96  \} $. The results of our method are averaged over five random seeds, and the standard deviation across the five runs is reported after the ``$\pm$'' symbol.}
\label{tab:full_baseline_results}
\renewcommand{\arraystretch}{0.85}
\centering
\resizebox{1\columnwidth}{!}{%
\begin{threeparttable}
\begin{small}
\renewcommand{\multirowsetup}{\centering}
\setlength{\tabcolsep}{1pt}
\begin{tabular}{c|c|cc|cc|cc|cc|cc|cc|cc|cc|cc|cc}
\toprule
\multicolumn{2}{c}{\multirow{2}{*}{Models}} &
\multicolumn{2}{c}{\rotatebox{0}{\scalebox{0.8}{\textbf{PatchTST+TEM}}}} &
\multicolumn{2}{c}{\rotatebox{0}{\scalebox{0.8}{PatchTST}}} &
\multicolumn{2}{c}{\rotatebox{0}{\scalebox{0.8}{\textbf{iTransformer+TEM}}}} &
\multicolumn{2}{c}{\rotatebox{0}{\scalebox{0.8}{iTransformer}}} &
\multicolumn{2}{c}{\rotatebox{0}{\scalebox{0.8}{\textbf{Transformer+TEM}}}} &
\multicolumn{2}{c}{\rotatebox{0}{\scalebox{0.8}{Transformer}}} &
\multicolumn{2}{c}{\rotatebox{0}{\scalebox{0.8}{Crossformer}}} &
\multicolumn{2}{c}{\rotatebox{0}{\scalebox{0.8}{TimesNet}}} &
\multicolumn{2}{c}{\rotatebox{0}{\scalebox{0.8}{DLinear}}} &
\multicolumn{2}{c}{\rotatebox{0}{\scalebox{0.8}{TIDE}}} \\

\multicolumn{2}{c}{} &
\multicolumn{2}{c}{\scalebox{0.8}{(Ours)}} &
\multicolumn{2}{c}{\scalebox{0.8}{(2023)}} &
\multicolumn{2}{c}{\scalebox{0.8}{(Ours)}} &
\multicolumn{2}{c}{\scalebox{0.8}{(2024)}} &
\multicolumn{2}{c}{\scalebox{0.8}{(Ours)}} &
\multicolumn{2}{c}{\scalebox{0.8}{(2017)}} &
\multicolumn{2}{c}{\scalebox{0.8}{(2023)}} &
\multicolumn{2}{c}{\scalebox{0.8}{(2023)}} &
\multicolumn{2}{c}{\scalebox{0.8}{(2023)}} &
\multicolumn{2}{c}{\scalebox{0.8}{(2023)}} \\

\cmidrule(lr){3-4}\cmidrule(lr){5-6}\cmidrule(lr){7-8}\cmidrule(lr){9-10}
\cmidrule(lr){11-12}\cmidrule(lr){13-14}\cmidrule(lr){15-16}\cmidrule(lr){17-18}
\cmidrule(lr){19-20}\cmidrule(lr){21-22}
\multicolumn{2}{c}{Metric} &
\scalebox{0.78}{MSE} & \scalebox{0.78}{MAE} & \scalebox{0.78}{MSE} & \scalebox{0.78}{MAE} &
\scalebox{0.78}{MSE} & \scalebox{0.78}{MAE} & \scalebox{0.78}{MSE} & \scalebox{0.78}{MAE} &
\scalebox{0.78}{MSE} & \scalebox{0.78}{MAE} & \scalebox{0.78}{MSE} & \scalebox{0.78}{MAE} &
\scalebox{0.78}{MSE} & \scalebox{0.78}{MAE} & \scalebox{0.78}{MSE} & \scalebox{0.78}{MAE} &
\scalebox{0.78}{MSE} & \scalebox{0.78}{MAE} & \scalebox{0.78}{MSE} & \scalebox{0.78}{MAE} \\ \toprule
\multirow{5}{*}{\rotatebox{90}{\scalebox{0.95}{PEMS03}}}
&  \scalebox{0.78}{12} & 0.095$\pm$0.001 & 0.207$\pm$0.001 & 0.099 & 0.216 & \textcolor{red}{0.067}$\pm$0.001 & \textcolor{red}{0.165}$\pm$0.001 & \textcolor{blue}{0.071} & \textcolor{blue}{0.174} & 0.133$\pm$0.002 & 0.290$\pm$0.001 & 0.139 & 0.302 & 0.090 & 0.203 & 0.085 & 0.192 & 0.122 & 0.243 & 0.178 & 0.305 \\[1pt]
&  \scalebox{0.78}{24} & 0.136$\pm$0.000 & 0.249$\pm$0.001 & 0.142 & 0.259 & \textcolor{red}{0.088}$\pm$0.001 & \textcolor{red}{0.191}$\pm$0.000 & \textcolor{blue}{0.093} & \textcolor{blue}{0.201} & 0.191$\pm$0.001 & 0.348$\pm$0.001 & 0.199 & 0.363 & 0.121 & 0.240 & 0.118 & 0.223 & 0.201 & 0.317 & 0.257 & 0.371 \\[1pt]
&  \scalebox{0.78}{48} & 0.203$\pm$0.001 & 0.306$\pm$0.001 & 0.211 & 0.319 & \textcolor{red}{0.119}$\pm$0.000 & \textcolor{red}{0.224}$\pm$0.001 & \textcolor{blue}{0.125} & \textcolor{blue}{0.236} & 0.283$\pm$0.002 & 0.429$\pm$0.001 & 0.295 & 0.447 & 0.202 & 0.317 & 0.155 & 0.260 & 0.333 & 0.425 & 0.379 & 0.463 \\[1pt]
&  \scalebox{0.78}{96} & 0.258$\pm$0.001 & 0.355$\pm$0.002 & 0.269 & 0.370 & \textcolor{red}{0.156}$\pm$0.001 & \textcolor{red}{0.261}$\pm$0.001 & \textcolor{blue}{0.164} & \textcolor{blue}{0.275} & 0.362$\pm$0.002 & 0.497$\pm$0.002 & 0.377 & 0.518 & 0.262 & 0.367 & 0.228 & 0.317 & 0.457 & 0.515 & 0.490 & 0.539 \\[1pt]
\cmidrule(lr){2-22}
&  \scalebox{0.78}{Avg} & 0.173 & 0.279 & 0.180 & 0.291 & \textcolor{red}{0.107} & \textcolor{red}{0.210} & \textcolor{blue}{0.113} & \textcolor{blue}{0.221} & 0.242 & 0.391 & 0.252 & 0.408 & 0.169 & 0.281 & 0.147 & 0.248 & 0.278 & 0.375 & 0.326 & 0.419 \\
\midrule
\multirow{5}{*}{\rotatebox{90}{\scalebox{0.95}{PEMS04}}}
&  \scalebox{0.78}{12} & 0.101$\pm$0.001 & 0.215$\pm$0.001 & 0.105 & 0.224 & \textcolor{red}{0.074}$\pm$0.000 & \textcolor{red}{0.174}$\pm$0.001 & \textcolor{blue}{0.078} & \textcolor{blue}{0.183} & 0.141$\pm$0.002 & 0.301$\pm$0.002 & 0.147 & 0.314 & 0.098 & 0.218 & 0.087 & 0.195 & 0.148 & 0.272 & 0.219 & 0.340 \\[1pt]
&  \scalebox{0.78}{24} & 0.147$\pm$0.000 & 0.264$\pm$0.001 & 0.153 & 0.275 & \textcolor{red}{0.090}$\pm$0.001 &\textcolor{red}{0.195}$\pm$0.000 & \textcolor{blue}{0.095} & \textcolor{blue}{0.205} & 0.205$\pm$0.002 & 0.370$\pm$0.001 & 0.214 & 0.385 & 0.131 & 0.256 & 0.103 & 0.215 & 0.224 & 0.340 & 0.292 & 0.398 \\[1pt]
&  \scalebox{0.78}{48} & 0.220$\pm$0.001 & 0.325$\pm$0.001 & 0.229 & 0.339 & \textcolor{red}{0.114}$\pm$0.000 & \textcolor{red}{0.221}$\pm$0.001 & \textcolor{blue}{0.120} & \textcolor{blue}{0.233} & 0.308$\pm$0.003 & 0.456$\pm$0.002 & 0.321 & 0.475 & 0.205 & 0.326 & 0.136 & 0.250 & 0.355 & 0.437 & 0.409 & 0.478 \\[1pt]
&  \scalebox{0.78}{96} & 0.279$\pm$0.001 & 0.373$\pm$0.002 & 0.291 & 0.389 & \textcolor{red}{0.142}$\pm$0.001 & \textcolor{red}{0.249}$\pm$0.001 & \textcolor{blue}{0.150} & \textcolor{blue}{0.262} & 0.391$\pm$0.002 & 0.523$\pm$0.002 & 0.407 & 0.545 & 0.402 & 0.457 & 0.190 & 0.303 & 0.452 & 0.504 & 0.492 & 0.532 \\[1pt]
\cmidrule(lr){2-22}
&  \scalebox{0.78}{Avg} & 0.187 & 0.295 & 0.195 & 0.307 & \textcolor{red}{0.105} & \textcolor{red}{0.210} & \textcolor{blue}{0.111} & \textcolor{blue}{0.221} & 0.261 & 0.412 & 0.272 & 0.430 & 0.209 & 0.314 & 0.129 & 0.241 & 0.295 & 0.388 & 0.353 & 0.437 \\
\midrule
\multirow{5}{*}{\rotatebox{90}{\scalebox{0.95}{PEMS07}}}
&  \scalebox{0.78}{12} & 0.091$\pm$0.001 & 0.199$\pm$0.001 & 0.095 & 0.207 & \textcolor{red}{0.064}$\pm$0.001 & \textcolor{red}{0.157}$\pm$0.000 & \textcolor{blue}{0.067} & \textcolor{blue}{0.165} & 0.128$\pm$0.002 & 0.278$\pm$0.001 & 0.133 & 0.290 & 0.094 & 0.200 & 0.082 & 0.181 & 0.115 & 0.242 & 0.173 & 0.304 \\[1pt]
&  \scalebox{0.78}{24} & 0.144$\pm$0.000 & 0.252$\pm$0.001 & 0.150 & 0.262 & \textcolor{red}{0.084}$\pm$0.001 & \textcolor{red}{0.180}$\pm$0.001 & \textcolor{blue}{0.088} & \textcolor{blue}{0.190} & 0.202$\pm$0.002 & 0.352$\pm$0.002 & 0.210 & 0.367 & 0.139 & 0.247 & 0.101 & 0.204 & 0.210 & 0.329 & 0.271 & 0.383 \\[1pt]
&  \scalebox{0.78}{48} & 0.243$\pm$0.001 & 0.326$\pm$0.001 & 0.253 & 0.340 & \textcolor{red}{0.104}$\pm$0.000 & \textcolor{red}{0.204}$\pm$0.001 & \textcolor{blue}{0.110} & \textcolor{blue}{0.215} & 0.340$\pm$0.002 & 0.457$\pm$0.002 & 0.354 & 0.476 & 0.311 & 0.369 & 0.134 & 0.238 & 0.398 & 0.458 & 0.446 & 0.495 \\[1pt]
&  \scalebox{0.78}{96} & 0.332$\pm$0.001 & 0.388$\pm$0.001 & 0.346 & 0.404 & \textcolor{red}{0.132}$\pm$0.001 & \textcolor{red}{0.233}$\pm$0.001 & \textcolor{blue}{0.139} & \textcolor{blue}{0.245} & 0.465$\pm$0.003 & 0.543$\pm$0.002 & 0.484 & 0.566 & 0.396 & 0.442 & 0.181 & 0.279 & 0.594 & 0.553 & 0.628 & 0.577 \\[1pt]
\cmidrule(lr){2-22}
&  \scalebox{0.78}{Avg} & 0.203 & 0.291 & 0.211 & 0.303 & \textcolor{red}{0.096} & \textcolor{red}{0.194} & \textcolor{blue}{0.101} & \textcolor{blue}{0.204} & 0.284 & 0.408 & 0.295 & 0.425 & 0.235 & 0.315 & 0.124 & 0.225 & 0.329 & 0.395 & 0.380 & 0.440 \\
\midrule
\multirow{5}{*}{\rotatebox{90}{\scalebox{0.95}{PEMS08}}}
&  \scalebox{0.78}{12} & 0.161$\pm$0.001 & 0.223$\pm$0.001 & 0.168 & 0.232 & \textcolor{red}{0.075}$\pm$0.000 & \textcolor{red}{0.173}$\pm$0.001 & \textcolor{blue}{0.079} & \textcolor{blue}{0.182} & 0.226$\pm$0.002 & 0.312$\pm$0.002 & 0.235 & 0.325 & 0.165 & 0.214 & 0.112 & 0.212 & 0.154 & 0.276 & 0.227 & 0.343 \\[1pt]
&  \scalebox{0.78}{24} & 0.215$\pm$0.001 & 0.270$\pm$0.001 & 0.224 & 0.281 & \textcolor{red}{0.109}$\pm$0.001 & \textcolor{red}{0.208}$\pm$0.001 & \textcolor{blue}{0.115} & \textcolor{blue}{0.219} & 0.301$\pm$0.002 & 0.377$\pm$0.002 & 0.314 & 0.393 & 0.215 & 0.260 & 0.141 & 0.238 & 0.248 & 0.353 & 0.318 & 0.409 \\[1pt]
&  \scalebox{0.78}{48} & 0.308$\pm$0.001 & 0.340$\pm$0.001 & 0.321 & 0.354 & \textcolor{red}{0.177}$\pm$0.001 & \textcolor{red}{0.223}$\pm$0.001 & \textcolor{blue}{0.186} & \textcolor{blue}{0.235} & 0.431$\pm$0.002 & 0.476$\pm$0.002 & 0.449 & 0.496 & 0.315 & 0.355 & 0.198 & 0.283 & 0.440 & 0.470 & 0.497 & 0.510 \\[1pt]
&  \scalebox{0.78}{96} & 0.392$\pm$0.001 & 0.400$\pm$0.002 & 0.408 & 0.417 & \textcolor{red}{0.21}0$\pm$0.001 & \textcolor{red}{0.254}$\pm$0.001 & \textcolor{blue}{0.221} & \textcolor{blue}{0.267} & 0.548$\pm$0.003 & 0.561$\pm$0.002 & 0.571 & 0.584 & 0.377 & 0.397 & 0.320 & 0.351 & 0.674 & 0.565 & 0.721 & 0.592 \\[1pt]
\cmidrule(lr){2-22}
&  \scalebox{0.78}{Avg} & 0.269 & 0.308 & 0.280 & 0.321 & \textcolor{red}{0.142} & \textcolor{red}{0.215} & \textcolor{blue}{0.150} & \textcolor{blue}{0.226} & 0.376 & 0.432 & 0.392 & 0.450 & 0.268 & 0.307 & 0.193 & 0.271 & 0.379 & 0.416 & 0.441 & 0.464 \\
\bottomrule
\end{tabular}
\end{small}
\end{threeparttable}}
\end{table*}

\begin{table*}
\centering
\caption{The performance of applying sinusoidal PE to iTransformer on the Weather dataset. The lookback length for all models is set to 96, and the prediction lengths $ \in \{ 96,192,336,720 \}$.}
\label{tab:vt_PE}
\resizebox{0.6\textwidth}{!}{ 
\begin{tabular}{l|c|c|c|c}
\midrule
\textbf{MSE/MAE} &  \textbf{Dataset} & \textbf{Prediction length} & \textbf{Original} & \textbf{+sinusoidal PE} \\
\midrule
iTransformer
&Weather&96  
& 0.174/0.214 & 0.172/0.212 \\
&&192 & 0.221/0.254 & 0.219/0.257 \\
&&336 & 0.278/0.296 & 0.276/0.295 \\
&&720 & 0.358/0.349 & 0.356/0.347 \\
&&Avg & 0.258/0.279 & 0.256/0.278 \\
\bottomrule
\end{tabular}
}
\end{table*}

\subsection{More Forecasting Results}
\label{full_res}
Due to space constraints in the main text, this section presents the more multivariate forecasting results. We conducted an extensive evaluation of several competitive baselines on challenging forecasting tasks, and the outcomes are summarized in Table \ref{tab:full_baseline_results}. All reported figures of our methods are the averages over five random seeds. As the table shows, integrating our TEM into various baseline models consistently yields improved forecasting performance.

\subsection{The Importance of Position Encoding for Variable Tokens}
\label{import_PE_V_tokens}
iTransformer considers the PE of variable tokens unimportant \cite{liu2023itransformer}, which is an unreasonable viewpoint. Although swapping the positions of variable tokens does not alter the semantics of the input sequence (i.e., permutation invariance), making the relative positions of variable tokens seemingly meaningless, the absolute positions of variable tokens still play a crucial role. Through the absolute positions, the model can directly recognize variables without needing to identify them indirectly through features. We believe that the PE of variable tokens remains meaningful. Experimental results (Table \ref{tab:vt_PE}) also validate this perspective: on the Weather dataset, adding sinusoidal PE to iTransformer and retraining it significantly reduces MSE/MAE. Furthermore, other methods that utilize variable tokens, such as Crossformer \cite{Crossformer}, also incorporate PE for variable tokens.

\section{Further Work: Applying TEM to a Dual-Branch Framework}
\label{Sec_further}
In the main text, we apply our TEM to three single-branch Transformer-based time prediction methods (PatchTST \cite{PatchTST}, iTransformer \cite{liu2023itransformer}, Transformer \cite{Transformer}). To further verify the generality of TEM, we apply it to a dual-branch Transformer-based time prediction method in this section. However, current dual-branch methods, for the sake of computational efficiency, either reduce the number of tokens in the output series as the network deepens \cite{Crossformer}, or replace the attention mechanism with more efficient convolution operations \cite{dai2025vtformer,huang2023dbaformer}. In the former work \cite{Crossformer}, the similarity matrix of the initial tokens has a shape inconsistent with the intermediate $QK^{\top}$ in the network, making direct addition infeasible and thus preventing the injection of OST via Equation \ref{eq_SNEM}. In the latter work \cite{dai2025vtformer}, $QK^{\top}$ is absent altogether, similarly making it impossible to inject OST through Equation \ref{eq_SNEM}. These approaches make it infeasible to apply STEM to them.

To address the aforementioned issues, we design a Concise Dual-branch Transformer Framework (CDTF) to incorporate our TEM method, aiming to validate TEM's applicability and effectiveness within a dual-branch architecture. The framework is straightforward, with each branch implemented as a basic Transformer encoder: the left branch processes temporal tokens, while the right branch handles variable tokens. In this section, we first introduce the architectural details of the model (see Appendix \ref{app_d_method}), followed by implementation details (see Appendix \ref{app_d_imp}), and finally present comparative results with existing methods as well as ablation studies (see Appendix \ref{app_d_res}).

\begin{figure*}
    \centering
    \includegraphics[width=0.7\textwidth]{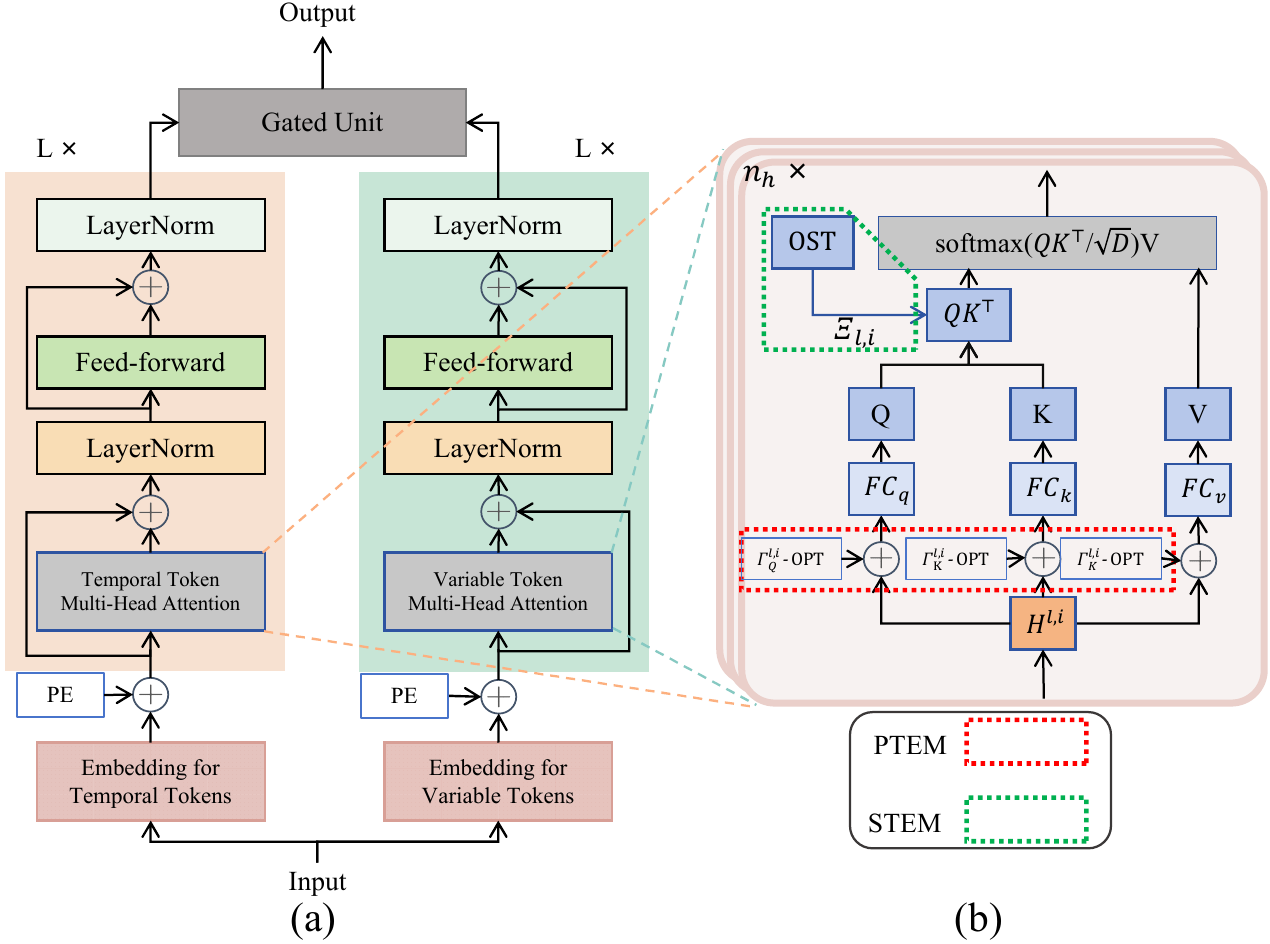}
    \caption{Overview of the Concise Dual-branch Transformer Framework (CDTF). (a) The structure of CDTF. (b) Illustration of the integration locations of PTEM and STEM within CDTF. }
    \label{fig:appendix_method}
\end{figure*}

\subsection{A Concise Dual-branch Transformer Framework with our TEM}
\label{app_d_method}
In this subsection, we present our Concise Dual-branch Transformer Framework (CDTF), which leverages temporal tokens and variable tokens to boost model performance. CDTF first processes the correlations between temporal tokens and the correlations between variable tokens in two separate branches, each built on the original Transformer encoder. It then merges the features from both branches through a gated unit to make predictions. An overview of CDTF is shown in Fig. \ref{fig:appendix_method}. Next, we detail the temporal-token branch, the variable-token branch, and the fusion module (the gating unit) in turn. 

\textbf{Temporal token branch:} 
This branch aims to handle the correlations between the temporal tokens. Specifically, we get enhanced input features ($H^{l-1,i}_Q$, $H^{l-1,i}_K$, $H^{l-1,i}_V$) of the $l$-th layer of this branch through Eq.(\ref{eq_GNEM}). Then, we calculate the $Q$, $K$, and $V$ of the $i$-th attention head in the $l$-th layer of this branch as follows:
\begin{equation}
\label{qkv_s}
\begin{array}{l}
    Q^{time}_{l,i} = \mathbf{FC}_{q}^{time}(H^{l-1,i}_Q) \in \mathbb{R}^{T\times D}, K^{time}_{l,i} = \mathbf{FC}_{k}^{time}(H^{l-1,i}_K) \in \mathbb{R}^{T\times D}, \\
    V^{time}_{l,i} = \mathbf{FC}_{v}^{time}(H^{l-1,i}_V) \in \mathbb{R}^{T\times D}, l=1,...,L, i=1,...,n_t
\end{array}
\end{equation}
where $\mathbf{FC}_{q}^{time}$, $\mathbf{FC}_{k}^{time}$ and $\mathbf{FC}_{v}^{time}$ are three fully connected layers, $L$ is the number of encoder layers, $n_t$ is the number of head. After obtaining $Q$, $K$, and $V$, according to Eq.(\ref{eq_SNEM}), we calculate the attention map of the $i$-th attention head in the $l$-th layer of this branch as follows:
\begin{equation}
\label{eq_att}
\begin{array}{l}
    A_{l,i}=\mathbf{softmax}(\frac{Q^{time}_{l,i}[K^{time}_{l,i}]^{\top}+\Xi_{l,i} \times \mathcal{S}^0}{\sqrt{D}}), l=1,...L , i=1,...,n_t. 
\end{array}
\end{equation}
The rest of the computation process remains consistent with the vanilla Transformer encoder. Firstly, we use attention mechanism to produce the refined feature $A_{l,i}V^{time}_{l,i}$, then we fuse the multi-head output, next the fused feature goes through a feedforward network (FFN) to obtain the output of the current layer, recorded as $H^{time}_{l}$ which serves as the input embedding for the next layer. This process is repeated $L$ times throughout the entire model. Finally, we obtain the output features of the temporal tokens branch, i.e., $H^{time}_{out} \in \mathbb{R}^{T\times D}$.

\textbf{Variable token branch}: 
This branch aims to handle the correlations between variable tokens. The computation process is the same as the temporal token branch. The only difference is that this branch processes variable tokens with shape $N \times D$. Finally, we obtain the output features of the variable tokens branch, i.e., $H^{var}_{out} \in \mathbb{R}^{N\times D}$.

\textbf{Fusion Module}: 
In this module, we aim to merge the outputs of the above two branches. We propose a gated unit specially designed for TSF to achieve this goal. The gated unit dynamically assesses the two branches' importance based on their outputs and performs weighted fusion. Specifically, we first use three fully connected layers to reshape the output features from both branches to the same shape:
\begin{equation}
\label{eq_10}
    \begin{aligned}
    \begin{array}{l}
   F_{t}=\mathbf{FC}_{t2}[\mathbf{FC}_{t1}(H_{out}^{time})]^{\top}\in \mathbb{R}^{N\times S},
   F_{v}=\mathbf{FC}_{v}(H_{out}^{var})\in \mathbb{R}^{N\times S},
    \end{array}
    \end{aligned}
\end{equation}
where $\mathbf{FC}_{t1}:\mathbb{R}^{D}\to \mathbb{R}^{N}$, $\mathbf{FC}_{t2}:\mathbb{R}^{T}\to \mathbb{R}^{S}$, $\mathbf{FC}_{v}:\mathbb{R}^{D}\to\mathbb{R}^{S}$ are three fully connected layers, $F_{t}$ is the refined feature of the temporal token branch, $F_{v}$ is the refined features of the variable token branch.
Next, we employ the gated unit to fuse $F_{t}$ and $F_{t}$:
\begin{equation}
\begin{array}{l}
    G=\sigma ([F_{t},F_{v}]W_{f})\in \mathbb{R}^{N\times S}, 
   \hat{Y}=G\odot F_v+(1-G)\odot F_v,
\end{array}
\end{equation}
where $\left [ \cdots  \right ] $ denotes the operation of concatenation, [$F_{t}, F_{v}$] $\in \mathbb{R}^{N\times 2\mathrm {S}}$, $W_{f}\in \mathbb{R}^{2\mathrm {S}\times \mathrm {S}}$, $\sigma$ denotes a sigmoid operation, $\odot$ denotes element-wise multiplication and $\hat{Y}^{\top}$$\in \mathbb{R}^{S\times N}$ is the final output. For simplicity, we let $\hat{Y}=\hat{Y}^{\top}$. 

We use the MSE loss and the bi-level optimization strategy mentioned in Section \ref{SEC_Loss} to update the model.

\subsection{Implementation Details}
\label{app_d_imp}
Data undergo mean-std normalization before input into the model and inverse mean-std normalization before output. The proposed models are trained using an NVIDIA 4090 GPU. MSE and MAE are employed as evaluation metrics. In our experiments, we set the number of layers of the encoder $L$ to 2 for ETT datasets, 3 for the Weather dataset, and 4 for other datasets. The token dimension $D$ is set to 128 for the ETT dataset and 512 for other datasets. The head number of attention is set to 8. The batch size is set to 16 for the ECL and Traffic datasets and 32 for others. For the bi-level optimization, in the inner loop, we adopt the Adam optimizer \cite{Adam}with a decaying learning rate to update the backbone parameters. The initial learning rate is shown in Table \ref{tab:model_hyperparameters}. In the outer loop, the positional/semantic topology weights are updated using the Adam optimizer with a decaying learning rate. The initial learning rate is set to $1e-3$. The total number of epochs is 40. The training process will stop early if the validation loss doesn't decrease within three epochs. 

\begin{table}
\centering
\caption{Hyperparameter setting of CDTF for different datasets.}
\resizebox{0.3\columnwidth}{!}{
\begin{tabular}{l|l|l}
\toprule
\textbf{Dataset} & \textbf{Prediction Length} & \textbf{$lr_{init}$} \\
\midrule
ETTh1     & 96   & 9.2e-3        \\
          & 192  & 1.33e-2        \\
          & 336  & 3e-4        \\
          & 720  & 1e-4        \\
\midrule
ETTh2     & 96   & 2.9e-3        \\
          & 192  & 5e-3        \\
          & 336  & 7e-3        \\
          & 720  & 7e-3        \\
\midrule
ETTm1     & 96   & 1.08e-2        \\
          & 192  & 6.2e-3        \\
          & 336  & 9.8e-3        \\
          & 720  & 4.5e-3        \\
\midrule
ETTm2     & 96   & 1.1e-3        \\
          & 192  & 1e-4        \\
          & 336  & 2e-3        \\
          & 720  & 6.6e-3        \\
\midrule
ECL       & 96   & 9.05e-4 \\
          & 192  & 1.08e-3  \\
          & 336  & 1.38e-3  \\
          & 720  & 9.2e-4   \\
\midrule
Traffic   & 96   & 1.07e-3  \\
          & 192  & 1.01e-3 \\
          & 336  & 9.2e-4 \\
          & 720  & 1.22e-3 \\
\midrule
Weather   & 96   & 1.3e-3        \\
          & 192  & 9.1e-5        \\
          & 336  & 7e-4        \\
          & 720  & 7e-4        \\
\midrule
Solar-Energy & 96   & 1.6e-4   \\
             & 192  & 1.6e-4       \\
             & 336  & 1.6e-4        \\
             & 720  & 1.6e-4        \\
\bottomrule
\end{tabular}
}
\label{tab:model_hyperparameters}
\end{table}

\subsection{Experiment Results}
\label{app_d_res}
\paragraph{Comparative Experimental Results}
Table \ref{tab:app_mse_results} and Table \ref{tab:app_mae_results} present the comprehensive prediction results. The best results are highlighted in red, while the second-best are marked in blue. Lower MSE/MAE values indicate higher prediction accuracy. Compared with other time series forecasting (TSF) models, our method achieves the best average performance on most benchmark datasets. These results demonstrate the superior performance of our proposed approach. Furthermore, the encoder of each branch of our model is identical to the original Transformer encoder, with the only difference being the introduction of the PTEM and STEM modules. The experimental results show that even with a minimal dual-branch Transformer structure, integrating our proposed modules can achieve competitive performance on most datasets. These findings further validate the effectiveness of our method.

\paragraph{Ablation Study of PTEM, STEM and Dual‑branch}
In this subsection, we conduct a systematic, quantitative study of the performance gains contributed by PTEM, STEM, and the dual‑branch architecture. Specifically, we independently remove PTEM, STEM, PTEM\&STEM, the temporal branch, and the variable branch. Experiments are performed on the ETTm1, Weather, and ECL datasets, which contain 7, 21, and 321 variables, respectively, representing three typical scenarios with small, medium, and large numbers of variables. The results are shown in Table \ref{tab:ablation_main_text}. From the results, we can draw three conclusions. (1) Using the original algorithm (Original) and the dual‑branch framework (corresponding ``–PTEM\&STEM'' column) as reference points, we find that adding PTEM alone to the dual‑branch framework (corresponding ``–STEM'' column) accounts for about 61.48\% of the gain achieved when both PTEM and STEM are added simultaneously (Original). Adding STEM alone (corresponding ``–PTEM'' column) accounts for about 56.71\% of that gain. (2) Taking the pure variable-branch algorithm (corresponding ``-Temporal Branch\&PTEM\&STEM'' column) as the baseline. Adding the temporal branch to it (corresponding ``\mbox{–PTEM \& STEM}'' column) lowers the MSE by $1.72\%$, $1.55\%$, and $1.12\%$ on ETTm1, Weather, and ECL, respectively. When PTEM \& STEM are further incorporated into this dual-branch framework (Original), the MSE reductions increase to $4.67\%$, $6.2\%$, and $7.30\%$ on the same datasets. Hence, the dual-branch design itself contributes $36.84\%$, $25\%$, and $15.34\%$ of the total MSE reduction on ETTm1 (7 variables), Weather (21 variables), and ECL (321 variables), respectively. This pattern suggests that the fewer the variables, the more pronounced the benefit of adding the temporal branch, while PTEM \& STEM remain the dominant source of performance improvement overall. (3) The columns ``–Temporal Branch'' and ``–Variable Branch'' reveal that the temporal branch alone performs worse than the variable branch. Nevertheless, in settings with scarce variables (ETTm1), the temporal branch provides valuable supplementary information, making its inclusion practically necessary.

\begin{table*}
\caption{Full \textbf{MSE} results for the TSF task. All baseline models' input length is 96, and the prediction lengths include $\{96,192,336,720\}$. The results of our method are averaged over five random seeds, and the standard deviation across the five runs is reported after the ``$\pm$'' symbol.}
\renewcommand{\arraystretch}{1}
\centering
\resizebox{0.8\columnwidth}{!}{
\begin{threeparttable}
\begin{small}
\renewcommand{\multirowsetup}{\centering}
\setlength{\tabcolsep}{1pt}
\begin{tabular}{c|c|c|c|c|c|c|c|c|c|c|c|c}
\toprule
\multicolumn{2}{c}{\multirow{2}{*}{Models}} &
  \multicolumn{1}{c}{\rotatebox{0}{\scalebox{0.8}{\textbf{CDTF}}}} &
  \multicolumn{1}{c}{\rotatebox{0}{\scalebox{0.8}{PatchTST}}} &
  \multicolumn{1}{c}{\rotatebox{0}{\scalebox{0.8}{iTransformer}}} &
  \multicolumn{1}{c}{\rotatebox{0}{\scalebox{0.8}{Crossformer}}} &
  \multicolumn{1}{c}{\rotatebox{0}{\scalebox{0.8}{TimesNet}}} &
  \multicolumn{1}{c}{\rotatebox{0}{\scalebox{0.8}{DLinear}}} &
  \multicolumn{1}{c}{\rotatebox{0}{\scalebox{0.8}{DSformer}}} &
  \multicolumn{1}{c}{\rotatebox{0}{\scalebox{0.8}{Transformer}}} &
  \multicolumn{1}{c}{\rotatebox{0}{\scalebox{0.8}{SpareTSF}}} &
  \multicolumn{1}{c}{\rotatebox{0}{\scalebox{0.8}{PRReg CI}}} &
  \multicolumn{1}{c}{\rotatebox{0}{\scalebox{0.8}{TIDE}}} \\
\multicolumn{2}{c}{} &
  \scalebox{0.8}{(Ours)} & \scalebox{0.8}{(2023)} & \scalebox{0.8}{(2024)} &
  \scalebox{0.8}{(2023)} & \scalebox{0.8}{(2023)} & \scalebox{0.8}{(2023)} &
  \scalebox{0.8}{(2023)} & \scalebox{0.8}{(2017)} & \scalebox{0.8}{(2024)} &
  \scalebox{0.8}{(2024)} & \scalebox{0.8}{(2023)} \\
\cmidrule(lr){3-13}
\multicolumn{2}{c}{Metric} & \multicolumn{11}{c}{MSE (↓)} \\
\toprule
\multirow{5}{*}{\rotatebox{90}{ETTh1}}
  & 96  & \textcolor{blue}{0.382}\scalebox{0.9}{$\pm$0.002} & 0.414 & 0.386 & 0.423 & 0.384 & 0.386 & 0.401 & 0.611 & \textcolor{red}{0.374} & 0.409 & 0.479 \\
  & 192 & \textcolor{blue}{0.436}\scalebox{0.9}{$\pm$0.003} & 0.460 & 0.441 & 0.471 & \textcolor{blue}{0.436} & 0.437 & 0.441 & 0.697 & \textcolor{red}{0.419} & 0.454 & 0.525 \\
  & 336 & \textcolor{blue}{0.476}\scalebox{0.9}{$\pm$0.002} & 0.501 & 0.487 & 0.570 & 0.491 & 0.481 & 0.486 & 0.770 & \textcolor{red}{0.463} & 0.514 & 0.565 \\
  & 720 & \textcolor{blue}{0.486}\scalebox{0.9}{$\pm$0.001} & 0.500 & 0.503 & 0.653 & 0.521 & 0.519 & 0.493 & 0.796 & \textcolor{red}{0.477} & 0.530 & 0.594 \\
\cmidrule(lr){2-13}
  & Avg & \textcolor{blue}{0.445} & 0.469 & 0.454 & 0.529 & 0.458 & 0.456 & 0.455 & 0.718 & \textcolor{red}{0.433} & 0.477 & 0.541 \\
\midrule
\multirow{5}{*}{\rotatebox{90}{ETTh2}}
  & 96  & \textcolor{red}{0.289}\scalebox{0.9}{$\pm$0.001} & 0.302 & \textcolor{blue}{0.297} & 0.745 & 0.340 & 0.333 & 0.319 & 0.470 & 0.308 & 0.300 & 0.400 \\
  & 192 & \textcolor{red}{0.377}\scalebox{0.9}{$\pm$0.001} & 0.388 & \textcolor{blue}{0.380} & 0.877 & 0.402 & 0.477 & 0.400 & 0.601 & 0.388 & 0.385 & 0.528 \\
  & 336 & \textcolor{red}{0.410}\scalebox{0.9}{$\pm$0.000} & \textcolor{blue}{0.426} & 0.428 & 1.043 & 0.452 & 0.594 & 0.462 & 0.676 & 0.431 & 0.440 & 0.643 \\
  & 720 & \textcolor{red}{0.411}\scalebox{0.9}{$\pm$0.002} & 0.431 & 0.427 & 1.104 & 0.462 & 0.831 & 0.457 & 0.676 & 0.432 & \textcolor{blue}{0.426} & 0.874 \\
\cmidrule(lr){2-13}
  & Avg & \textcolor{red}{0.372} & 0.387 & \textcolor{blue}{0.383} & 0.942 & 0.414 & 0.559 & 0.410 & 0.605 & 0.390 & 0.388 & 0.611 \\
\midrule
\multirow{5}{*}{\rotatebox{90}{ETTm1}}
  & 96  & \textcolor{red}{0.317}\scalebox{0.9}{$\pm$0.002} & \textcolor{blue}{0.329} & 0.334 & 0.404 & 0.338 & 0.345 & 0.342 & 0.528 & 0.349 & 0.344 & 0.364 \\
  & 192 & \textcolor{red}{0.367}\scalebox{0.9}{$\pm$0.002} & \textcolor{red}{0.367} & 0.377 & 0.450 & \textcolor{blue}{0.374} & 0.380 & 0.385 & 0.596 & 0.385 & 0.384 & 0.398 \\
  & 336 & 0.403\scalebox{0.9}{$\pm$0.003} & \textcolor{red}{0.399} & 0.426 & 0.532 & 0.410 & 0.413 & \textcolor{blue}{0.402} & 0.674 & 0.412 & 0.442 & 0.428 \\
  & 720 & \textcolor{blue}{0.466}\scalebox{0.9}{$\pm$0.000} & \textcolor{red}{0.454} & 0.491 & 0.666 & 0.478 & 0.474 & 0.472 & 0.777 & 0.474 & 0.479 & 0.487 \\
\cmidrule(lr){2-13}
  & Avg & \textcolor{blue}{0.388} & \textcolor{red}{0.387} & 0.407 & 0.513 & 0.400 & 0.403 & 0.400 & 0.643 & 0.405 & 0.412 & 0.419 \\
\midrule
\multirow{5}{*}{\rotatebox{90}{ETTm2}}
  & 96  & \textcolor{red}{0.172}\scalebox{0.9}{$\pm$0.001} & \textcolor{blue}{0.175} & 0.180 & 0.287 & 0.187 & 0.193 & 0.184 & 0.285 & 0.180 & 0.187 & 0.207 \\
  & 192 & \textcolor{red}{0.238}\scalebox{0.9}{$\pm$0.002} & \textcolor{blue}{0.241} & 0.250 & 0.414 & 0.249 & 0.284 & 0.248 & 0.396 & 0.243 & 0.253 & 0.290 \\
  & 336 & \textcolor{blue}{0.305}\scalebox{0.9}{$\pm$0.002} & \textcolor{blue}{0.305} & 0.311 & 0.597 & 0.321 & 0.369 & 0.335 & 0.492 & \textcolor{red}{0.302} & 0.320 & 0.377 \\
  & 720 & 0.405\scalebox{0.9}{$\pm$0.003} & \textcolor{blue}{0.402} & 0.412 & 1.730 & 0.408 & 0.554 & 0.412 & 0.652 & \textcolor{red}{0.398} & 0.440 & 0.558 \\
\cmidrule(lr){2-13}
  & Avg & \textcolor{red}{0.280} & \textcolor{blue}{0.281} & 0.288 & 0.757 & 0.291 & 0.350 & 0.295 & 0.456 & \textcolor{blue}{0.281} & 0.300 & 0.358 \\
\midrule
\multirow{5}{*}{\rotatebox{90}{ECL}}
  & 96  & \textcolor{red}{0.136}\scalebox{0.9}{$\pm$0.001} & 0.195 & 0.148 & 0.219 & 0.168 & 0.197 & 0.173 & 0.260 & 0.194 & \textcolor{blue}{0.146} & 0.237 \\
  & 192 & \textcolor{red}{0.154}\scalebox{0.9}{$\pm$0.000} & 0.199 & 0.162 & 0.231 & 0.184 & 0.196 & 0.183 & 0.266 & 0.197 & \textcolor{blue}{0.160} & 0.236 \\
  & 336 & \textcolor{red}{0.168}\scalebox{0.9}{$\pm$0.001} & 0.215 & 0.178 & 0.246 & 0.198 & 0.209 & 0.203 & 0.280 & 0.209 & \textcolor{blue}{0.172} & 0.249 \\
  & 720 & \textcolor{red}{0.203}\scalebox{0.9}{$\pm$0.004} & 0.256 & 0.225 & 0.280 & 0.220 & 0.245 & 0.259 & 0.302 & 0.251 & \textcolor{blue}{0.209} & 0.284 \\
\cmidrule(lr){2-13}
  & Avg & \textcolor{red}{0.165} & 0.216 & 0.178 & 0.244 & 0.192 & 0.212 & 0.205 & 0.277 & 0.213 & \textcolor{blue}{0.172} & 0.251 \\
\midrule
\multirow{5}{*}{\rotatebox{90}{Traffic}}
  & 96  & \textcolor{red}{0.384}\scalebox{0.9}{$\pm$0.003} & 0.544 & \textcolor{blue}{0.395} & 0.522 & 0.593 & 0.650 & 0.529 & 0.647 & 0.543 & 0.403 & 0.805 \\
  
  & 192 & \textcolor{red}{0.399}\scalebox{0.9}{$\pm$0.002} & 0.540 & 0.417 & 0.530 & 0.617 & 0.598 & 0.533 & 0.649 & 0.555 & \textcolor{blue}{0.416} & 0.756 \\
  & 336 & \textcolor{red}{0.418}\scalebox{0.9}{$\pm$0.002} & 0.551 & \textcolor{blue}{0.433} & 0.558 & 0.629 & 0.605 & 0.545 & 0.667 & 0.565 & 0.448 & 0.762 \\
  & 720 & \textcolor{red}{0.456}\scalebox{0.9}{$\pm$0.004} & 0.586 & \textcolor{blue}{0.467} & 0.589 & 0.640 & 0.645 & 0.583 & 0.697 & 0.639 & 0.471 & 0.719 \\
\cmidrule(lr){2-13}
  & Avg & \textcolor{red}{0.414} & 0.555 & \textcolor{blue}{0.428} & 0.550 & 0.620 & 0.625 & 0.548 & 0.665 & 0.576 & 0.435 & 0.760 \\
\midrule
\multirow{5}{*}{\rotatebox{90}{Weather}}
  & 96  & \textcolor{red}{0.156}\scalebox{0.9}{$\pm$0.001} & 0.177 & 0.174 & \textcolor{blue}{0.158} & 0.172 & 0.196 & 0.162 & 0.395 & 0.180 & 0.171 & 0.202 \\
  & 192 & \textcolor{red}{0.206}\scalebox{0.9}{$\pm$0.001} & 0.225 & 0.221 & \textcolor{red}{0.206} & 0.219 & 0.237 & \textcolor{blue}{0.211} & 0.619 & 0.226 & 0.219 & 0.242 \\
  & 336 & \textcolor{red}{0.263}\scalebox{0.9}{$\pm$0.001} & 0.278 & 0.278 & 0.272 & 0.280 & 0.283 & \textcolor{blue}{0.267} & 0.689 & 0.281 & 0.286 & 0.287 \\
  & 720 & \textcolor{blue}{0.344}\scalebox{0.9}{$\pm$0.000} & 0.354 & 0.358 & 0.398 & 0.365 & 0.345 & \textcolor{red}{0.343} & 0.926 & 0.358 & 0.357 & 0.351 \\
\cmidrule(lr){2-13}
  & Avg & \textcolor{red}{0.242} & 0.259 & 0.258 & 0.259 & 0.259 & 0.265 & \textcolor{blue}{0.246} & 0.657 & 0.261 & 0.258 & 0.271 \\
\midrule
\multirow{5}{*}{\rotatebox{90}{Solar-Energy}}
  & 96  & \textcolor{red}{0.197}\scalebox{0.9}{$\pm$0.002} & 0.234 & \textcolor{blue}{0.203} & 0.310 & 0.250 & 0.290 & 0.247 & 0.386 & 0.245 & 0.204 & 0.312 \\
  & 192 & \textcolor{red}{0.228}\scalebox{0.9}{$\pm$0.002} & 0.267 & \textcolor{blue}{0.233} & 0.734 & 0.296 & 0.320 & 0.288 & 0.444 & 0.255 & 0.249 & 0.339 \\
  & 336 & \textcolor{red}{0.242}\scalebox{0.9}{$\pm$0.002} & 0.290 & \textcolor{blue}{0.248} & 0.750 & 0.319 & 0.353 & 0.329 & 0.471 & 0.259 & 0.260 & 0.368 \\
  & 720 & \textcolor{red}{0.248}\scalebox{0.9}{$\pm$0.003} & 0.289 & \textcolor{blue}{0.249} & 0.769 & 0.338 & 0.356 & 0.341 & 0.473 & 0.262 & 0.271 & 0.370 \\
\cmidrule(lr){2-13}
  & Avg & \textcolor{red}{0.229} & 0.270 & \textcolor{blue}{0.233} & 0.641 & 0.301 & 0.330 & 0.301 & 0.442 & 0.255 & 0.246 & 0.347 \\
\bottomrule
\end{tabular}
\end{small}
\end{threeparttable}}
\label{tab:app_mse_results}
\end{table*}

\begin{table*}
\caption{Full \textbf{MAE} results for the TSF task. All baseline models' input length is 96, and the prediction lengths include $\{96,192,336,720\}$. The results of our method are averaged over five random seeds, and the standard deviation across the five runs is reported after the ``$\pm$'' symbol.}
\renewcommand{\arraystretch}{1}
\centering
\resizebox{0.8\columnwidth}{!}{
\begin{threeparttable}
\begin{small}
\renewcommand{\multirowsetup}{\centering}
\setlength{\tabcolsep}{1pt}
\begin{tabular}{c|c|c|c|c|c|c|c|c|c|c|c|c}
\toprule
\multicolumn{2}{c}{\multirow{2}{*}{Models}} &
  \multicolumn{1}{c}{\rotatebox{0}{\scalebox{0.8}{\textbf{CDTF}}}} &
  \multicolumn{1}{c}{\rotatebox{0}{\scalebox{0.8}{PatchTST}}} &
  \multicolumn{1}{c}{\rotatebox{0}{\scalebox{0.8}{iTransformer}}} &
  \multicolumn{1}{c}{\rotatebox{0}{\scalebox{0.8}{Crossformer}}} &
  \multicolumn{1}{c}{\rotatebox{0}{\scalebox{0.8}{TimesNet}}} &
  \multicolumn{1}{c}{\rotatebox{0}{\scalebox{0.8}{DLinear}}} &
  \multicolumn{1}{c}{\rotatebox{0}{\scalebox{0.8}{DSformer}}} &
  \multicolumn{1}{c}{\rotatebox{0}{\scalebox{0.8}{Transformer}}} &
  \multicolumn{1}{c}{\rotatebox{0}{\scalebox{0.8}{SpareTSF}}} &
  \multicolumn{1}{c}{\rotatebox{0}{\scalebox{0.8}{PRReg CI}}} &
  \multicolumn{1}{c}{\rotatebox{0}{\scalebox{0.8}{TIDE}}} \\
\multicolumn{2}{c}{} &
  \scalebox{0.8}{(Ours)} & \scalebox{0.8}{(2023)} & \scalebox{0.8}{(2024)} &
  \scalebox{0.8}{(2023)} & \scalebox{0.8}{(2023)} & \scalebox{0.8}{(2023)} &
  \scalebox{0.8}{(2023)} & \scalebox{0.8}{(2017)} & \scalebox{0.8}{(2024)} &
  \scalebox{0.8}{(2024)} & \scalebox{0.8}{(2023)} \\
\cmidrule(lr){3-13}
\multicolumn{2}{c}{Metric} & \multicolumn{11}{c}{MAE (↓)} \\
\toprule
\multirow{5}{*}{\rotatebox{90}{ETTh1}}
  & 96  & \textcolor{blue}{0.399}\scalebox{0.9}{$\pm$0.001} & 0.419 & 0.405 & 0.448 & 0.402 & 0.400 & 0.412 & 0.539 & \textcolor{red}{0.390} & 0.408 & 0.464 \\
  & 192 & 0.431\scalebox{0.9}{$\pm$0.002} & 0.445 & 0.436 & 0.474 & \textcolor{blue}{0.429} & 0.432 & 0.435 & 0.580 & \textcolor{red}{0.420} & 0.447 & 0.492 \\
  & 336 & \textcolor{blue}{0.450}\scalebox{0.9}{$\pm$0.003} & 0.466 & 0.458 & 0.546 & 0.469 & 0.459 & 0.456 & 0.609 & \textcolor{red}{0.440} & 0.483 & 0.515 \\
  & 720 & 0.477\scalebox{0.9}{$\pm$0.003} & 0.488 & 0.491 & 0.621 & 0.500 & 0.516 & \textcolor{blue}{0.475} & 0.653 & \textcolor{red}{0.472} & 0.490 & 0.558 \\
\cmidrule(lr){2-13}
  & Avg & \textcolor{blue}{0.439} & 0.454 & 0.447 & 0.522 & 0.450 & 0.452 & 0.446 & 0.594 & \textcolor{red}{0.431} & 0.457 & 0.507 \\
\midrule
\multirow{5}{*}{\rotatebox{90}{ETTh2}}
  & 96  & \textcolor{red}{0.342}\scalebox{0.9}{$\pm$0.000} & 0.348 & \textcolor{blue}{0.347} & 0.584 & 0.374 & 0.387 & 0.359 & 0.461 & 0.354 & 0.363 & 0.440 \\
  & 192 & \textcolor{blue}{0.400}\scalebox{0.9}{$\pm$0.001} & \textcolor{blue}{0.400} & \textcolor{blue}{0.400} & 0.656 & 0.414 & 0.476 & 0.410 & 0.532 & \textcolor{red}{0.398} & 0.414 & 0.509 \\
  & 336 & \textcolor{red}{0.425}\scalebox{0.9}{$\pm$0.002} & 0.433 & \textcolor{blue}{0.432} & 0.731 & 0.452 & 0.541 & 0.451 & 0.575 & 0.443 & 0.452 & 0.571 \\
  & 720 & \textcolor{red}{0.437}\scalebox{0.9}{$\pm$0.001} & 0.446 & \textcolor{blue}{0.445} & 0.763 & 0.468 & 0.657 & 0.463 & 0.592 & 0.447 & 0.466 & 0.679 \\
\cmidrule(lr){2-13}
  & Avg & \textcolor{red}{0.401} & \textcolor{blue}{0.407} & \textcolor{blue}{0.407} & 0.684 & 0.427 & 0.515 & 0.421 & 0.541 & 0.411 & 0.424 & 0.550 \\
\midrule
\multirow{5}{*}{\rotatebox{90}{ETTm1}}
  & 96  & \textcolor{red}{0.357}\scalebox{0.9}{$\pm$0.002} & \textcolor{blue}{0.367} & 0.368 & 0.426 & 0.375 & 0.372 & 0.375 & 0.489 & 0.381 & 0.384 & 0.387 \\
  & 192 & \textcolor{blue}{0.386}\scalebox{0.9}{$\pm$0.002} & \textcolor{red}{0.385} & 0.391 & 0.451 & 0.387 & 0.389 & 0.392 & 0.520 & 0.397 & 0.399 & 0.404 \\
  & 336 & \textcolor{red}{0.407}\scalebox{0.9}{$\pm$0.002} & \textcolor{blue}{0.410} & 0.420 & 0.515 & 0.411 & 0.413 & 0.411 & 0.559 & 0.417 & 0.426 & 0.425 \\
  & 720 & \textcolor{blue}{0.443}\scalebox{0.9}{$\pm$0.003} & \textcolor{red}{0.439} & 0.459 & 0.589 & 0.450 & 0.453 & 0.448 & 0.610 & 0.450 & 0.455 & 0.461 \\
\cmidrule(lr){2-13}
  & Avg & \textcolor{red}{0.398} & \textcolor{blue}{0.400} & 0.410 & 0.496 & 0.406 & 0.407 & 0.407 & 0.545 & 0.411 & 0.416 & 0.419 \\
\midrule
\multirow{5}{*}{\rotatebox{90}{ETTm2}}
  & 96  & \textcolor{red}{0.258}\scalebox{0.9}{$\pm$0.001} & \textcolor{blue}{0.259} & 0.264 & 0.366 & 0.267 & 0.292 & 0.269 & 0.351 & \textcolor{blue}{0.259} & 0.278 & 0.305 \\
  & 192 & \textcolor{red}{0.300}\scalebox{0.9}{$\pm$0.002} & \textcolor{blue}{0.302} & 0.309 & 0.492 & 0.309 & 0.362 & 0.306 & 0.411 & 0.308 & 0.329 & 0.364 \\
  & 336 & 0.344\scalebox{0.9}{$\pm$0.002} & \textcolor{blue}{0.343} & 0.348 & 0.542 & 0.351 & 0.427 & 0.365 & 0.463 & \textcolor{red}{0.342} & 0.366 & 0.422 \\
  & 720 & \textcolor{blue}{0.400}\scalebox{0.9}{$\pm$0.003} & \textcolor{blue}{0.400} & 0.407 & 1.042 & 0.403 & 0.522 & 0.409 & 0.541 & \textcolor{red}{0.398} & 0.428 & 0.524 \\
\cmidrule(lr){2-13}
  & Avg & \textcolor{red}{0.326} & \textcolor{red}{0.326} & 0.332 & 0.610 & 0.333 & 0.401 & 0.337 & 0.442 & \textcolor{blue}{0.327} & 0.350 & 0.404 \\
\midrule
\multirow{5}{*}{\rotatebox{90}{ECL}}
  & 96  & \textcolor{red}{0.233}\scalebox{0.9}{$\pm$0.000} & 0.285 & 0.240 & 0.314 & 0.272 & 0.282 & 0.269 & 0.358 & 0.269 & \textcolor{blue}{0.239} & 0.329 \\
  & 192 & \textcolor{red}{0.251}\scalebox{0.9}{$\pm$0.001} & 0.289 & \textcolor{blue}{0.253} & 0.322 & 0.289 & 0.285 & 0.280 & 0.367 & 0.273 & 0.262 & 0.330 \\
  & 336 & \textcolor{red}{0.266}\scalebox{0.9}{$\pm$0.002} & 0.305 & \textcolor{blue}{0.269} & 0.337 & 0.300 & 0.301 & 0.297 & 0.375 & 0.287 & 0.276 & 0.344 \\
  & 720 & \textcolor{red}{0.294}\scalebox{0.9}{$\pm$0.003} & 0.337 & 0.317 & 0.363 & 0.320 & 0.333 & 0.340 & 0.386 & 0.321 & \textcolor{blue}{0.306} & 0.373 \\
\cmidrule(lr){2-13}
  & Avg & \textcolor{red}{0.261} & 0.304 & \textcolor{blue}{0.270} & 0.334 & 0.295 & 0.300 & 0.297 & 0.372 & 0.288 & 0.271 & 0.344 \\
\midrule
\multirow{5}{*}{\rotatebox{90}{Traffic}}
  & 96  & \textcolor{blue}{0.269}\scalebox{0.9}{$\pm$0.002} & 0.359 & \textcolor{red}{0.268} & 0.290 & 0.321 & 0.396 & 0.370 & 0.357 & 0.332 & 0.279 & 0.493 \\
  & 192 & \textcolor{red}{0.275}\scalebox{0.9}{$\pm$0.002} & 0.354 & \textcolor{blue}{0.276} & 0.293 & 0.336 & 0.370 & 0.366 & 0.356 & 0.356 & 0.301 & 0.474 \\
  & 336 & \textcolor{blue}{0.285}\scalebox{0.9}{$\pm$0.002} & 0.358 & \textcolor{red}{0.283} & 0.305 & 0.336 & 0.373 & 0.370 & 0.364 & 0.360 & 0.305 & 0.477 \\
  & 720 & \textcolor{red}{0.299}\scalebox{0.9}{$\pm$0.003} & 0.375 & \textcolor{blue}{0.302} & 0.328 & 0.350 & 0.394 & 0.386 & 0.376 & 0.352 & 0.322 & 0.449 \\
\cmidrule(lr){2-13}
  & Avg & \textcolor{red}{0.282} & 0.362 & \textcolor{red}{0.282} & 0.304 & 0.336 & 0.383 & 0.373 & 0.363 & 0.350 & \textcolor{blue}{0.302} & 0.473 \\
\midrule
\multirow{5}{*}{\rotatebox{90}{Weather}}
  & 96  & \textcolor{red}{0.203}\scalebox{0.9}{$\pm$0.000} & 0.218 & 0.214 & 0.230 & 0.220 & 0.255 & \textcolor{blue}{0.207} & 0.427 & 0.230 & 0.217 & 0.261 \\
  & 192 & \textcolor{red}{0.251}\scalebox{0.9}{$\pm$0.002} & 0.259 & 0.254 & 0.277 & 0.261 & 0.296 & \textcolor{blue}{0.252} & 0.560 & 0.271 & 0.270 & 0.298 \\
  & 336 & \textcolor{red}{0.290}\scalebox{0.9}{$\pm$0.001} & 0.297 & 0.296 & 0.335 & 0.306 & 0.335 & \textcolor{blue}{0.294} & 0.594 & 0.311 & 0.314 & 0.335 \\
  & 720 & \textcolor{red}{0.341}\scalebox{0.9}{$\pm$0.002} & 0.348 & 0.349 & 0.418 & 0.359 & 0.381 & \textcolor{blue}{0.343} & 0.710 & 0.360 & 0.349 & 0.386 \\
\cmidrule(lr){2-13}
  & Avg & \textcolor{red}{0.271} & 0.281 & 0.279 & 0.315 & 0.287 & 0.317 & \textcolor{blue}{0.274} & 0.572 & 0.293 & 0.288 & 0.320 \\
\midrule
\multirow{5}{*}{\rotatebox{90}{Solar-Energy}}
  & 96  & \textcolor{red}{0.229}\scalebox{0.9}{$\pm$0.001} & 0.286 & \textcolor{blue}{0.237} & 0.331 & 0.292 & 0.378 & 0.292 & 0.372 & 0.291 & 0.251 & 0.399 \\
  & 192 & \textcolor{red}{0.260}\scalebox{0.9}{$\pm$0.002} & 0.310 & \textcolor{blue}{0.261} & 0.725 & 0.318 & 0.398 & 0.320 & 0.409 & 0.294 & 0.272 & 0.416 \\
  & 336 & \textcolor{red}{0.271}\scalebox{0.9}{$\pm$0.002} & 0.315 & \textcolor{blue}{0.273} & 0.735 & 0.330 & 0.415 & 0.344 & 0.429 & 0.300 & 0.285 & 0.430 \\
  & 720 & \textcolor{blue}{0.276}\scalebox{0.9}{$\pm$0.002} & 0.317 & \textcolor{red}{0.275} & 0.765 & 0.337 & 0.413 & 0.352 & 0.432 & 0.303 & 0.292 & 0.425 \\
\cmidrule(lr){2-13}
  & Avg & \textcolor{red}{0.259} & 0.307 & \textcolor{blue}{0.262} & 0.639 & 0.319 & 0.401 & 0.327 & 0.411 & 0.297 & 0.275 & 0.417 \\
\bottomrule
\end{tabular}
\end{small}
\end{threeparttable}}
\label{tab:app_mae_results}
\end{table*}

\begin{table*}
  \centering
  \caption{Ablation study of PTEM, STEM, temporal branch, and variable branch. Look-back length is fixed at 96; prediction lengths $\,\in\!\{96,192,336,720\}$.}
  \label{tab:ablation_main_text}
  \resizebox{\textwidth}{!}{%
  \begin{tabular}{l|c|c|c|c|c|c|c|c}
    \toprule
    \textbf{Dataset} & \textbf{Pred Len} &
    \textbf{Original} & \textbf{-PTEM} & \textbf{-STEM} & \textbf{-PTEM \& STEM} &
    \textbf{-Temporal Branch\&PTEM \& STEM} & \textbf{-Temporal Branch} & \textbf{-Variable Branch} \\
    \midrule
    \multirow{5}{*}{ETTm1}
        & 96  & \textbf{0.317 / 0.357} & 0.326 / 0.361 & 0.321 / 0.359 & 0.327 / 0.366 & 0.334 / 0.368 & 0.328 / 0.365& 0.408 / 0.415\\
        & 192 & \textbf{0.367 / 0.386} & 0.372 / 0.387 & 0.371 / 0.388 & 0.374 / 0.390 & 0.377 / 0.391 & 0.374 / 0.390 & 0.461 / 0.441\\
        & 336 & \textbf{0.403 / 0.407} & 0.411 / 0.410 & 0.407 / 0.409 & 0.415 / 0.415 & 0.426 / 0.420 & 0.421 / 0.415 & 0.489 / 0.481\\
        & 720 & \textbf{0.466 / 0.443} & 0.472 / 0.446 & 0.473 / 0.445 & 0.482 / 0.452 & 0.491 / 0.459 & 0.485 / 0.456 & 0.542 / 0.512\\
        & Avg & \textbf{0.388 / 0.398} & 0.395 / 0.401 & 0.393 / 0.400 & 0.400 / 0.406 & 0.407 / 0.410 & 0.402 / 0.406 & 0.494 / 0.462 \\[2pt]
    \cmidrule{1-9}
    \multirow{5}{*}{ECL}
    & 96  & \textbf{0.136 / 0.233} & 0.140 / 0.234 & 0.138 / 0.233 & 0.149 / 0.238 & 0.148 / 0.240 & 0.141 / 0.235 & 0.270 / 0.362 \\
    & 192 & \textbf{0.154 / 0.251} & 0.156 / 0.253 & 0.157 / 0.252 & 0.160 / 0.254 & 0.162 / 0.253 & 0.156 / 0.252 & 0.282 / 0.374 \\
    & 336 & \textbf{0.168 / 0.266} & 0.171 / 0.269 & 0.171 / 0.267 & 0.175 / 0.272 & 0.178 / 0.269 & 0.169 / 0.264 & 0.288 / 0.382 \\
    & 720 & \textbf{0.203 / 0.294} & 0.207 / 0.297 & 0.204 / 0.304 & 0.221 / 0.315 & 0.225 / 0.317 & 0.209 / 0.305 & 0.291 / 0.387 \\
    & Avg & \textbf{0.165 / 0.261} & 0.168 / 0.263 & 0.169 / 0.264 & 0.176 / 0.270 & 0.178 / 0.270 & 0.169 / 0.264 & 0.283 / 0.376 \\[2pt]
    \cmidrule{1-9}
    \multirow{5}{*}{Weather}
    & 96  & \textbf{0.156 / 0.203} & 0.166 / 0.209 & 0.165 / 0.210 & 0.172/0.212 & 0.174 / 0.214 & 0.165 / 0.208 & 0.184 / 0.275 \\
    & 192 & \textbf{0.206 / 0.251} &  0.212 / 0.253 & 0.210 / 0.252 & 0.219/0.254 & 0.221 / 0.254 & 0.210 / 0.252 & 0.225 / 0.311 \\
    & 336 & \textbf{0.263 / 0.290} & 0.268 / 0.292 & 0.271 / 0.292 & 0.274/0.295 & 0.278 / 0.296 & 0.266 / 0.293 & 0.344 / 0.372 \\
    & 720 & \textbf{0.344 / 0.341} & 0.348 / 0.344 & 0.347  / 0.346 & 0.351/0.346 & 0.358 / 0.349 & 0.350 / 0.345 & 0.480 / 0.498 \\
    & Avg & \textbf{0.242 / 0.271} & 0.249 / 0.275 & 0.248 / 0.275 & 0.254/0.277 & 0.258 / 0.279 & 0.248 / 0.275 & 0.316 / 0.364 \\
    \bottomrule
  \end{tabular}}
\end{table*}

\printcredits

\end{document}